\documentclass{article} 
\usepackage[preprint]{colm2025_conference}

\usepackage{microtype}
\usepackage{hyperref}
\usepackage{url}
\usepackage{booktabs}
\usepackage{subcaption}
\usepackage{graphicx}
\usepackage{lineno}

\usepackage{amsmath,amsfonts,bm}
\usepackage{wrapfig}

\usepackage{spverbatim}
\usepackage{tikz}
\usepackage{xspace}
\definecolor{darkred}{RGB}{150,0,0}
\definecolor{darkgreen}{RGB}{0,150,0}
\definecolor{darkblue}{RGB}{0,0,200}
\hypersetup{colorlinks=true, linkcolor=darkred, citecolor=darkgreen, urlcolor=darkblue}

\newcommand{\alg}{\textsc{TLDR}\xspace}

\def\eqref#1{equation~\ref{#1}}

\def\1{\bm{1}}

\DeclareMathAlphabet{\mathsfit}{\encodingdefault}{\sfdefault}{m}{sl}
\SetMathAlphabet{\mathsfit}{bold}{\encodingdefault}{\sfdefault}{bx}{n}

\usepackage{cleveref}
\definecolor{darkblue}{rgb}{0, 0, 0.5}
\hypersetup{colorlinks=true, citecolor=darkblue, linkcolor=darkblue, urlcolor=darkblue}
\usepackage{enumitem}

\title{Understanding and Improving Token-Efficiency of Reasoning Models}
\title{Enhancing the Efficiency of Small Reasoning Models through Length Penalized Reinforcement Learning}
\title{Efficient Small Reasoning Models through Length Penalized Reinforcement Learning}
\title{Efficient Small Reasoning Models via Response Length Control}
\title{Controllable Response Length for Efficient Small Reasoning Models}
\title{How to make SLMs efficient reasoners?}
\title{Making Small Language Models Efficient Reasoners}
\title{Toward Efficient and Accurate Reasoning in Small-Scale Language Models}
\title{Making Small Language Models Efficient Reasoners:\\Intervention, Supervision, Reinforcement}

\author{Xuechen Zhang\thanks{These authors contributed equally to this work.}, Zijian Huang\footnotemark[1], Chenshun Ni, Ziyang Xiong, Jiasi Chen, Samet Oymak\\
University of Michigan\\
\texttt{\{zxuechen,zijianh,nichensh,xziyang,jiasi,oymak\}@umich.edu} 
}

\begin{document}

\ifcolmsubmission
\linenumbers
\fi

\maketitle

\begin{abstract}

Recent research enhances language model reasoning by scaling test-time compute via longer chain-of-thought traces. This often improves accuracy but also introduces redundancy and high computational cost, especially for small language models distilled with supervised fine-tuning (SFT). In this work, we propose new algorithms to improve token-efficient reasoning with small-scale models by effectively trading off accuracy and computation. We first show that the post-SFT model fails to determine the optimal stopping point of the reasoning process, resulting in verbose and repetitive outputs. Verbosity also significantly varies across wrong vs correct responses. To address these issues, we propose two solutions: (1) Temperature scaling (TS) to control the stopping point for the \emph{thinking phase} and thereby trace length, and (2) \alg: a length-regularized reinforcement learning method based on GRPO that facilitates multi-level trace length control (e.g.~short, medium, long reasoning). Experiments on four reasoning benchmarks, MATH500, AMC, AIME24 and OlympiadBench, demonstrate that TS is highly effective compared to s1's budget forcing approach and \alg significantly improves token efficiency by about 50\% with minimal to no accuracy loss over the SFT baseline. Moreover, \alg also facilitates flexible control over the response length, offering a practical and effective solution for token-efficient reasoning in small models. Ultimately, our work reveals the importance of stopping time control, highlights shortcomings of pure SFT, and provides effective algorithmic recipes.

\end{abstract}
\section{Introduction}
\begin{wrapfigure}{r}{0.5\textwidth}
\centering\vspace{-22pt}
\includegraphics[width=0.5\textwidth]{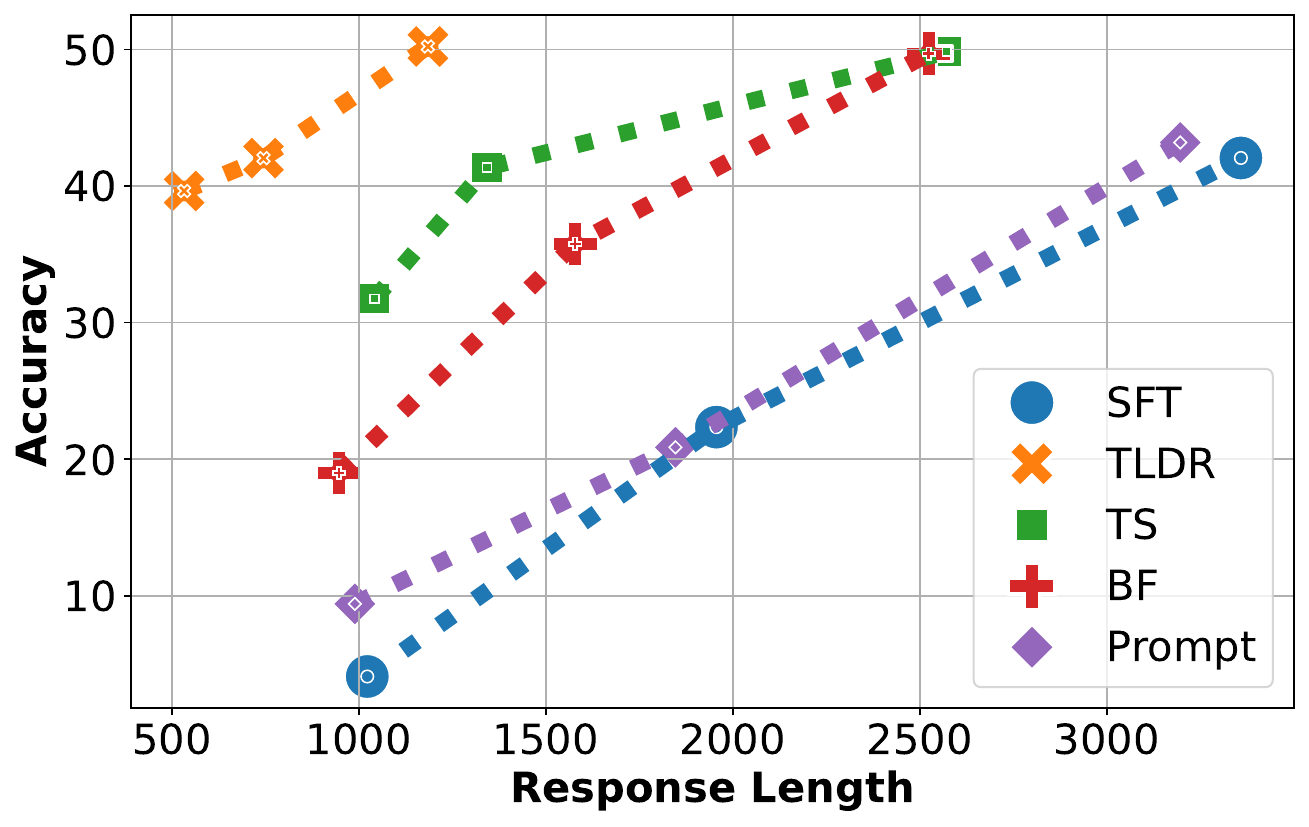}
\vspace{-20pt}
\caption{\small{The average performances across four reasoning benchmarks for 7B models. Our RL-based length-control method \alg enhances token efficiency by over 50\% compared to SFT. Our temperature scaling (TS) method outperforms other test-time intervention techniques that avoid the need for training, such as Budget Forcing (BF) \cite{muennighoff2025s1} and Prompting. The detailed explanation of baselines and our methods are provided in Sec.~\ref{sec:method}.}}\label{fig:3_level_acc_avg_new}
\vspace{-20pt}
\end{wrapfigure}
Recent works, such as OpenAI's o1, DeepSeek-R1, and s1~\cite{jaech2024openai,guo2025deepseek,muennighoff2025s1} have focused on enhancing the reasoning capabilities of language models by generating longer traces equipped with a \emph{thinking} phase. These traces can be extremely long for challenging queries such as AIME or IOI problems. If we wish to make reasoning more efficient, we could either ask for a shorter reasoning trace or use a smaller language model to solve the problem at hand. This raises the question: Given a potentially small language model (SLM) and a maximum trace length, how can we achieve effective reasoning? In this work, we provide new insights and algorithms toward addressing this question.
For training a small model, a go to approach is distillation where we use supervised fine-tuning (SFT) on the samples curated by a more powerful reasoning model like o1 or R1~\cite{guo2025deepseek}. Recent works~\cite{muennighoff2025s1, guo2025deepseek, li2025small} indicate that this straightforward distillation approach on properly curated data significantly enhances the reasoning abilities of small models. We find that pure SFT has a key limitation: these models tend to generate excessively long answers that often contain repetitions. This highlights a major weakness of SFT compared to direct reward optimization via reinforcement learning (e.g.~GRPO). 

We advocate that optimizing the efficiency-accuracy trade-off necessitates explicitly incorporating length regularization within reward or better intervention strategies if an RL training phase is not allowed. To this aim, we introduce \emph{\underline{T}race \underline{L}ength Control for \underline{D}ynamic \underline{R}easoning} (\alg): a length-penalized variation of GRPO. Our evaluations show that, by utilizing a mild penalization, \alg substantially reduces the average response length over GRPO while maintaining or even improving the overall accuracy. This could be viewed as a \emph{free lunch} for the efficiency-performance trade-off. Additionally, we introduce a multi-level variation of \alg that allows for dynamically controlling the response length by directly prompting the model. This eliminates the need for additional model training and offers a flexible approach to controlled generation. Experiments on multiple reasoning benchmarks demonstrate that our method significantly improves token efficiency and efficiency-performance pareto-front. As a training-free intervention method, we also introduce an intuitive temperature scaling strategy that provide a fine-grained control on the sampling of \texttt{<end-of-sequence (EOS)>} token.  Our specific contributions are as follows:



\begin{figure}
\centering
\vspace{-35pt}
\includegraphics[width=\textwidth]{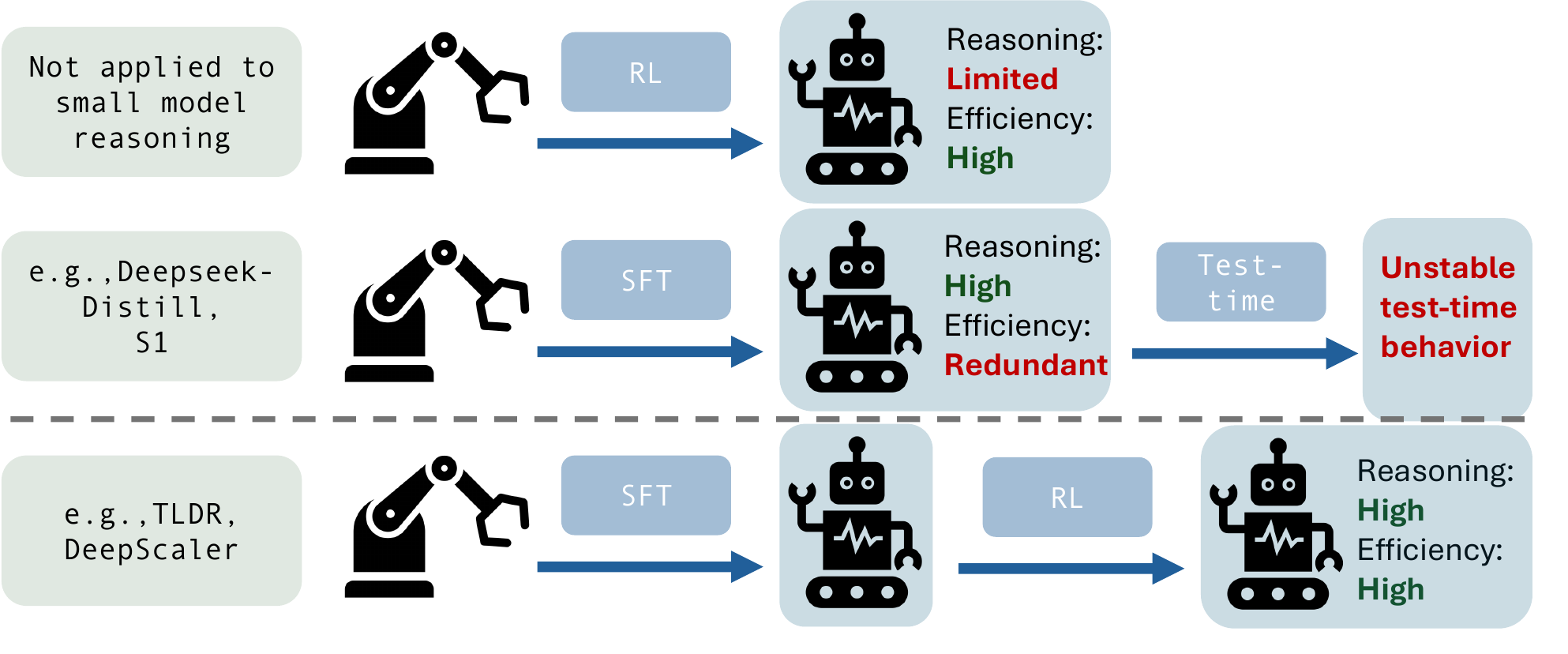}
\vspace{-15pt}
\caption{We explore training strategies for small language models (SLM), focusing on the effects on reasoning capability and test-time compute efficiency. RL improves efficiency, but when used alone, can compromise reasoning performance. While SFT enhances reasoning, it often leads to redundant outputs due to the lack of deliberate stopping point control. Test-time strategies offer limited control over output length and fail to improve performance with increased length consistently. To address these challenges, we propose TLDR, a method that combines SFT with RL to achieve both strong reasoning and token-efficient generation.}
\label{fig:framework}
\end{figure}

\begin{itemize}[leftmargin=*]
    \item \textbf{Rethinking SFT for distillation:} Through extensive experiments comparing SFT, SFT with test-time compute, and reinforcement learning (RL) across models of different sizes, we find that SFT-based methods result in substantial inefficiencies. These arise from redundant generation and repetitions, and notably can result in non-monotonic accuracy progress as the test-time compute budget increases (cf.~Figure \ref{fig: ds-qwen-math}). 
    \item \textbf{Temperature scaling to control stopping time in SFT models:} We propose TS as a training-free model-agnostic method to improve token-efficiency by controlling the sampling probability of the EOS token. Compared to budget forcing or textual prompting, this provides a finer-grained control and yields a better accuracy-efficiency pareto-front with up to 50\% reduction in response length while maintaining accuracy across benchmarks. 
    \item \textbf{Controlled generation via RL with multi-level length penalties} 
    We introduce \alg which integrates length-penalty within GRPO formulation to improve response efficiency. Remarkably, \alg can reduce the response length while preserving overall performance compared to base GRPO. Within \alg, we incorporate multiple penalty levels that can be specified by the user prompt. Through this, we show that \alg can sweep the efficiency-accuracy pareto-front. Comparing pareto-fronts of different model sizes reveal scenarios where an SLM with longer trace (smaller penalty) outperforms an LLM with shorter trace while using fewer total FLOPs.
    
    
\end{itemize}

We remark that recent approaches \cite{muennighoff2025s1, aggarwal2025l1, butcher2024precise, yuan2024following,xu2025scalable,hou2025thinkprune,yang2025towards} have also highlighted the importance of length control. However, with the exception of the recent work L1 \cite{aggarwal2025l1}, these methods do not provide an optimized control over the average or maximum output length. Our evaluations show that lack of explicit optimization results in sub-optimal tradeoffs. \alg compares on par with L1 under an max trace length constraint, while training from a weaker base model. We also contend that enforcing a maximum length constraint is often more desirable in real-world settings and refer the reader to \Cref{sec:experiments} for details. The remainder of this paper is organized as follows.
In \Cref{sec:observations}, we present our experimental characterization of SFT and RL strategies regarding their response length.
Motivated by these observations, in \Cref{sec:observation6} we present our training-free TS method and, in \Cref{sec:method}, we present our length penalty framework.
\Cref{sec:experiments} showcases the empirical results, and \Cref{sec:conclusions} concludes the paper with a discussion.


\begin{figure}
\centering
\begin{subfigure}[b]{0.31\textwidth}
\includegraphics[width=\textwidth]{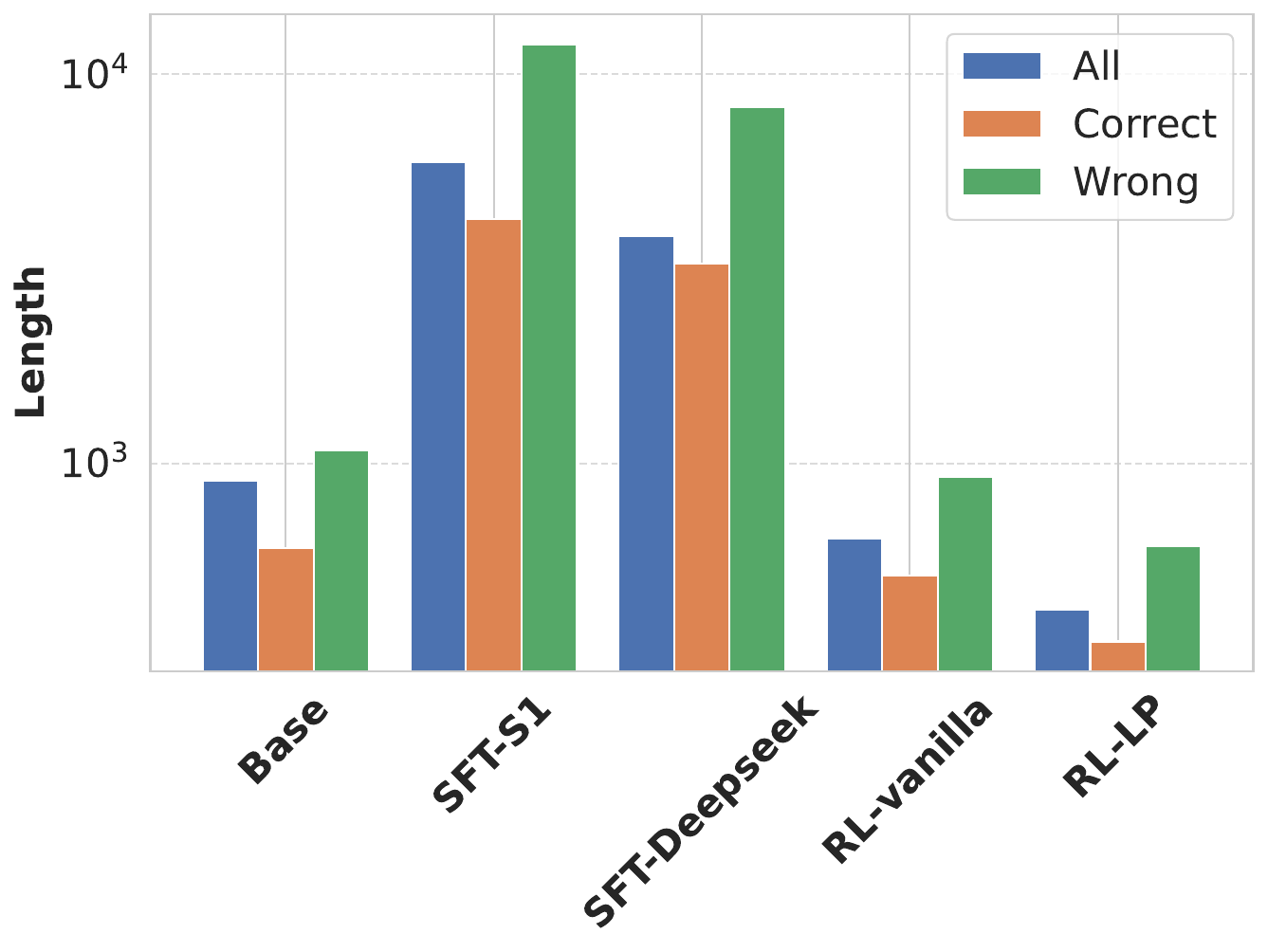}
\vspace{-15pt}
  \caption{Response length for different FT strategies}\label{fig:rl_sft_length}
\end{subfigure}
~~
\begin{subfigure}[b]{0.31\textwidth}
\includegraphics[width=\textwidth]{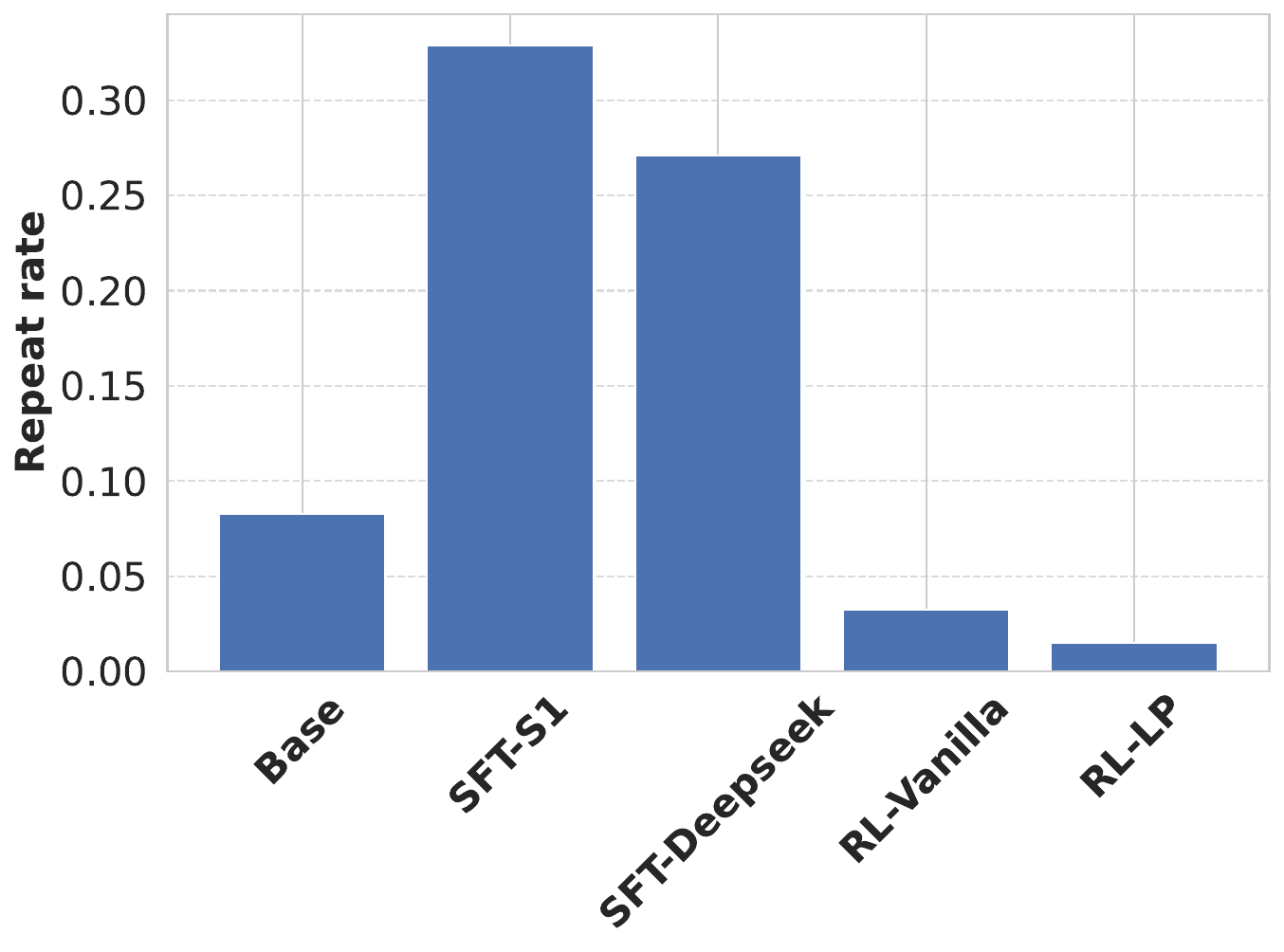}
\vspace{-15pt}
  \caption{Repeat rate for different FT strategies}\label{fig:rl_sft_repeat}
\end{subfigure}
~~
\begin{subfigure}[b]{0.31\textwidth}
\centering
\includegraphics[width=\textwidth]{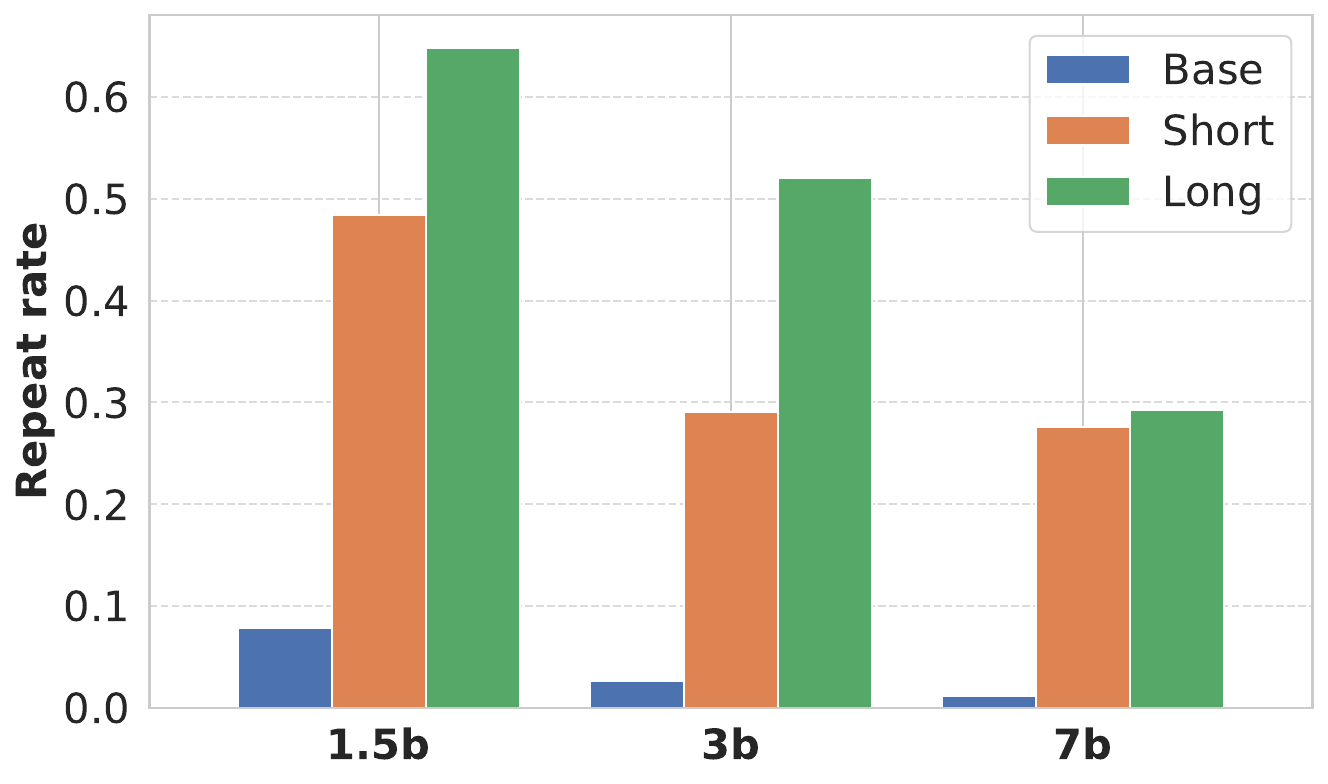}
\vspace{5pt}
  \caption{Repeat rate for different model sizes and FT trace length }\label{fig:scale_repeat}
\end{subfigure}
~~
\caption{SFT models have longer and more repetitive answers when wrong, especially for small models fine-tuned under supervision by long traces.}
\label{fig:rl_sft}
\vspace{-5pt}
\end{figure}

\section{Related Work}
\label{sec:related}

\textbf{Efficient LLMs.} LLMs are widly popular for tasks such as conversational AI, code generation, healthcare, and scientific research. However, their increasing size and computational demands pose challenges for scalability, efficiency, and deployment. This has become a central challenge with the rising importance of test-time compute scaling \cite{snell2024scaling} where the model generates long chain-of-thought traces \cite{muennighoff2025s1,jaech2024openai,guo2025deepseek} or conducts other computation/training \cite{akyurek2024surprising,gozeten2025test} to solve challenging problems. Speculative decoding \cite{leviathan2023fast} accelerates autoregressive generation by leveraging draft models to predict multiple tokens in parallel. MoE-based architectures \cite{shazeer2017outrageously} reduces computational cost while maintaining performance by activating only a subset of parameters per input. Quantization techniques, such as GPTQ \cite{frantar2022gptq} and AWQ \cite{lin2024awq}, compress models into lower precision formats, achieving speed-ups with minimal accuracy loss. Cascade-based methods \cite{chen2023frugalgpt,zhang2024efficient,gupta2024language} save the inference cost by choosing the appropriate models and prompts by either rule-based selection or training an RL algorithm. Instead of complicated model architecture designs or multi model dependence, we build a simple method to make language models more efficient by training from common models. 

\textbf{RL vs SFT.} To obtain smaller language models with relatively high performance, Deepseek-R1-distill family~\cite{guo2025deepseek} shows that smaller language models can also get high performance by SFT with the long CoT reasoning answer generated by the strong reasoning LLM. s1 \cite{muennighoff2025s1} shows that SFT on only 1,000 examples suffices to build a competitive reasoning model matching o1-preview and produces a model that lies on the pareto frontier. However, \cite{setlur2025scaling} argues that scaling test-time compute without verification or RL is suboptimal, meaning that distillation samll models do not understand the reasoning process but replicate the solution. We demonstrate that only SFT can either make the model generate meaningless and redundant responses when querying difficult questions or do not reach the best trade-off point between reasoning ability and efficiency. 


\textbf{Length Control in LLMs.} To fulfill different hardware settings and different users' requirements, several recent studies focus on controlling the reasoning or response length, to have the ability either to save the inference cost by limiting the length or to dig out the reasoning potential of language models. s1 \cite{muennighoff2025s1} utilizes a inference strategy called budget forcing on the distilled small model to control the thinking length, while L1 \cite{aggarwal2025l1} empower the model with length control by adding penalty term with different token length targets during RL training on a super powerful 1.5b model. {There are also a few contemporaneous works \cite{xu2025scalable,hou2025thinkprune,yang2025towards}. Compared to these, we focus on and contrast the impact of different strategies (intervention, SFT, RL) on reasoning length control. As a result, we provide temperature scaling as a more natural and effective intervention strategy compared to Budget Forcing. Compared to L1 which uses DeepScaler as their base model,  we provide a stronger understanding of the shortcomings of SFT and demonstrate that RL can unlock token-efficiency even with weaker base models.}


\section{Observations of Supervised Finetuning and Reinforcement Learning Regarding Response Length}
\label{sec:observations}

In this section, we scrutinize a range of models, including Qwen-2.5 family~\cite{yang2024qwen2-a,yang2024qwen2-b}, Deepseek-R1-distill family~\cite{guo2025deepseek}, DeepScaleR~\cite{deepscaler2025} and our RL fine-tuned models. The objective of this section is to explore the benefits and drawbacks of the two post-training finetuning methods, SFT and RL, in terms of their reasoning capability and efficiency. We make four main observations below, which later inform the application of our intervention method in \Cref{sec:observation6} and the design of our \alg framework in \Cref{sec:method}.

\subsection{Observation 1: Supervised fine-tuning introduces redundancy}
To equip more efficient smaller models with reasoning capabilities, open-source models, like Qwen~\cite{yang2024qwen2-a,yang2024qwen2-b}, with SFT, use samples curated by powerful RL-trained long reasoning models such as Deepseek-R1~\cite{guo2025deepseek}.
Recent works indicate that this straightforward distillation approach significantly enhances the reasoning abilities of small models.
Our main observation is that LLMs with SFT sometimes generate excessively long answers, especially for smaller models.
 
\textbf{Experiment setup.}
We use the Qwen-2.5 family~\cite{yang2024qwen2-a,yang2024qwen2-b} as the base models as well as short CoT models. For SFT models, we use the Deepseek-R1-distill family~\cite{guo2025deepseek}, labeled as ``SFT-Deepseek'',  which has been fine-tuned on the Deepseek-R1 curated dataset containing approximately 800k samples of reasoning and non-reasoning data. 
Additionally, we study the S1 family of models~\cite{muennighoff2025s1}, labeled ``SFT-S1'', which has been fine-tuned on S1K, the dataset containing 1k questions paired with reasoning traces. To obtain S1 SFT models of different sizes, we use the officially released S1K SFT models. For models fine-tuned with RL, we train using either standard GRPO or our proposed (described later in \Cref{sec:method}). 
We evaluate on the MATH500 dataset~\cite{lightman2023let}. 

\begin{figure}
\centering
\includegraphics[width=\textwidth]{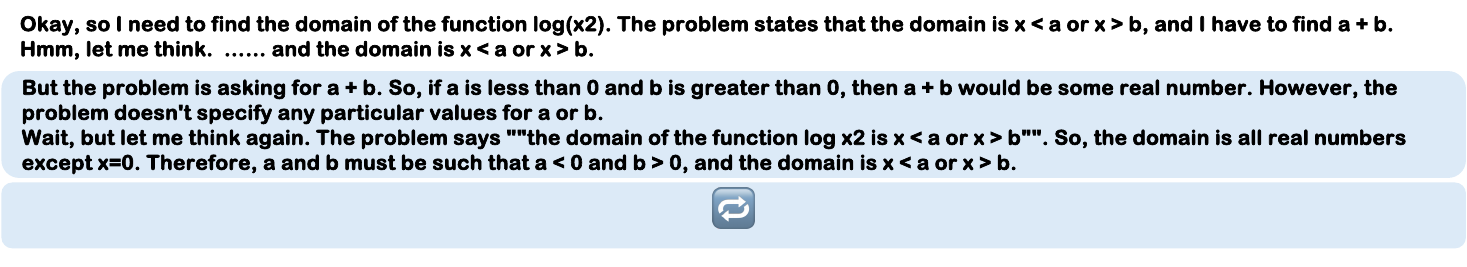}
\caption{Example of wrong and lengthy response, generated by the Deepseek-R1-Distill-Qwen-1.5B model for a Math500 question. The response repeatedly cycles through the paragraph shown in the blue block until it reaches the maximum context length, without providing a final answer.} 
\label{fig:bad_2}
\end{figure}

\textbf{SFT models generate excessively long and repetitive responses.}
First, we investigate the two SFT models with improved performance compared to the Qwen base model. The results of models with 7b parameters are shown in \Cref{tab:model_perfromance}. We also visualize the length distribution in \Cref{fig:rl_sft_length}. In \Cref{fig:rl_sft_length}, the blue bars are the average length of all the responses, the orange bars are the average length of all the correct responses and the green bars show the average length of all the incorrect responses. 
We can see that SFT models generate longer responses compared to the base model and RL fine-tuned models. 
Further, the average length of wrong answers are always longer than correct answers. 
Such responses are undesirable because they are costly, inefficient, and unlikely to be correct.
An example wrong answer is shown in \Cref{fig:bad_2}. 

To understand the prevalence of such repetitive answers, we compute the ``repeat rate'', defined as the proportion of incorrect answers that are excessively long and repetitive among all wrong answers. Repetition is identified by querying GPT-4o-mini~\cite{GPT-4o-mini} with a custom prompt (\Cref{app:repeat_prompt}).
The repeat rate results are presented in~\Cref{fig:rl_sft_repeat}.
We can see that the repeated error rate of SFT models is much higher than that of base model and RL fine-tuned models. We also propose hypothesis why RL natually provides more effective control in \Cref{sec:observation5}.

\subsection{Observation 2: Longer traces for FT causes redundant responses for small models}
\label{sec:repeat_rate}
\textbf{Experiment setup.}
We include the S1 family of models~\cite{muennighoff2025s1}, which has been fine-tuned on S1K or S1K-1.1, two datasets containing 1k questions paired with reasoning traces. These traces were generated using either the Google Gemini Flash Thinking API~\cite{Gemini} (S1K) or Deepseek-R1 (S1K-1.1), which are labeled as ``short'' and ``long'' respectively in \Cref{fig:scale_repeat}. Notably, reasoning traces generated by Deepseek-R1 are of higher quality but are also significantly longer, often including the ``aha'' moment that encourages deeper reasoning. The length distribution of the two kinds of traces are shown in \Cref{app:s1k}. 

\textbf{Longer traces for FT hurt response efficiency.}
Previous research~\cite{muennighoff2025s1} suggests that incorporating difficulty, and high-quality traces can enhance reasoning performance. However, we argue that complex traces can decrease the efficiency of small models and may even adversely affect performance, as they might exceed the maximum context length before completing their reasoning process.
In \Cref{fig:scale_repeat}, we present the repeat rate for different model sizes and trace lengths. The orange bars indicate the results of models trained with shorter Gemini traces, while the green bars show models trained with significantly longer Deepseek traces. We observe that smaller models tend to exhibit higher repeat rates. Additionally, using long reasoning traces for fine-tuning increases this rate. Among all incorrect responses, over 60\% of failures are due to repetition in the 1.5B parameter model that was fine-tuned with supervision using long Deepseek traces. We provide an example of a repeating incorrect answer in \Cref{app:bad_example}, \Cref{fig:bad_1}. In this example, the model generates the word ``wait'' 577 times, continuously repeats the same sentences, and fails to provide a final answer before reaching the maximum context length, with one paragraph repeating 180 times. Even worse, when the model finally escapes this repetitive cycle, it often falls back into it. We argue that fine-tuning with high-quality, longer traces severely hurts response efficiency. 

\vspace{-5pt}
\subsection{Observation 3: Controlling context length trades off performance for efficiency}
\textbf{Experiment setup.}
We experiment with three different sizes of state-of-the-art SFT models, drawn from the Deepseek-R1-distill models~\cite{guo2025deepseek} on the MATH500 dataset~\cite{lightman2023let} with a state-of-the-art S1 ``budget forcing''  method~\cite{muennighoff2025s1} that caps the thinking trajectory length using test-time compute. ``Auto'' denotes responses generated without any extra constraints on response length.

The results are shown in~\Cref{fig:ds_qwen-math-tokens_vs_accuracy(Mainbody)} for different model sizes. The x-axis denotes the average context length, and the y-axis shows the corresponding accuracy. Dots on the upper left of the figure have a better efficiency-performance trade-off. ``Budget forcing'' achieves shorter response lengths with comparable or even better performance compared to responses generated without constraints (``Auto'') across all three sizes. The improvement of budget forcing methods is most significant on the smaller model, showing an almost 50\% length decrease on Deepseek-R1-distill-1.5b (blue). 
However, without training, such test-time length control strategies sometimes fail to find an optimal trade-off (results shown in the Appendix in \Cref{fig:three-benchmark-token_accuracy}). These results suggest significant room for improvement in achieving an efficient performance trade-off, a key motivation for our \alg.

\subsection{Observation 4: Test-time compute cannot precisely control response length}
\textbf{Experiment setup.}
We conducted experiments using various test-time length control strategies, including the S1 ``budget forcing'' as in the previous subsection, and ``exact control''. These methods force the thinking trajectory to be of fixed length through a collection of strategies: suppressing the generation of the end-of-thinking token delimiter to enforce a minimum response length, appending a ``Wait'' to the model's current reasoning trace to encourage reflection on its generation \cite{aggarwal2025l1}), and adding the prompt ``Think for up to $n$ tokens'' after the question. These experiments were performed on three different datasets: MATH500 \cite{lightman2023let}, GPQA \cite{rein2024gpqa}, and AIME24.

\begin{figure}
\centering
\begin{subfigure}[b]{0.32\textwidth}
\includegraphics[width=\textwidth]{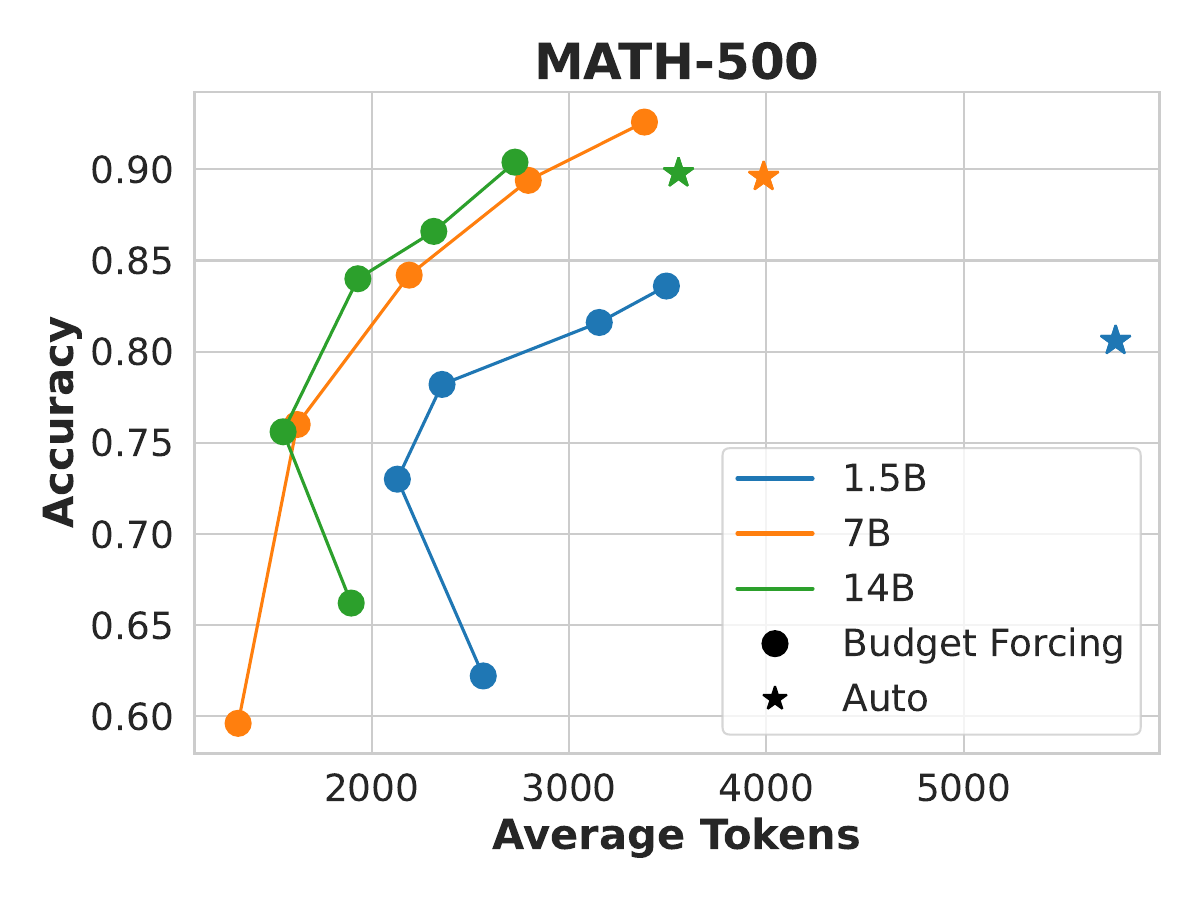}
\caption{\small{Efficiency-performance trade-off}}\label{fig:ds_qwen-math-tokens_vs_accuracy(Mainbody)}
\end{subfigure}
~
\begin{subfigure}[b]{0.32\textwidth}
\includegraphics[width=\textwidth]{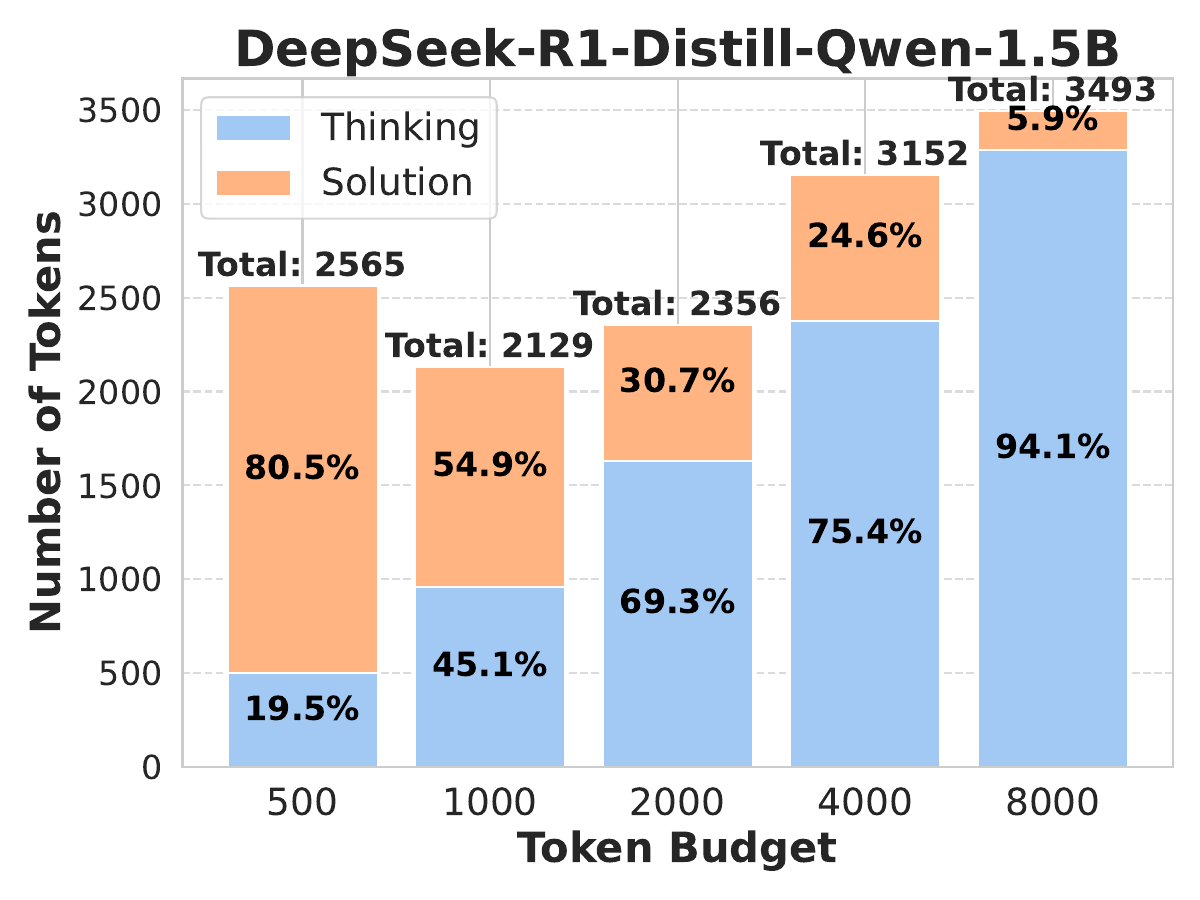}
\caption{\small{Thinking:solution token ratio of Deepseek-R1-distill-1.5b}}\label{fig:ds-qwen1p5b-force-math-len_dist}
\end{subfigure}
~
\begin{subfigure}[b]{0.32\textwidth}
\includegraphics[width=\textwidth]{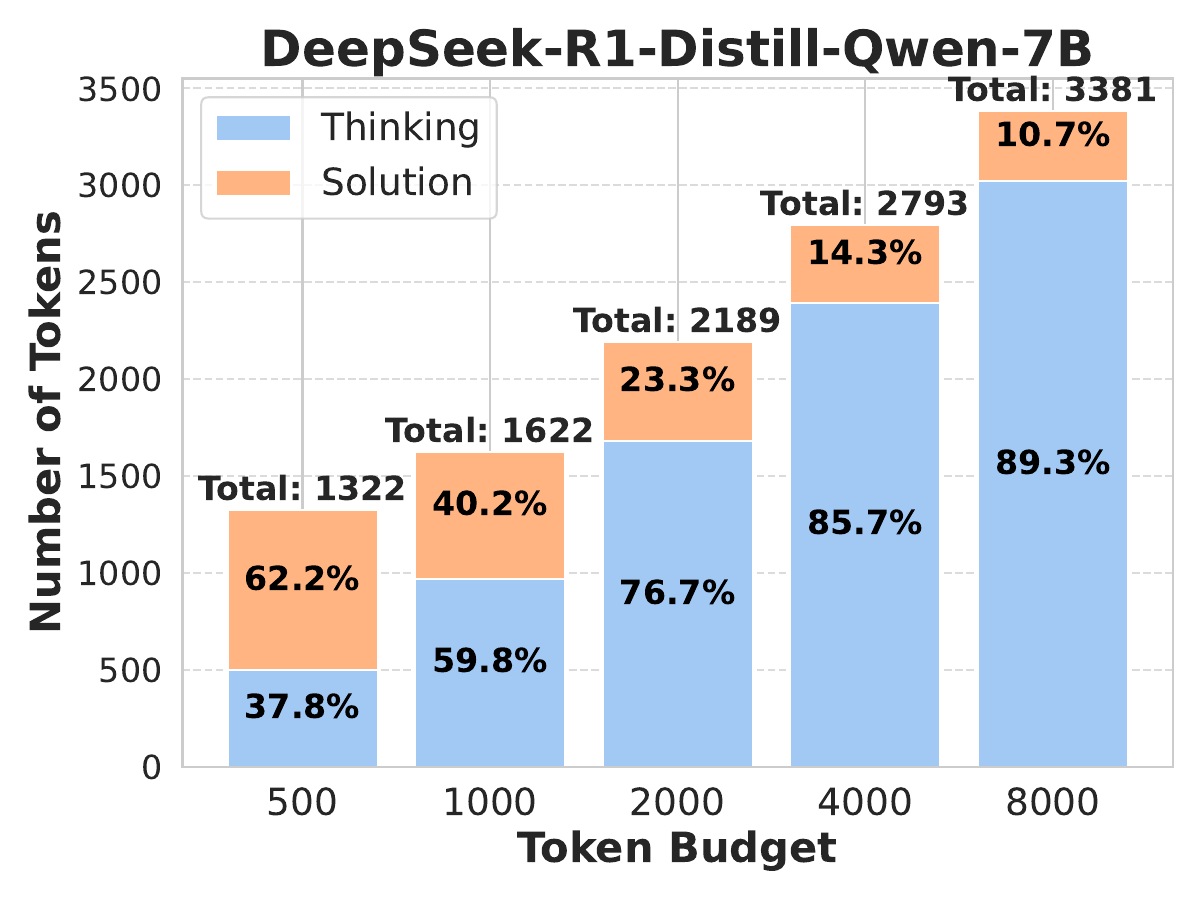}
\caption{\small{Thinking:solution token ratio of Deepseek-R1-distill-7B}}\label{fig:ds-qwen7b-force-math-len_dist}
\end{subfigure}
\caption{Test time compute strategies can trade off performance for efficiency, but cannot precisely control response length, in part due to many tokens spent on the final solution.} 
\label{fig: ds-qwen-math}
\end{figure}

The results on the MATH500 dataset using budget forcing are shown in \Cref{fig: ds-qwen-math}. The budget forcing method cannot precisely control the total response length, as demonstrated in \Cref{fig:ds_qwen-math-tokens_vs_accuracy(Mainbody)}. Despite a maximum 500-token constraint for the thinking trajectory, the final response often extends to about 2000 tokens or more (first green/blue dot), which is longer than the responses generated with a maximum 1000-token constraint. 
This is because forcing a shorter thinking trajectory sometimes leads to a significantly longer final solution.
To illustrate this, we plot the ratio of tokens spent during thinking vs the final solution in \Cref{fig:ds-qwen1p5b-force-math-len_dist,fig:ds-qwen7b-force-math-len_dist}, for different token budgets. These length control methods performed poorly. Additional results for all length control strategies and the datasets are in \Cref{fig:three-benchmark-token_accuracy,fig:three-benchmark-length-dist} in the Appendix.

\subsection{Observation 5: SFT introduces redundancy due to equal treatment of end token, while RL-based models can learn when to stop} \label{sec:observation5}
\begin{figure}
\centering
\includegraphics[width=0.4\textwidth]{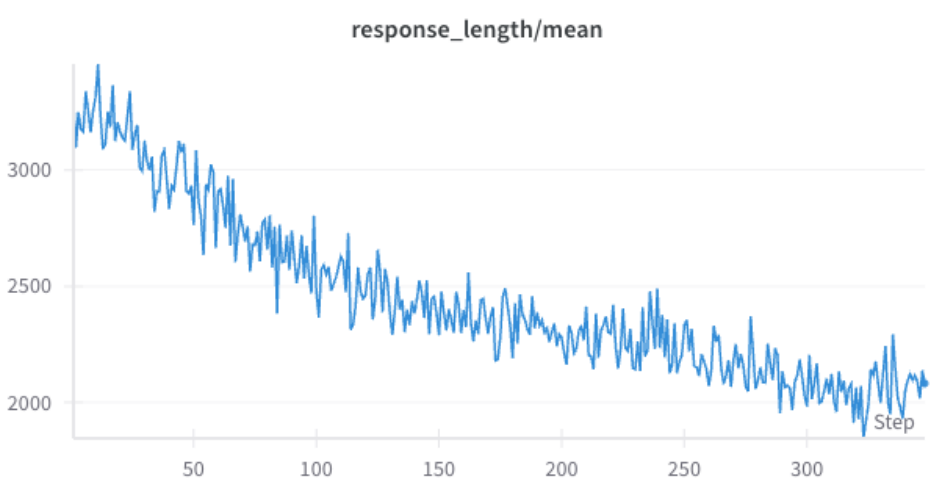}
\caption{Changes in response length during reinforcement training without length penalty.} \label{fig:rl_reduce_length}
\end{figure}

In SFT, the model is trained to predict the next token based on ground truth sequences, with each token, including special tokens such as end-of-sequence (EOS), $<$think$>$, $<$answer$>$, treated equally during optimization. This method can lead to redundancy in reasoning tasks for several reasons:
\begin{itemize}[leftmargin=*]
	\item \textbf{Lack of Explicit Stopping Signal:} Since the EOS token is treated like other tokens, the model is not explicitly encouraged to optimize for brevity. Instead, it learns to imitate long reasoning traces, even when shorter reasoning suffices.
	\item \textbf{Bias:} During training, the model only sees EOS tokens once at the end of the reasoning traces. 
	\item \textbf{Overfitting to Long Traces:} If the training data contains long, detailed reasoning sequences, SFT models tend to memorize these structures instead of learning when a concise answer would be sufficient.
\end{itemize}
RL can provide more effective control. Unlike SFT, RL-based models can learn when to stop reasoning. RL treats stopping as a decision-making problem instead of treating it equally with other tokens. The model can learn when to terminate reasoning to maximize rewards. This is evidenced by decreases in response length during training, as shown in \Cref{fig:rl_reduce_length}.

In RL, we can also define a reward function that explicitly balances reasoning quality and efficiency. Unlike SFT, where the model passively learns from long training sequences, RL allows us to actively penalize excessive length, while rewarding correct and concise answers, by adding a length penalty. 

\section{Proposed Method: Post-hoc Temperature Scaling} 
\label{sec:observation6}
Motivated by the bad example shown in \Cref{fig:bad_2}, we check the next-token probabilities at the end of each block loop and find that in 87.5\% of the cases, the end-of-sequence (EOS) token ranks among the top 5 candidates, despite not being selected. So, we introduce post-hoc temperature scaling to increase the probability of the end-of-sequence (EOS) token, to reduce the response length of the SFT-distilled models. The temperature scaling does not change the internal reasoning process of the model. Temperature scaling is a fundamental method for controlling model behavior, influencing aspects such as stochasticity of generative LLMs, calibration and imbalanced data, as highlighted in several studies~\cite{zhang2024selective,menon2020long,li2021autobalance,zhang2024class}. $\mathbf{l} \in \mathbb{R}^V$ is the original logits output by the model, where $V$ is the vocabulary size, and the EOS token has index $i_{\text{eos}}$, we modify the EOS logit as $l_{i_{\text{eos}}}{\prime} = \frac{z_{i_{\text{eos}}}}{T}, \quad T < 1$. By increasing the likelihood of EOS, the model is more likely to terminate its generation earlier, thus producing shorter outputs. Empirically, we find that this simple adjustment effectively reduces redundancy, especially repeating, while preserving the core reasoning steps and answer accuracy. The result is shown in \Cref{tab:ts}.

\begin{table}[h]
\scriptsize
\centering
\begin{tabular}{c|c|c|c|c|c|c}
    \hline
      &  \multicolumn{2}{|c|}{Base} & \multicolumn{2}{|c|}{BF} & \multicolumn{2}{|c}{TS} \\ \hline
      &  acc & length  & acc & length  &  acc & length  \\ \hline
    SFT-S1-7b & 77.00 & 4842.68 & 77.17 & 3591.21 & 77.01 & 1983.93  \\ \hline
    SFT-DeepSeek-1.5b & 80.60 & 5869.60 & 81.03 & 3162.49 & 81.09 &  2615.66 \\ \hline
    SFT-DeepSeek-7b & 88.17 & 4078.26 & 88.03 & 2839.95 & 88.95 &  2547.42 \\ \hline
\end{tabular}
\caption{Results of post-hoc temperature scaling. Since budget forcing (BF) requires a predefined target length, we sweep across 500, 1k, 2k, and 4k tokens, and select the config that achieves better performance than the base model while producing the shortest output as the BF baseline.
}
\label{tab:ts}
\end{table}

Based on the above observations, we ask two research questions: 
\begin{itemize}[leftmargin=*]
    \item \textbf{QS1:} Given a small reasoning model, can we obtain a sweet spot between accuracy and response length, which means a model generates shorter responses without hurting the reasoning ability? 
    \item \textbf{QS2:} Given a small reasoning model, can we incorporate the model with the ability to generate responses in different length levels by following the users' prompts? 
\end{itemize}

\section{Proposed Method: Reinforcement Learning with Length Penalty} \label{sec:method}



 To tackle the above two research questions, we propose penalized reward function, conditioned on the response length.
The proposed reward functions can integrate with the Group Relative Policy Optimization (GRPO) algorithm, a popular RL algorithm for efficient training nowadays. 
We denote the response length penalty function as $\zeta(L)$, where $L$ is the length of the response.
The final reward function is $r=\hat{r}-\zeta(L)$, where $\hat{r}$ is the original reward function (e.g., accuracy reward, format reward, etc.).

A unique aspect of our approach is that different parameter settings can be indirectly controlled by the end user through a special prompt, for example ``[Response Length: LONG] Provide a detailed step-by-step solution.''
In other words, the model is trained with the prompt and its corresponding penalty function.
The model thus learns to pair the special prompt with the trajectories associated with that length penalty.
Then during inference, the model should automatically produce responses that match that length penalty.

\begin{wrapfigure}{r}{0.4\textwidth}
\centering
\includegraphics[width=0.4\textwidth]{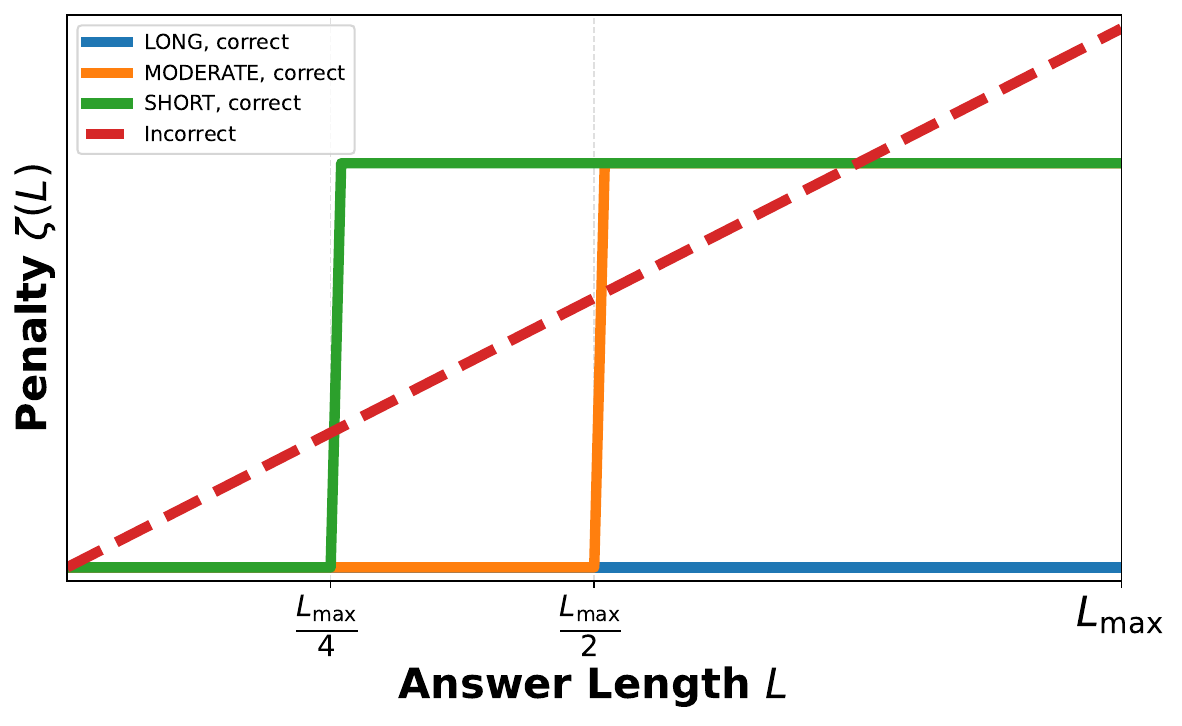}
\caption{Length Penalty function.} \label{fig:lp_function}
\end{wrapfigure}

\textbf{``Sweet Spot'' of Length Penalty.} To find the ``sweet spot'' of response efficiency without hurting the reasoning performance, we design the penalty function as $\eta(L) = \alpha \frac{L}{L_{\text{max}}}$, where $\alpha$ is a hyperparameter for the strength of the length penalty, and $L_{\text{max}}$ is the maximum length of the model's response. The $\alpha$ value should be set to a relatively small value to ensure that the penalty term $\eta(L)$ does not perturb the final reward $r$ too much. Note that in this setting, the reward function $r=\hat{r}-\eta(L)$ correspondingly. 

\textbf{Length Penalty for Multi-level Length Control.} To address the problem of multi-level LP control, we define three different LP levels: short, moderate and long.
Intuitively, the penalty function should be designed to penalize correct answers from being too long.
We also set a special penalty $\eta{L}$ for wrong answers, which is same as "Sweet Spot" experiments, since we don't want LLMs to waste tokens when it cannot solve the problem with the current model size. Formally, the length penalty is defined as:

{\scriptsize{
\begin{align}
    \zeta(L)=
    \begin{cases}
    0, & \text{if } c=1\text{ and (LONG or (MODERATE and }L\leq \frac{L_{\text{max}}}{2}\text{) or (SHORT and }L\leq \frac{L_{\text{max}}}{4}\text{)}, \\
    \beta,  & \text{if } c=1\text{ and (MODERATE and }L>\frac{L_{\text{max}}}{2}\text{ or (SHORT and }L>\frac{L_{\text{max}}}{4}\text{)}, \\
    \eta(L),  & \text{if }c=0 
    \end{cases}
    \label{eq:length_penalty_multi_level}
\end{align}
}}

where LONG, MODERATE, and SHORT correspond to the desired LP level (paired with the corresponding prompts), $L>\frac{L}{2}$ and $L>\frac{L}{4}$ are the max length thrshold corresponds to MODERATE, and SHORT prompts, $c \in \{0,1\}$ means the answer is incorrect/correct, and $\beta$ represents the penalty for correct answers whose length is greater than a threshold. The function are shown in \Cref{fig:lp_function}.
\section{Experiments}
\label{sec:experiments}

\subsection{Sweet Spot through Reinforcement Learning + Length Penalty}
\textbf{Setup.}
In this experiment, we use 3 base models as shown in \Cref{tab:sweet_spot}, which contains models of different types and sizes (1.5b and 7b). Specifically, Qwen2.5-Math-1.5B, Qwen2.5-Math-7B \cite{yang2024qwen2-a,yang2024qwen2-b} are short CoT reasoning models, while DeepSeek-R1-Distill-Qwen-7B is a long CoT reasoning model. We warm up the models by training them on the MATH training dataset with GRPO, where $\hat{r}=0.9\cdot r_\text{acc}+0.1\cdot r_{\text{format}}$, $r_\text{acc}$, $r_{\text{format}}$ are the reward for accuracy and format, and then train with length penalty as defined in \textbf{``Sweet Spot" of Length Penalty}, where $\alpha \in \{0.0, 0.1\}$ for RL without length penalty and RL with length penalty correspondingly. After training finishes, we test the models' performance on the MATH500 dataset. For all RL training, we set the response max length to 4096 because this is max length that can be trained with our resource, and 4096 is generally long enough to solve problems in MATH dataset.


\textbf{Results.} In \Cref{tab:sweet_spot}, we report the accuracy and token length for the base model and for different settings of the length penalty.
We can see in the second column ``RL'' that RL training directly within 4096 response length limitation can shorten the response length directly, while at the same time generally improving or maitaining the accuracy of SFT models compared to the Base model.
From the third ``RL + Length Penalty'' column, we can see that by applying length penalty, we can further shorten the response length while preserving the accuracy. With the comparison between the results of ``RL'' and ``RL + Length Penalty'', we can see that while RL without length penalty can already gain better response efficiency, a better ``Sweet Spot'' can be obtained when training with length penalty. 

\begin{table}[h]
\scriptsize
\centering
\begin{tabular}{c|c|c|c|c|c|c}
    \hline
      &  \multicolumn{2}{|c|}{Base} & \multicolumn{2}{|c|}{RL} & \multicolumn{2}{|c}{RL+Length Penalty} \\ \hline
      &  acc & length & acc & length & acc & length  \\ \hline
    Qwen2.5-Math-1.5B(4k) & 23.4 & 1976.874 & 68.2 & 685.896 & 71.4 & 495.842  \\ \hline
    Qwen2.5-Math-7B(4k) & 57.2 & 1109.13 & 70 & 671.58 & 71.2 & 404.962  \\ \hline
    DeepSeek-R1-Distill-Qwen-1.5B(32768) & 80.6 & 5769.596 & 77.2 & 1929.208 & 80.4 & 1104.748  \\ \hline
\end{tabular}
\caption{We investigate a range of models (size, short/long CoT, SFT/RL) and show that different values of the length penalty can produce shorter responses while maintaining good accuracy. 
}
\label{tab:sweet_spot}
\end{table}



\subsection{Length Control through Reinforcement Learning + Multi-Level Length Penalty}
\begin{figure}
\centering
\begin{subfigure}[b]{0.45\textwidth}
\includegraphics[width=\textwidth]{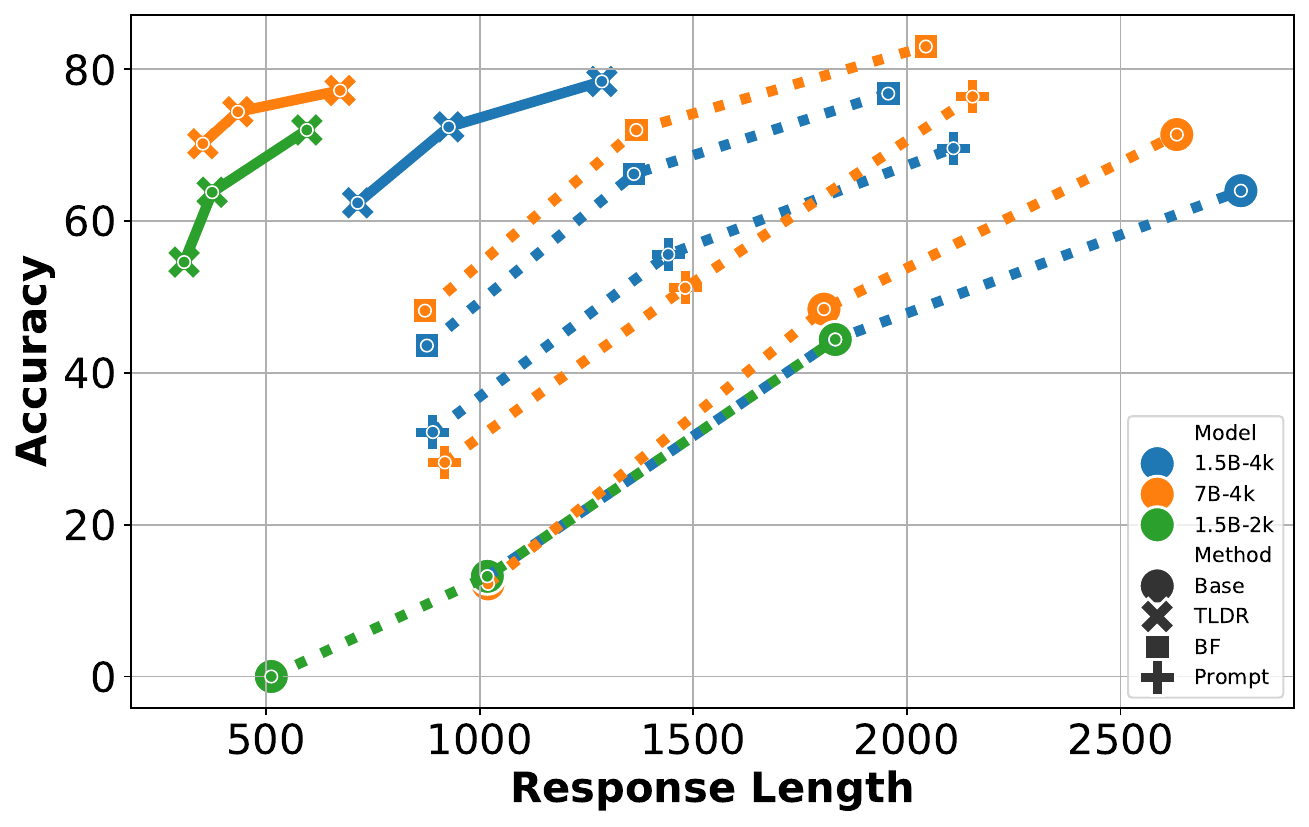}
\vspace{-15pt}
\caption{\small{MATH500}}\label{fig:3_level_acc_length_MATH}
\end{subfigure}
~
\begin{subfigure}[b]{0.45\textwidth}
\includegraphics[width=\textwidth]{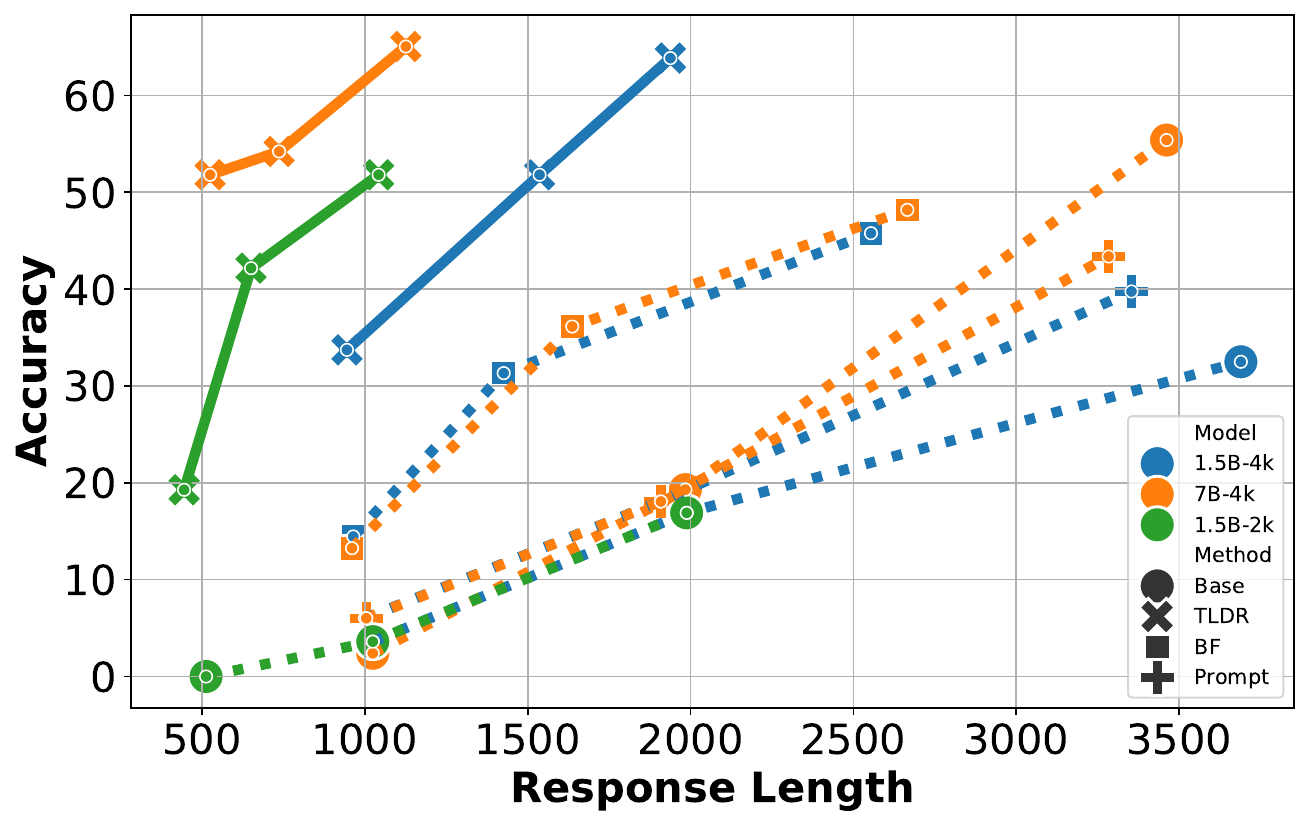}
\vspace{-15pt}
\caption{\small{AMC}}\label{fig:3_level_acc_length_AMC}
\end{subfigure}
~\\
\begin{subfigure}[b]{0.45\textwidth}
\includegraphics[width=\textwidth]{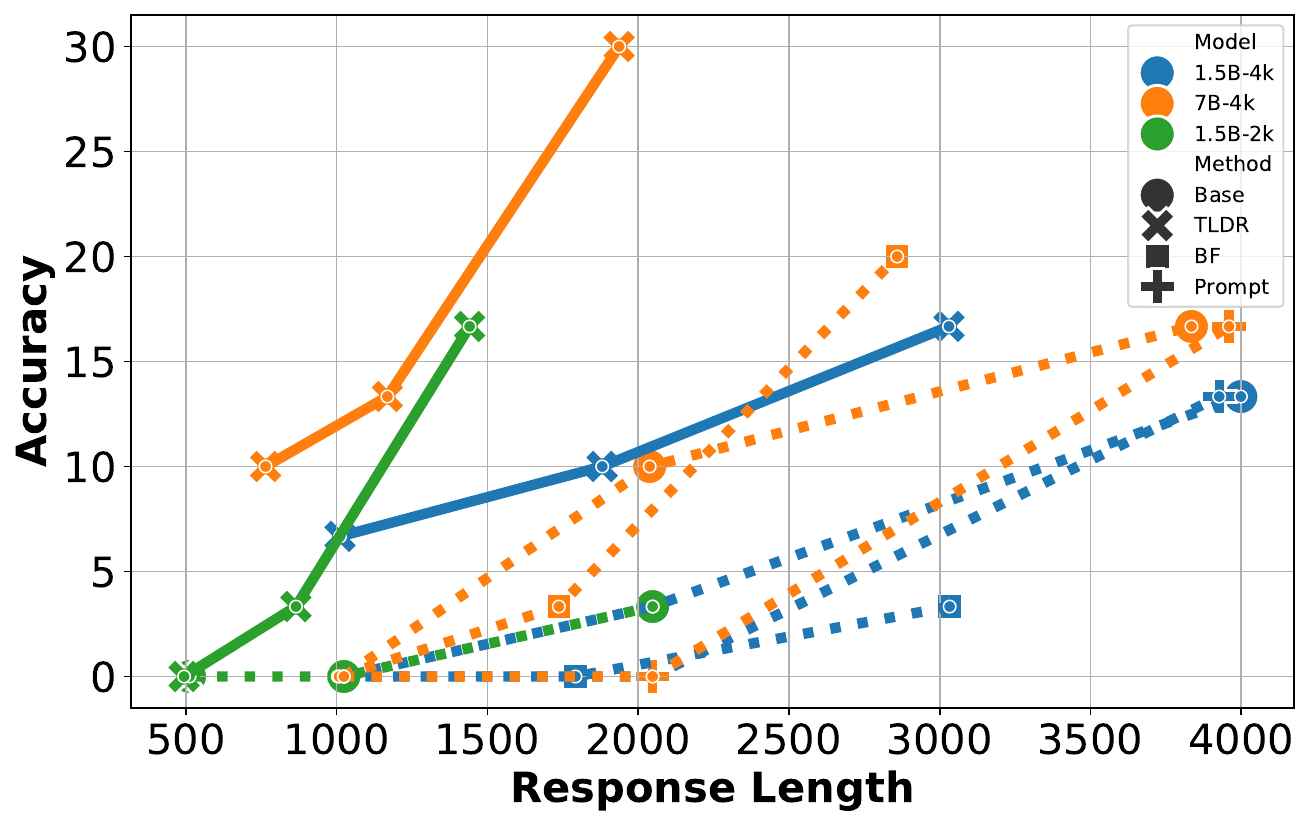}
\vspace{-15pt}
\caption{\small{AIME24}}\label{fig:3_level_acc_length_AIME}
\end{subfigure}
~
\begin{subfigure}[b]{0.45\textwidth}
\includegraphics[width=\textwidth]{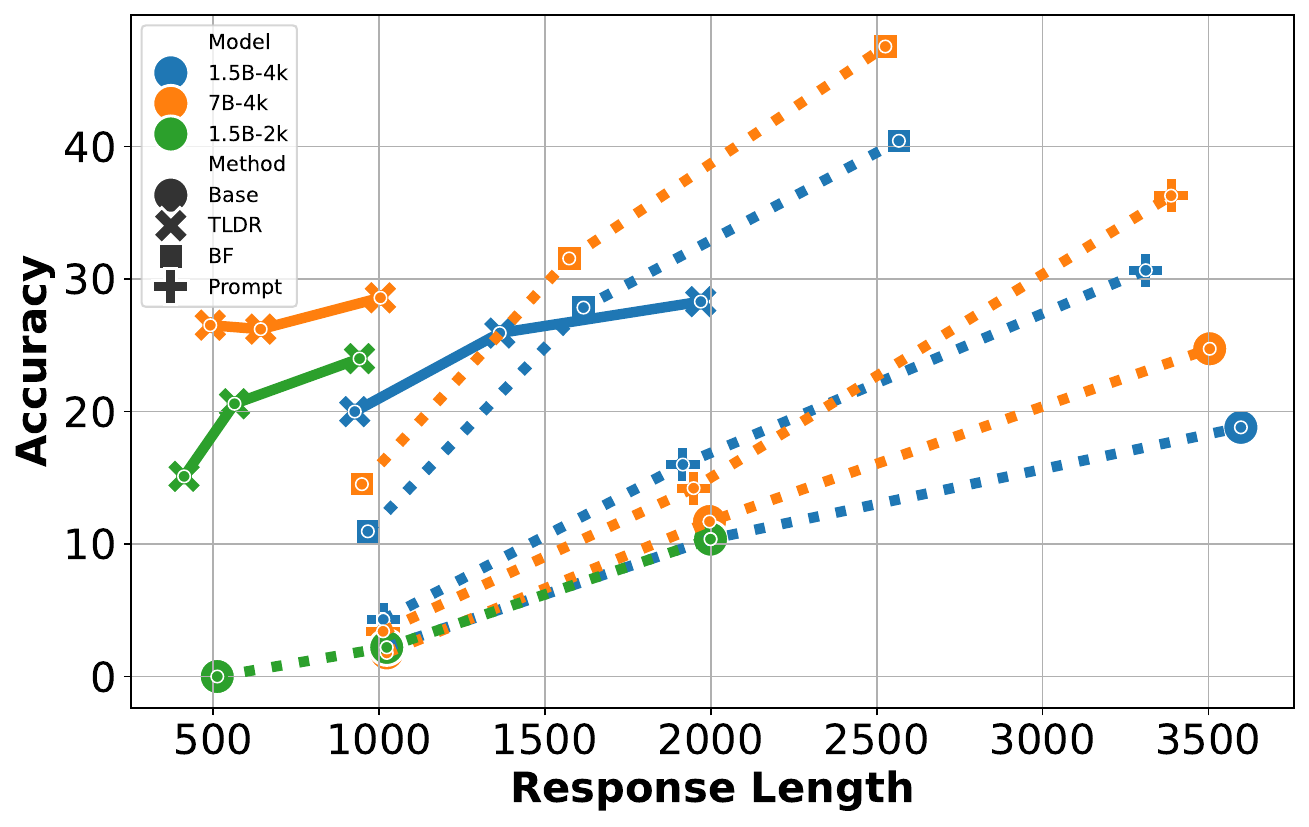}
\vspace{-15pt}
\caption{\small{OlympiadBench}}\label{fig:3_level_acc_length_oly}
\end{subfigure}
\caption{Performance of \alg under varying model sizes and token budget. \alg outperforms length-controlled baselines (S1) both in performance and token efficiency. }
\label{fig:3_level_acc_length}
\vspace{-20pt}
\end{figure}
\textbf{Setup.}
In these experiments, we use two base models: DeepSeek-R1-Distill-Qwen-1.5B and DeepSeek-R1-Distill-Qwen-7B. The training dataset is a combination of the training set of MATH, AIME, AMC, STILL, OlympiadBench, which is same as the training set of DeepScaleR-1.5B-Preview. We first warm up the models by training them to the convergence in the sense of the original reward $\hat{r}$, then we do the training with multi-level response length penalty as defined in \Cref{eq:length_penalty_multi_level}, where $\gamma=0.1, \beta=0.3$ in our setting. After training finishes, we test the models’ performance on the MATH500 dataset and the test sets of AMC, AIME24 and OlympiadBench datasets. For DeepSeek-R1-Distill-Qwen-1.5B, we experiment with $L_{\text{max}}=2048, 4096$, and for DeepSeek-R1-Distill-Qwen-7B, we experiment with $L_{\text{max}}=4096$. The prompts for low/medium/high length penalty are ``[Response Length: LONG] Provide a detailed step-by-step solution.'', ``[Response Length: MODERATE] Provide a concise but clear solution.'', ``[Response Length: SHORT] Provide only the essential steps.'' respectively. These three levels corresponds to the $\frac{1}{4},\frac{1}{2}$ length and the full length of the max response length.

\begin{figure}
\centering
\begin{subfigure}[b]{0.3\textwidth}
\includegraphics[width=\textwidth]{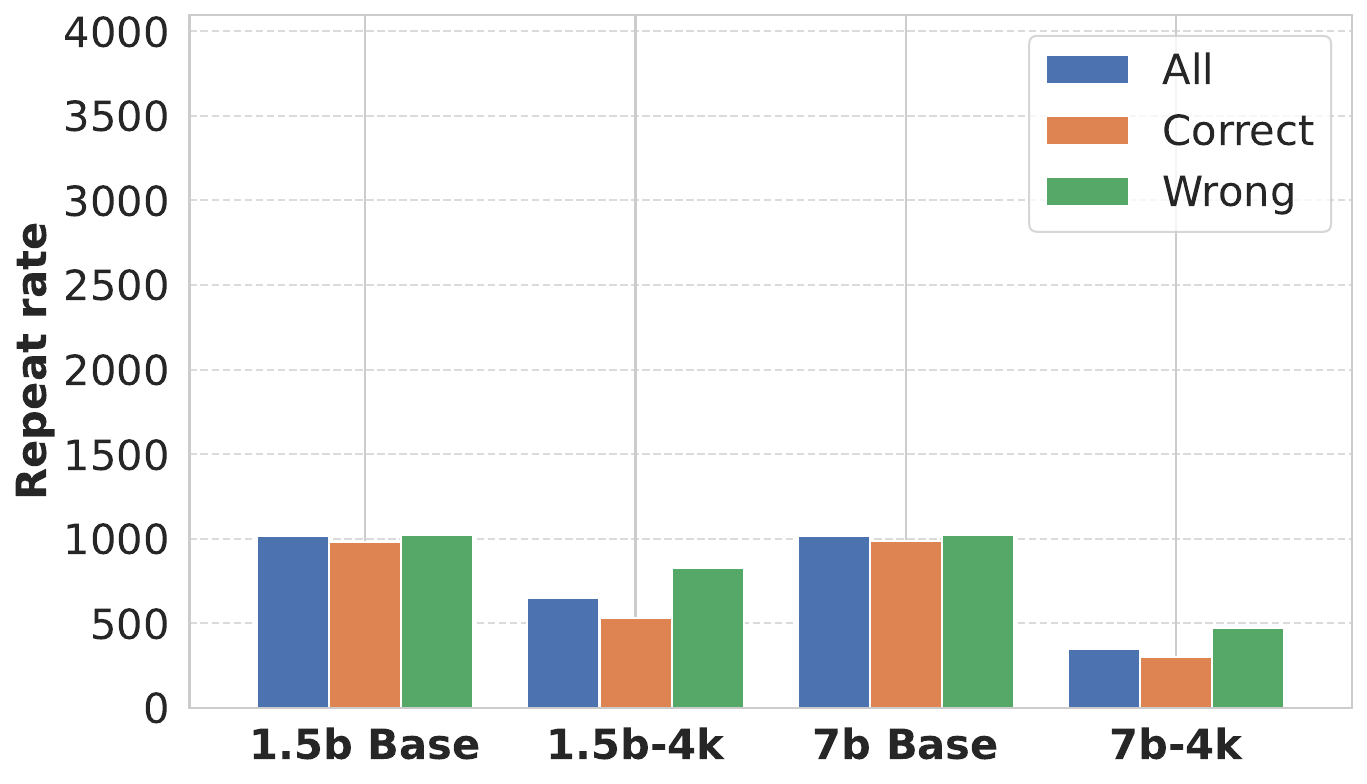}
\vspace{-15pt}
\caption{\small{SHORT}}\label{fig:3_level_length_short}
\end{subfigure}
~
\begin{subfigure}[b]{0.3\textwidth}
\includegraphics[width=\textwidth]{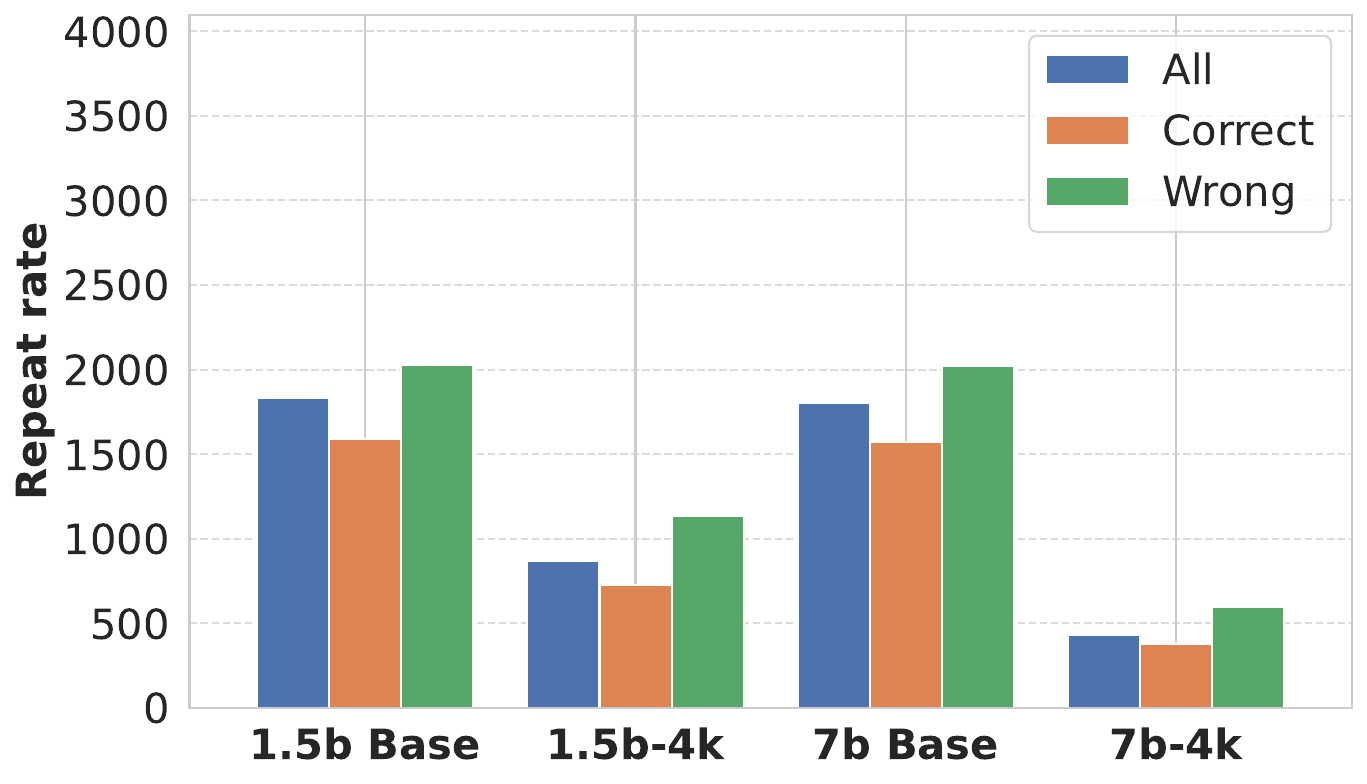}
\vspace{-15pt}
\caption{\small{MODERATE}}\label{fig:3_level_length_moderate}
\end{subfigure}
\begin{subfigure}[b]{0.3\textwidth}
\includegraphics[width=\textwidth]{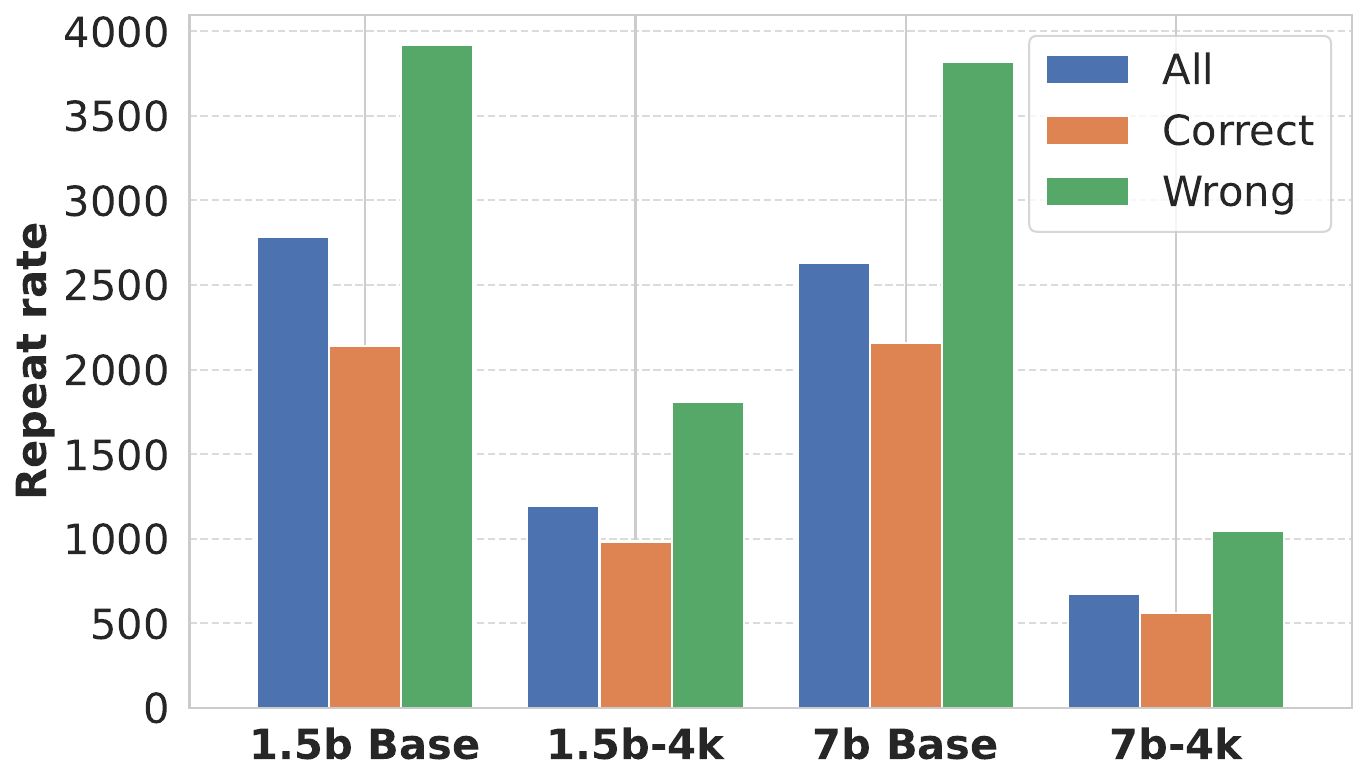}
\vspace{-15pt}
\caption{\small{LONG}}\label{fig:3_level_length_long}
\end{subfigure}
\caption{The response length of models before and after trained with \alg in different length control levels (SHORT/MODERATE/HIGH)). After training \alg, both of the correct and wrong responses' length decreases.}
\label{fig: 3_level_length}
\end{figure}

\textbf{Results.}
\Cref{fig:3_level_acc_length} shows the accuracy and token length for different models and settings of the length levels.
From the results, we can see that the base DeepSeek-R1-Distill models tend to struggle with these strong reasoning tasks. This issue can be more severe when the response length is limited, which is shown by the accuracy gap between \Cref{fig:3_level_acc_length_MATH,fig:3_level_acc_length_AIME} for the base model, and the results reported in DeepSeek-R1~\cite{guo2025deepseek}.
From \Cref{fig:3_level_acc_length}, we can see that after training the distilled models by GRPO with TLDR, we not only achieve much better accuracy on all four reasoning datasets by at most 50\%, but shorten the response length by at most 50\%. 
In particular, the lowest length penalty (LONG) generally results in the best accuracy albeit with slightly more tokens. We beat the base model, prompt control method and even  the SOTA s1 \cite{muennighoff2025s1} both in accuracy and inference efficiency. In \Cref{fig: 3_level_length}, we show that both correct response length and wrong response length are shortened after training with \alg, and the shortened ratio grows with the response length limitation. This means that \alg can incorporate small language models both the ability to reduce redundancy during reasoning but also stop early when the problem is too hard for it. The detailed experiment results are at \Cref{tab:multi-level_length_control}.

\section{Conclusion}
\label{sec:conclusions}
In conclusion, we propose a novel approach to improving token efficiency in reasoning tasks, addressing the inefficiencies of long reasoning traces generated by language models. Our work demonstrates that supervised fine-tuning (SFT) alone is insufficient to avoid redundant or repetitive outputs, especially for smaller models. To overcome this, we introduce TLDR, a length-penalized variation of Group Relative Policy Optimization (GRPO), which effectively reduces response length while maintaining accuracy. The flexibility of TLDR, through its multi-level penalty system, allows for dynamic control of response length, making it adaptable to different reasoning scenarios. Our experiments across four reasoning benchmarks show that TLDR achieves a significant improvement in token efficiency—around 50\%—without a noticeable loss in accuracy. Furthermore, TLDR’s performance compares favorably to existing methods, offering a superior trade-off between efficiency and accuracy. This work provides valuable insights into optimizing the computational cost of reasoning models, and the proposed method is a practical step toward more efficient and scalable reasoning systems. 
\vspace{-10pt}
\section*{Acknowledgements}\vspace{-10pt}

This work is supported by the National Science Foundation grants CCF-2046816, CCF-2403075, CCF-2212426, the Office of Naval Research grant N000142412289, and an Adobe Data Science Research Award. The computational aspects of the research is generously supported by computational resources provided by the Amazon Research Award on Foundation Model Development. 

\bibliography{colm2025_conference}
\bibliographystyle{plain}
\newpage
\appendix
\section{Model performance} 
\label{app:model_perfromance}
\Cref{tab:model_perfromance} shows the performance of different models on three reasoning datasets, where \text{Qwen2.5-\{\}b-Instruct\_s1K} models are \text{Qwen2.5-\{\}b-Instruct} supervised finetuned with s1K dataset, DeepSeek-R1-Distill-Qwen-\{\}B models are distilled models officially released by DeepSeek, s1.1-3B models are models released by the s1 team. From \Cref{tab:model_perfromance}, we can see that distillation with a small amount of high quality curated long CoT data can have comparable performance when the model size is larger (7B and 14B) and the task is not very difficult (MATH and GPQA), but it's still not close to the distilled models with a large amount of long when the model is small (1.5B, 3B, 7B) or the task is reasoning intensive (AIME). 

\begin{table}[h]
\scriptsize
\centering
\begin{tabular}{|c|c|c|c|c|c|c|c|}
    \hline
    \multicolumn{2}{|c|}{Model} & \multicolumn{2}{|c|}{MATH Accuracy} & \multicolumn{2}{|c|}{GPQA Accuracy} & \multicolumn{2}{|c|}{AIME Accuracy} \\ \hline
    \multicolumn{2}{|c|}{Qwen2.5-1.5b-Instruct\_s1K} & \multicolumn{2}{|c|}{0.426000} & \multicolumn{2}{|c|}{0.176768} & \multicolumn{2}{|c|}{0.066667} \\ \hline
    \multicolumn{2}{|c|}{Qwen2.5-3b-Instruct\_s1K} & \multicolumn{2}{|c|}{0.579158} & \multicolumn{2}{|c|}{0.247475} & \multicolumn{2}{|c|}{0.066667} \\ \hline
    \multicolumn{2}{|c|}{Qwen2.5-7b-Instruct\_s1K} & \multicolumn{2}{|c|}{0.740000} & \multicolumn{2}{|c|}{0.166667} & \multicolumn{2}{|c|}{0.100000} \\ \hline
    \multicolumn{2}{|c|}{Qwen2.5-14b-Instruct\_s1K} & \multicolumn{2}{|c|}{0.794000} & \multicolumn{2}{|c|}{0.414141} & \multicolumn{2}{|c|}{0.233333} \\ \hline
    \multicolumn{2}{|c|}{Qwen2.5-32b-Instruct\_s1K} & \multicolumn{2}{|c|}{0.870000} & \multicolumn{2}{|c|}{0.601010} & \multicolumn{2}{|c|}{0.366667} \\ \hline
    \multicolumn{2}{|c|}{DeepSeek-R1-Distill-Qwen-1.5B} & \multicolumn{2}{|c|}{0.818000} & \multicolumn{2}{|c|}{0.404040} & \multicolumn{2}{|c|}{0.400000} \\ \hline
    \multicolumn{2}{|c|}{DeepSeek-R1-Distill-Qwen-7B} & \multicolumn{2}{|c|}{0.881764} & \multicolumn{2}{|c|}{0.515152} & \multicolumn{2}{|c|}{0.366667} \\ \hline
    \multicolumn{2}{|c|}{DeepSeek-R1-Distill-Llama-8B} & \multicolumn{2}{|c|}{0.856000} & \multicolumn{2}{|c|}{0.510101} & \multicolumn{2}{|c|}{0.533333} \\ \hline
    \multicolumn{2}{|c|}{DeepSeek-R1-Distill-Qwen-14B} & \multicolumn{2}{|c|}{0.903808} & \multicolumn{2}{|c|}{0.558376} & \multicolumn{2}{|c|}{0.586207} \\ \hline
    \multicolumn{2}{|c|}{DeepSeek-R1-Distill-Qwen-32B} & \multicolumn{2}{|c|}{0.914000} & \multicolumn{2}{|c|}{0.595960} & \multicolumn{2}{|c|}{0.833333} \\ \hline
    \multicolumn{2}{|c|}{s1.1-3B} & \multicolumn{2}{|c|}{0.630000} & \multicolumn{2}{|c|}{0.287879} & \multicolumn{2}{|c|}{0.133333} \\ \hline
    \multicolumn{2}{|c|}{s1.1-7B} & \multicolumn{2}{|c|}{0.770000} & \multicolumn{2}{|c|}{0.393939} & \multicolumn{2}{|c|}{0.266667} \\ \hline
    \multicolumn{2}{|c|}{s1.1-14B} & \multicolumn{2}{|c|}{0.840000} & \multicolumn{2}{|c|}{0.575758} & \multicolumn{2}{|c|}{0.300000} \\ \hline
    \multicolumn{2}{|c|}{s1.1-32B} & \multicolumn{2}{|c|}{0.896000} & \multicolumn{2}{|c|}{0.621212} & \multicolumn{2}{|c|}{0.433333} \\ \hline
\end{tabular}
\caption{Model Performance}
\label{tab:model_perfromance}
\end{table}
\section{Model length distribution}\label{app:model_length}
\begin{figure}
\centering
\begin{subfigure}[b]{0.31\textwidth}
\includegraphics[width=\textwidth]{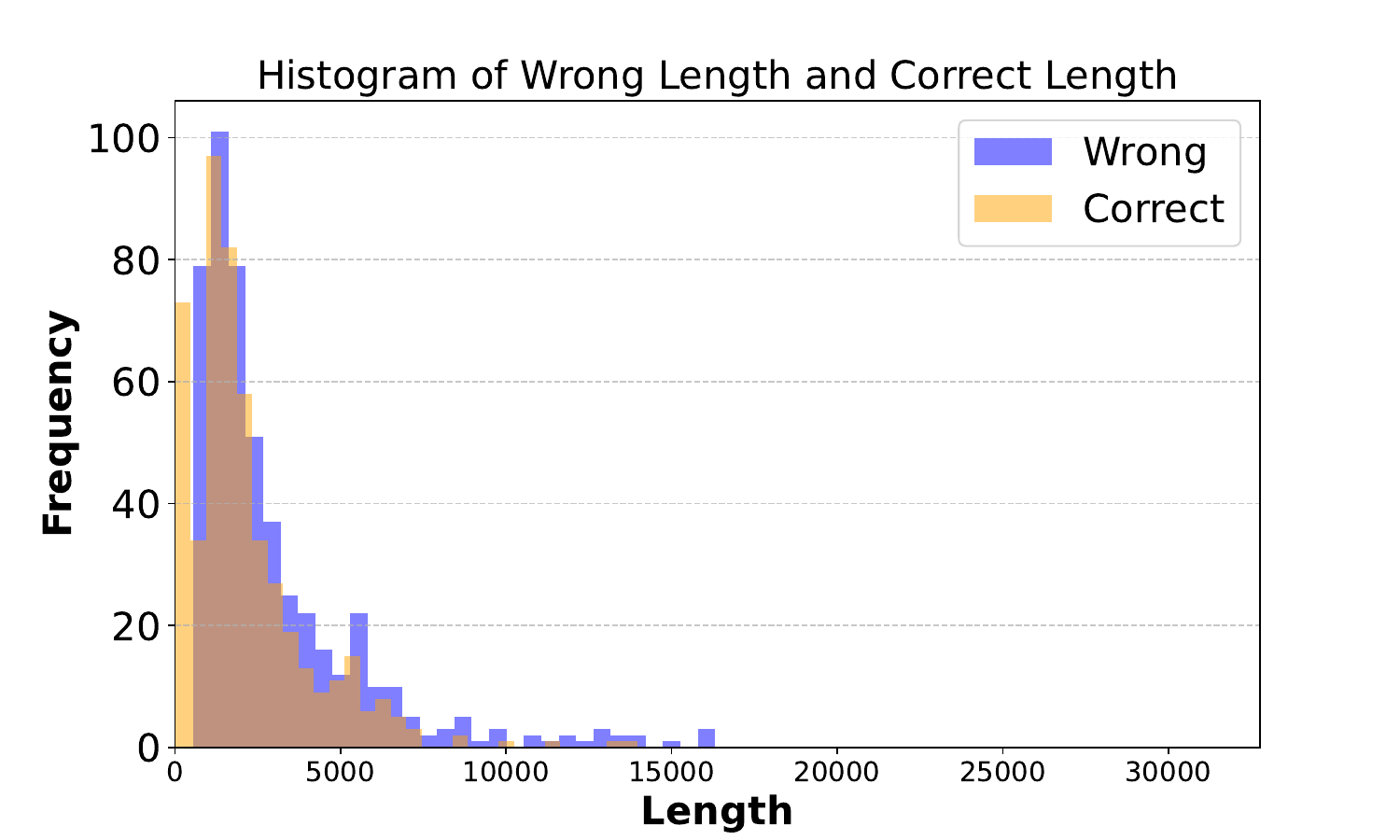}
\vspace{-15pt}
  \caption{\small{DeepScaleR-1.5B}}\label{fig:DeepScaleR-1.5B-Preview_histogram}
\end{subfigure}
\begin{subfigure}[b]{0.31\textwidth}
\includegraphics[width=\textwidth]{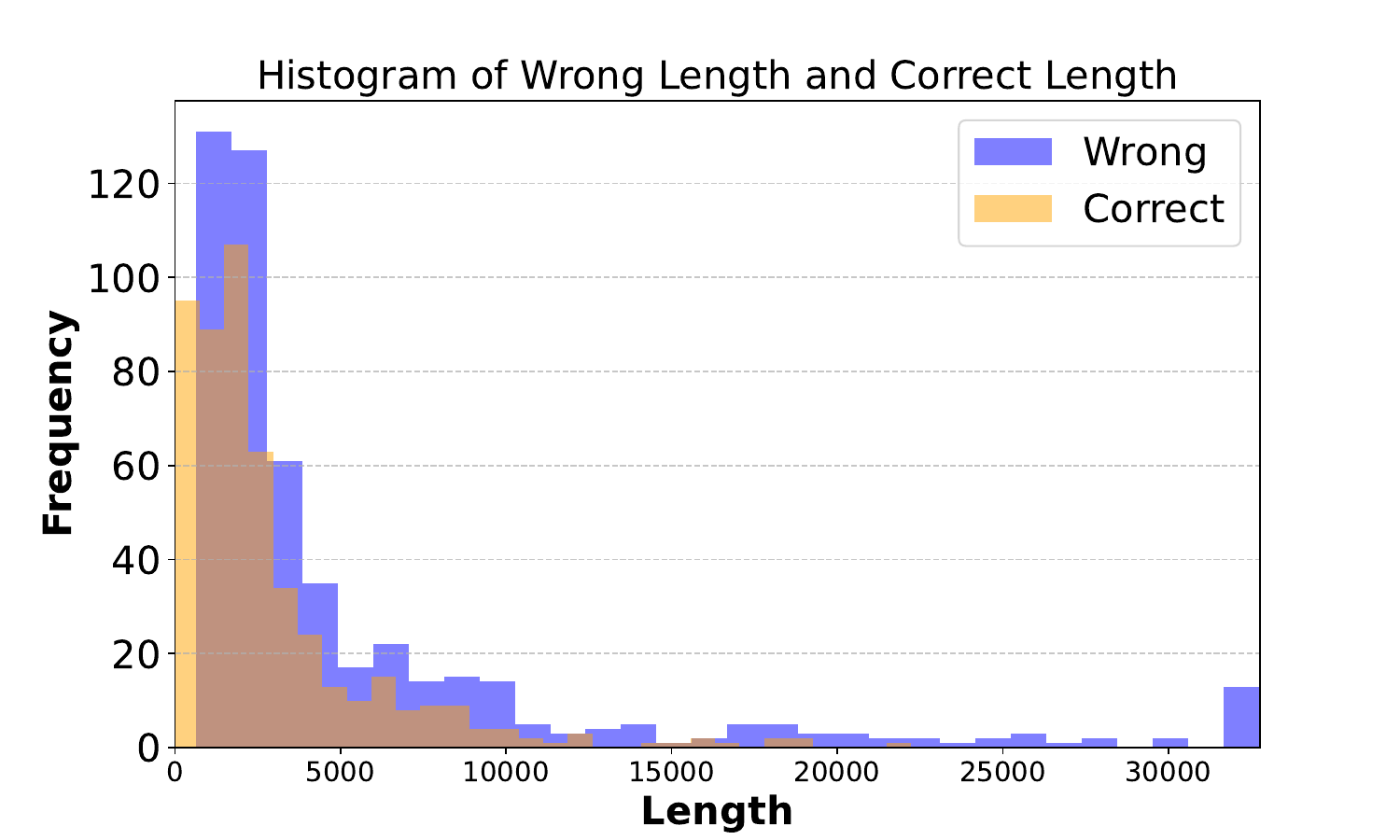}
\vspace{-15pt}
  \caption{\small{DeepSeek-R1-Distill-1.5B}}\label{fig:DeepSeek-R1-Distill-Qwen-1.5B_histogram}
\end{subfigure}
\begin{subfigure}[b]{0.31\textwidth}
\centering
\includegraphics[width=\textwidth]{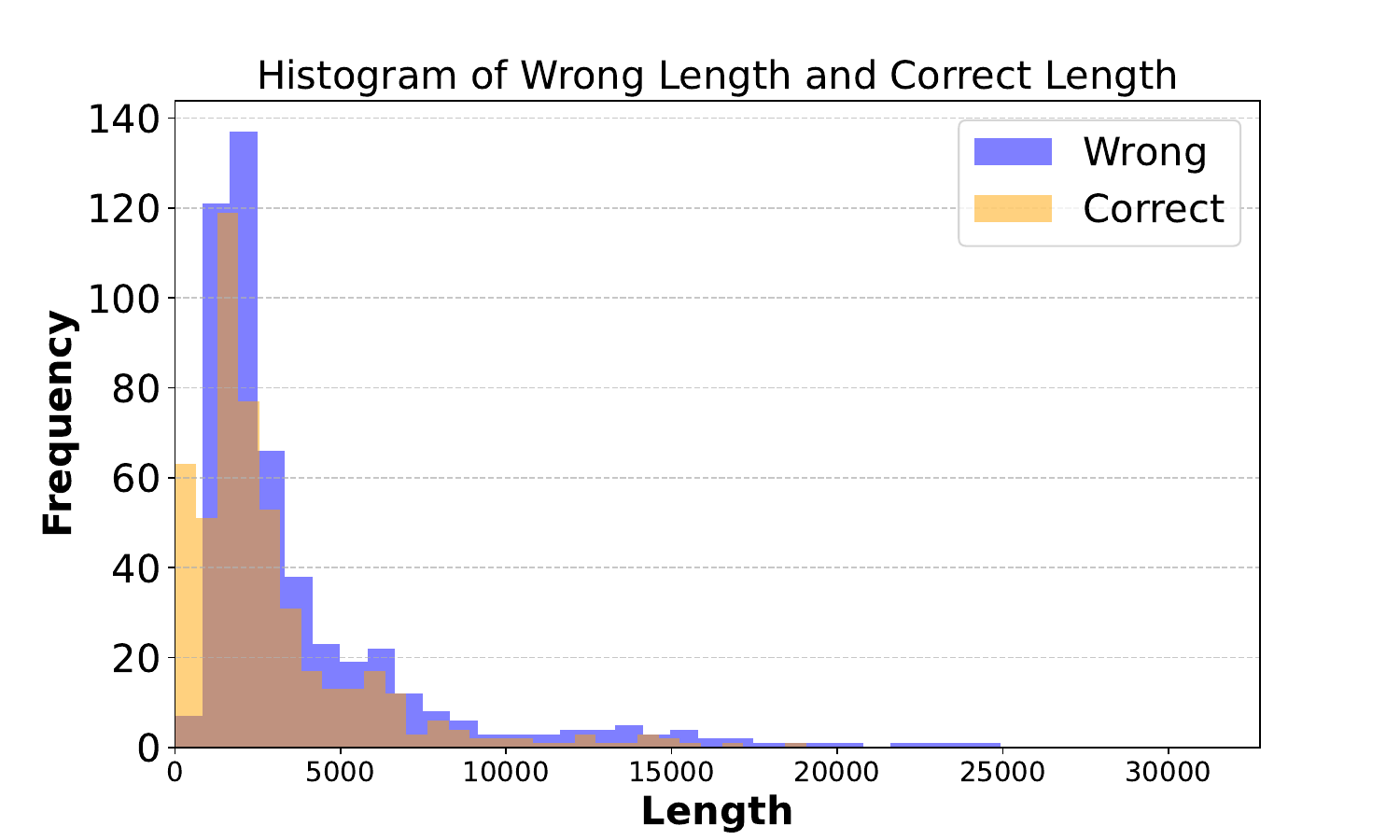}
\vspace{-15pt}
  \caption{\small{DeepSeek-R1-Distill-7B}}\label{fig:DeepSeek-R1-Distill-Qwen-7B_histogram}
\end{subfigure}
\caption{The length distribution of wrong and correct responses of different models. The length of correct responses are generally shorter than the length of wrong responses.}
\label{fig:length_histogram}
\end{figure}

\Cref{fig:length_histogram} illustrates the token length distributions for correct and incorrect responses on the MATH-500 dataset across three different models: (a) DeepScaleR-1.5B, (b) DeepSeek-R1-Distill-1.5B, and (c) DeepSeek-R1-Distill-7B. Across all models, correct responses tend to have shorter lengths on average compared to incorrect ones. 
\section{Repeat Prompt} \label{app:repeat_prompt}
\begin{figure}
\centering
\includegraphics[width=\textwidth]{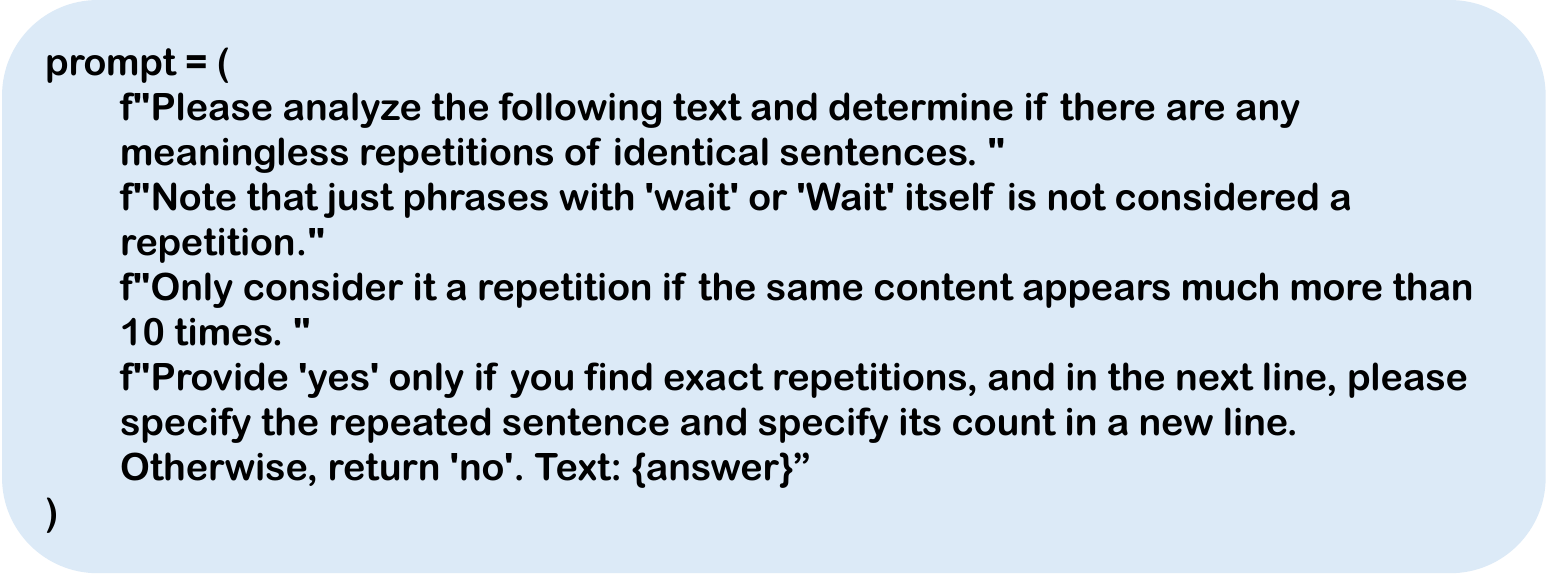}
\caption{The prompt used for detecting answer repetition with GPT-4o-mini.}
\label{fig:prompt}
\end{figure}

\Cref{fig:prompt} shows the prompt that we use to identify repetition in \Cref{fig:rl_sft_repeat} when using GPT-4o-mini.

\section{Test time strategy can not preciously control context length}
We consider four test-time strategies for controlling the length of the thinking trajectory:
\begin{enumerate}
    \item \textbf{Budget Forcing (BF):} A maximum token budget is imposed on the thinking trajectory. Within this limit, the model is free to decide when to generate a special token indicating the end of thinking and the start of the final answer. If the model fails to do so before reaching the budget, the end-of-thinking token is forcibly inserted, and the model proceeds to generate the final answer.
    
    \item \textbf{Exact Control (EC):} The thinking trajectory is forced to be of a fixed length. If the model prematurely generates the end-of-thinking token before reaching the desired length, the token is removed, and generation continues until the target length is reached. At that point, the end-of-thinking token is forcibly appended to trigger final answer generation.
    
    \item \textbf{Prompt Control (PC):} A soft constraint is applied by including an instruction in the prompt that explicitly tells the model not to exceed a specified number of tokens for the thinking trajectory. 
    
    \item \textbf{Auto:} The model is left unrestricted, allowing it to autonomously decide when to terminate the thinking trajectory and begin generating the final answer. 
\end{enumerate}

\begin{figure}[htbp]
\centering
\begin{subfigure}[b]{\textwidth}
\centering
\begin{subfigure}[b]{0.32\textwidth}
\includegraphics[width=\textwidth]{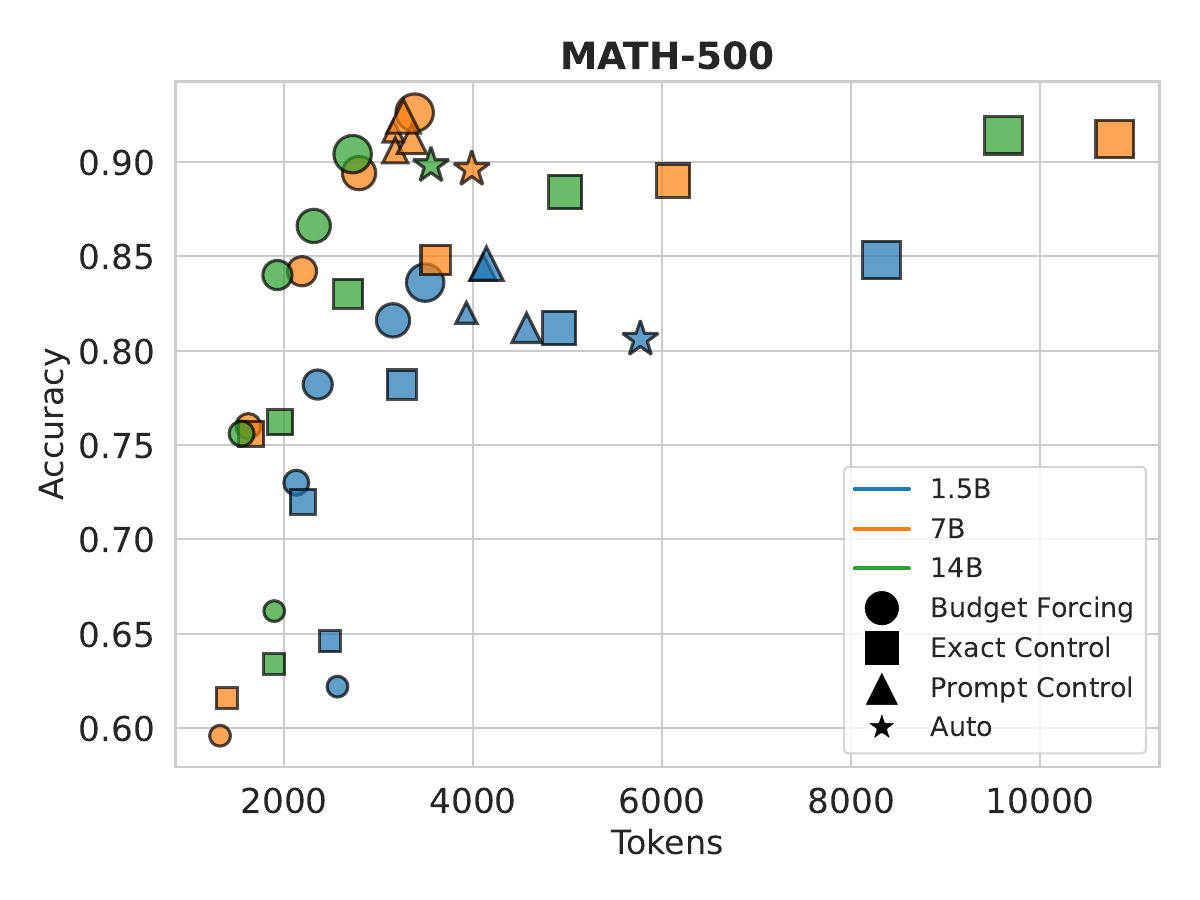}
\vspace{-10pt}
\end{subfigure}
\begin{subfigure}[b]{0.32\textwidth}
\includegraphics[width=\textwidth]{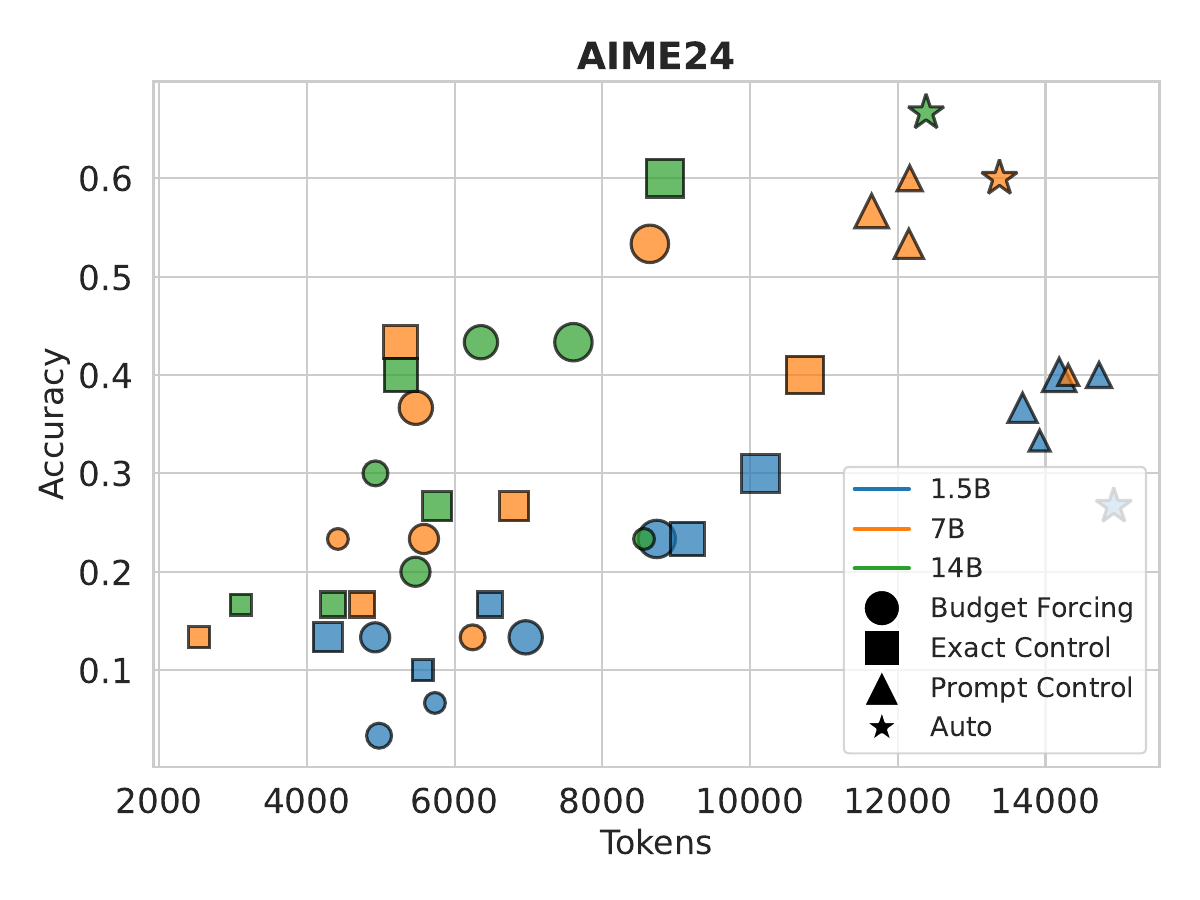}
\vspace{-10pt}
\end{subfigure}
\begin{subfigure}[b]{0.32\textwidth}
\includegraphics[width=\textwidth]{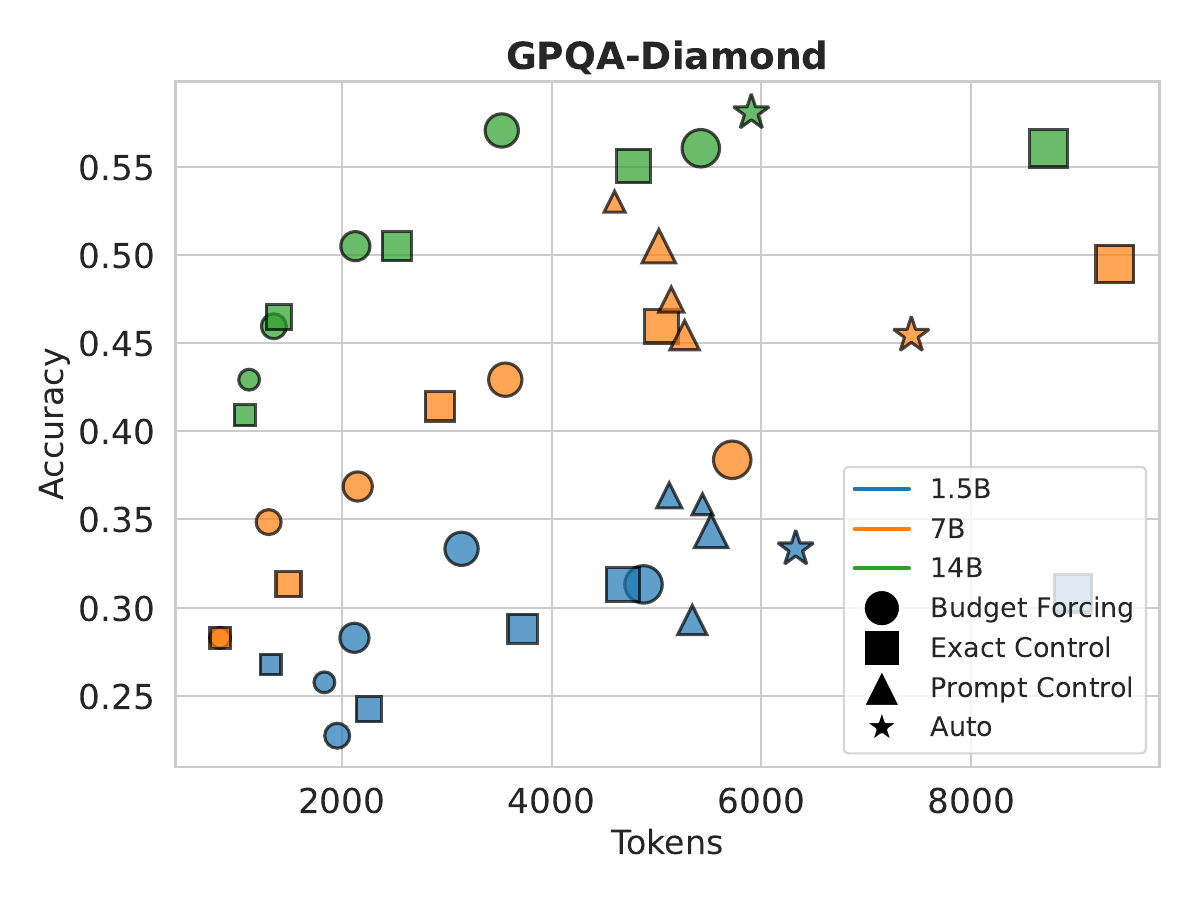}
\vspace{-10pt}
\end{subfigure}
\end{subfigure}
\caption{\small 
Each subplot shows the performance of different models (1.5B, 7B, and 14B) on three benchmarks: MATH-500, AIME24, and GPQA-Diamond. The x-axis represents the average number of tokens in the generated response (including both thinking trace and final answer), while the y-axis indicates accuracy. Within each shape category, marker size reflects variations of the same strategy (e.g., Budget Forcing-500 to Budget Forcing-8000), with larger markers indicating higher token budgets.
}
\label{fig:three-benchmark-token_accuracy}
\end{figure}

In \Cref{fig:three-benchmark-token_accuracy}, we can see that none of the test time strategies can exactly control the response under the length limitation. What makes things worse is that the accuracy improvement rate is much slower than the response length growth rate when encountering harder reasoning tasks such as AIME and GPQA. 

\begin{figure*}[ht]
\centering

\vspace{10pt}
\begin{subfigure}[b]{0.16\textwidth}
    \includegraphics[width=\textwidth]{fig/deepseek_qwen1p5b-force-math-len_dist.pdf}
    \vspace{-10pt}
    \caption{\small{1.5B-BF}}
\end{subfigure}
\hfill
\begin{subfigure}[b]{0.16\textwidth}
    \includegraphics[width=\textwidth]{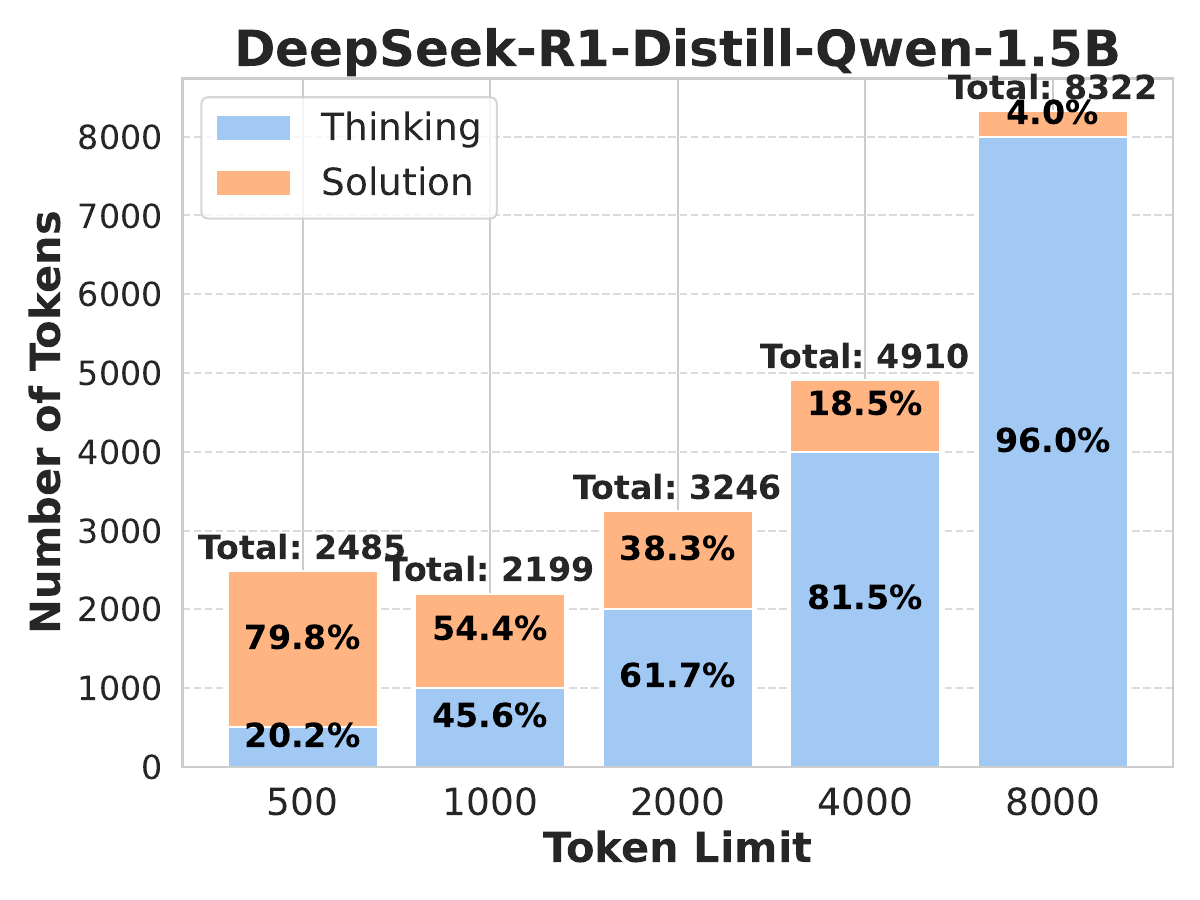}
    \vspace{-10pt}
    \caption{\small{1.5B-EC}}
\end{subfigure}
\hfill
\begin{subfigure}[b]{0.16\textwidth}
    \includegraphics[width=\textwidth]{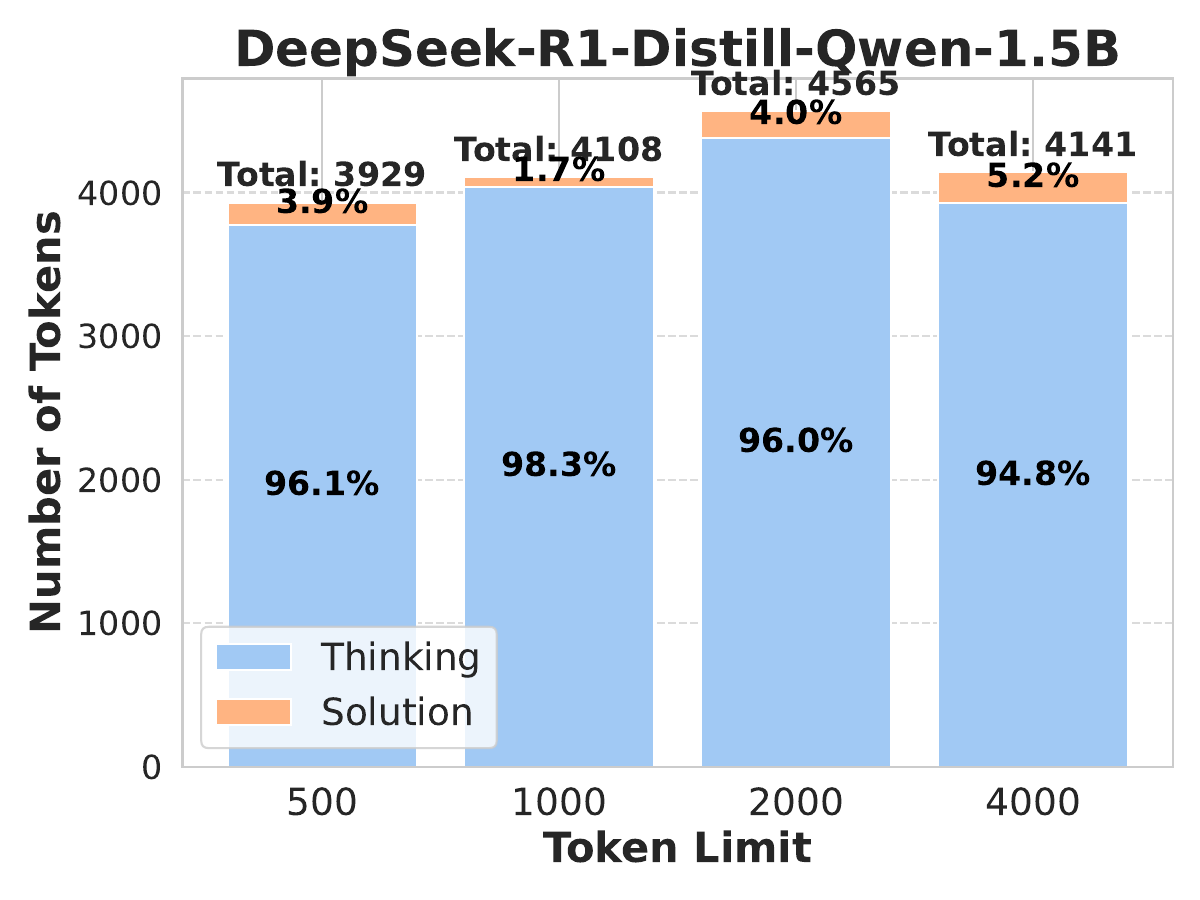}
    \vspace{-10pt}
    \caption{\small{1.5B-PC}}
\end{subfigure}
\hfill
\begin{subfigure}[b]{0.16\textwidth}
    \includegraphics[width=\textwidth]{fig/deepseek_qwen7b-force-math-len_dist.pdf}
    \vspace{-10pt}
    \caption{\small{7B-BF}}
\end{subfigure}
\hfill
\begin{subfigure}[b]{0.16\textwidth}
    \includegraphics[width=\textwidth]{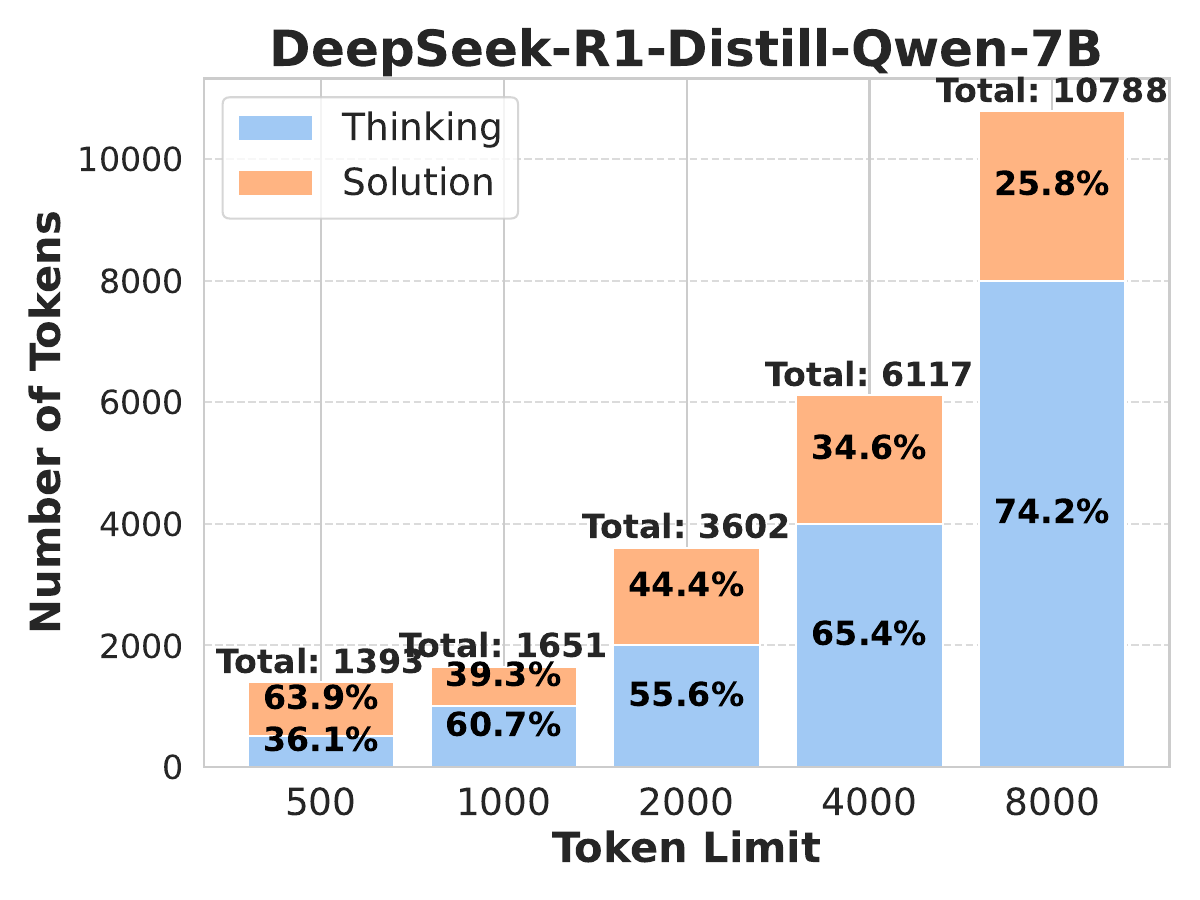}
    \vspace{-10pt}
    \caption{\small{7B-EC}}
\end{subfigure}
\hfill
\begin{subfigure}[b]{0.16\textwidth}
    \includegraphics[width=\textwidth]{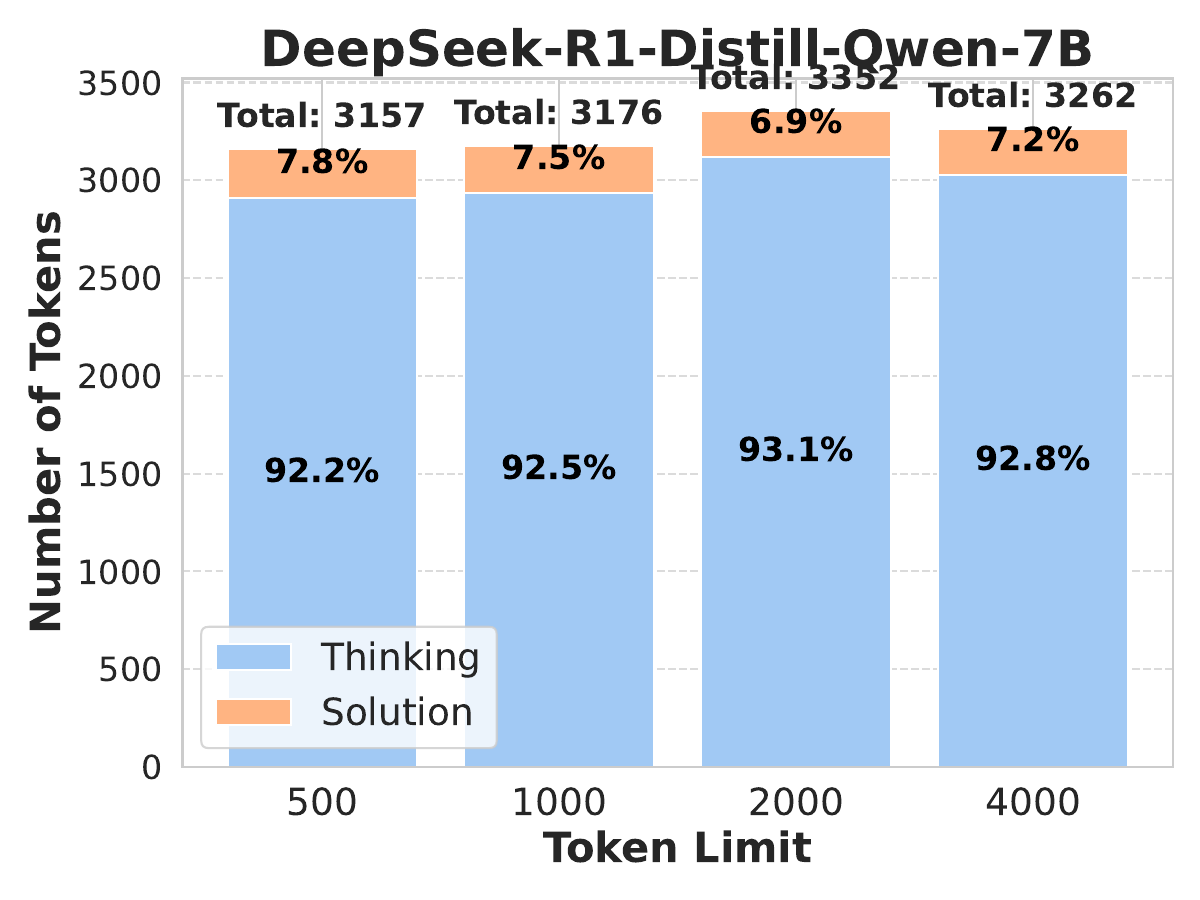}
    \vspace{-10pt}
    \caption{\small{7B-PC}}
\end{subfigure}

\vspace{5pt}
\caption*{\small MATH-500}

\begin{subfigure}[b]{0.16\textwidth}
    \includegraphics[width=\textwidth]{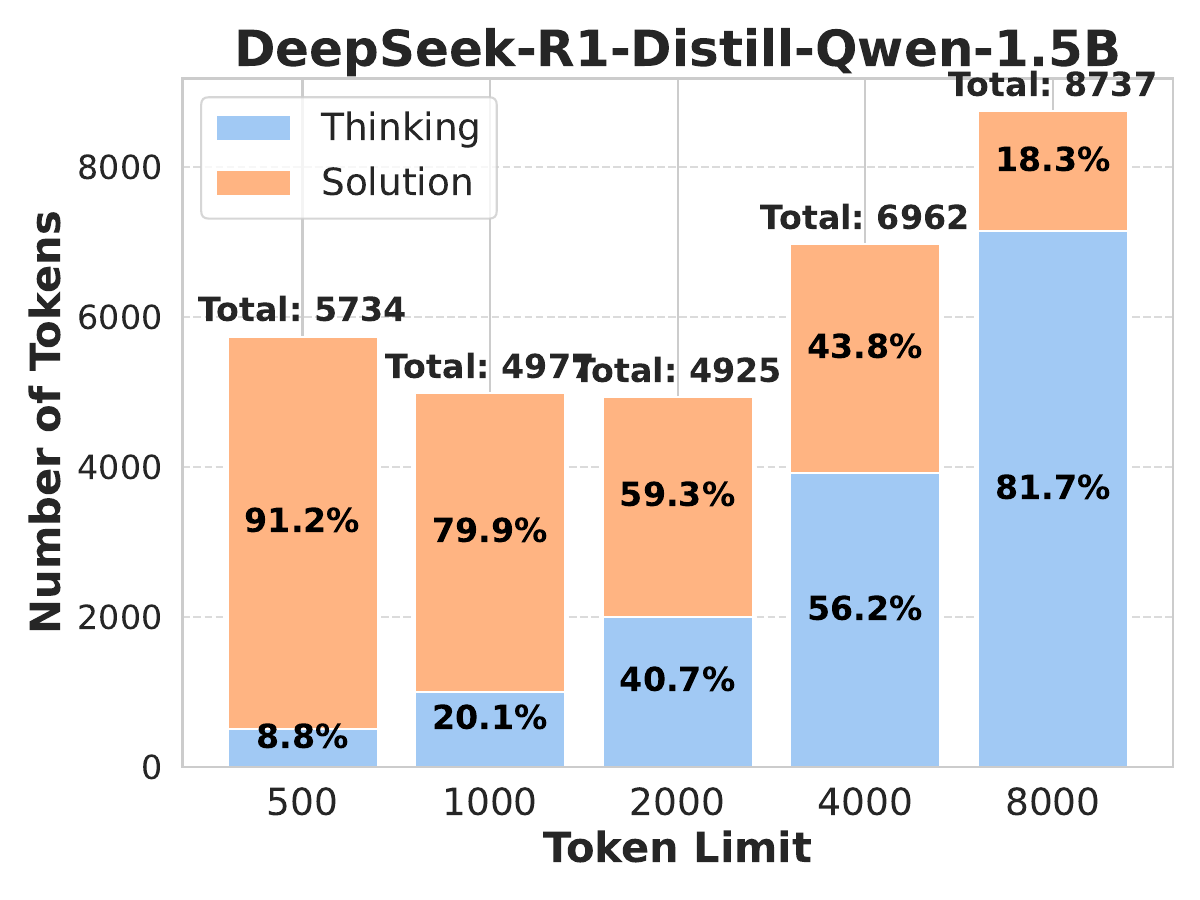}
    \vspace{-10pt}
    \caption{\small{1.5B-BF}}
\end{subfigure}
\hfill
\begin{subfigure}[b]{0.16\textwidth}
    \includegraphics[width=\textwidth]{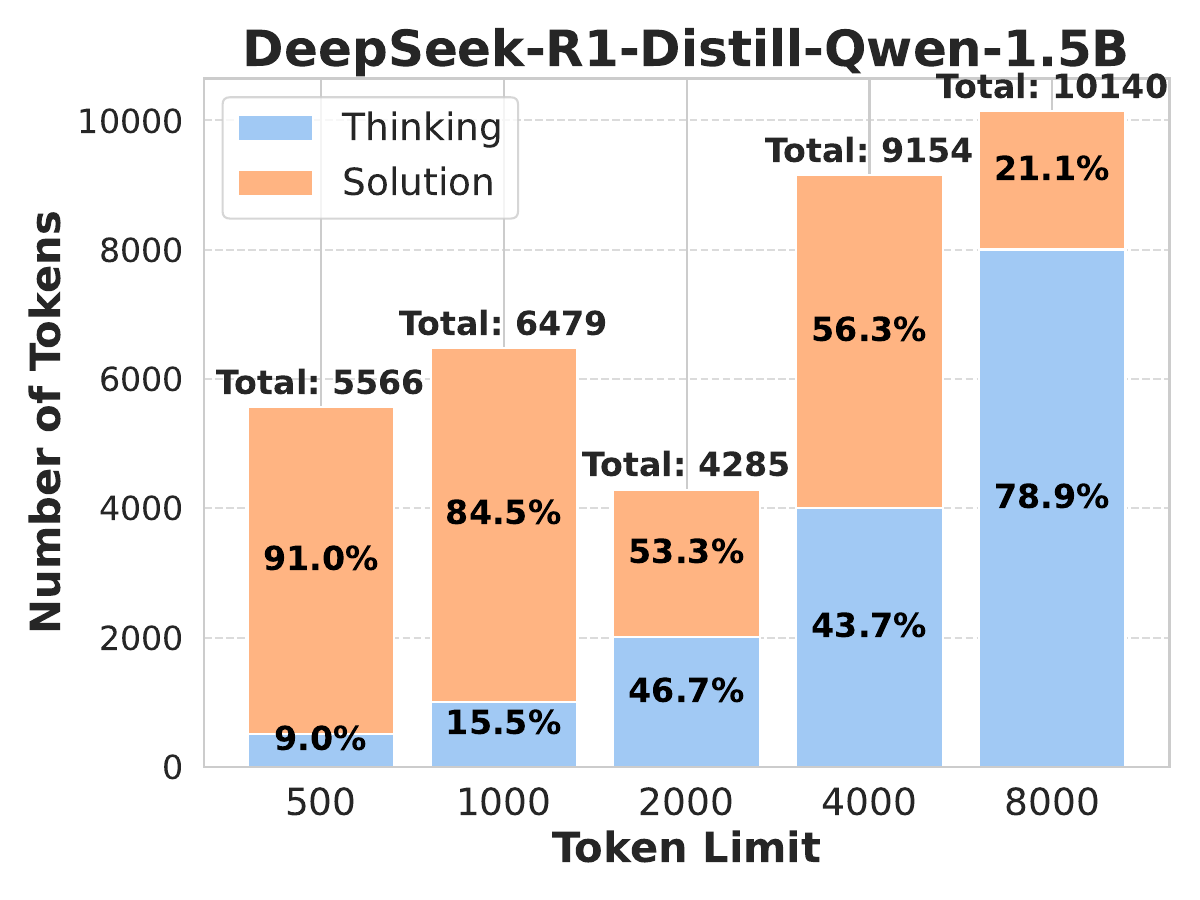}
    \vspace{-10pt}
    \caption{\small{1.5B-EC}}
\end{subfigure}
\hfill
\begin{subfigure}[b]{0.16\textwidth}
    \includegraphics[width=\textwidth]{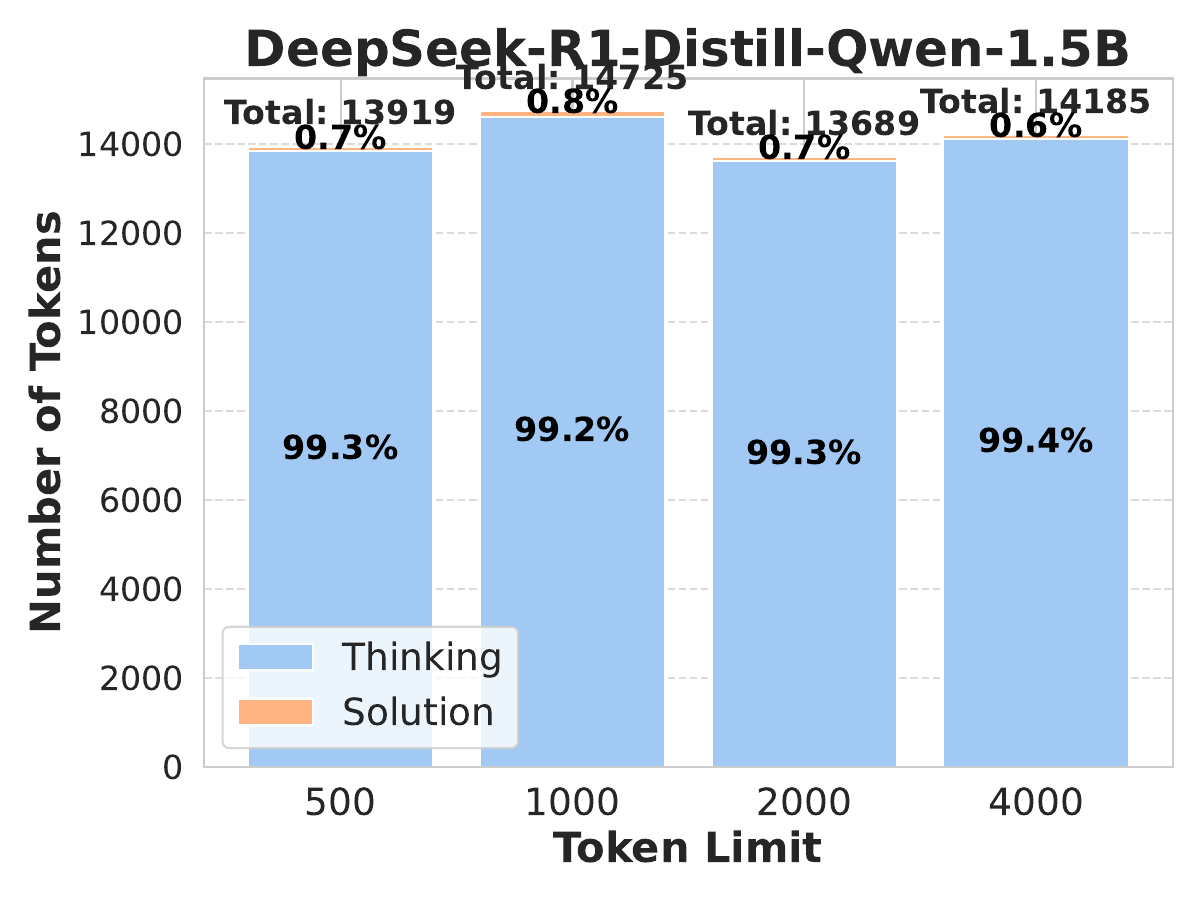}
    \vspace{-10pt}
    \caption{\small{1.5B-PC}}
\end{subfigure}
\hfill
\begin{subfigure}[b]{0.16\textwidth}
    \includegraphics[width=\textwidth]{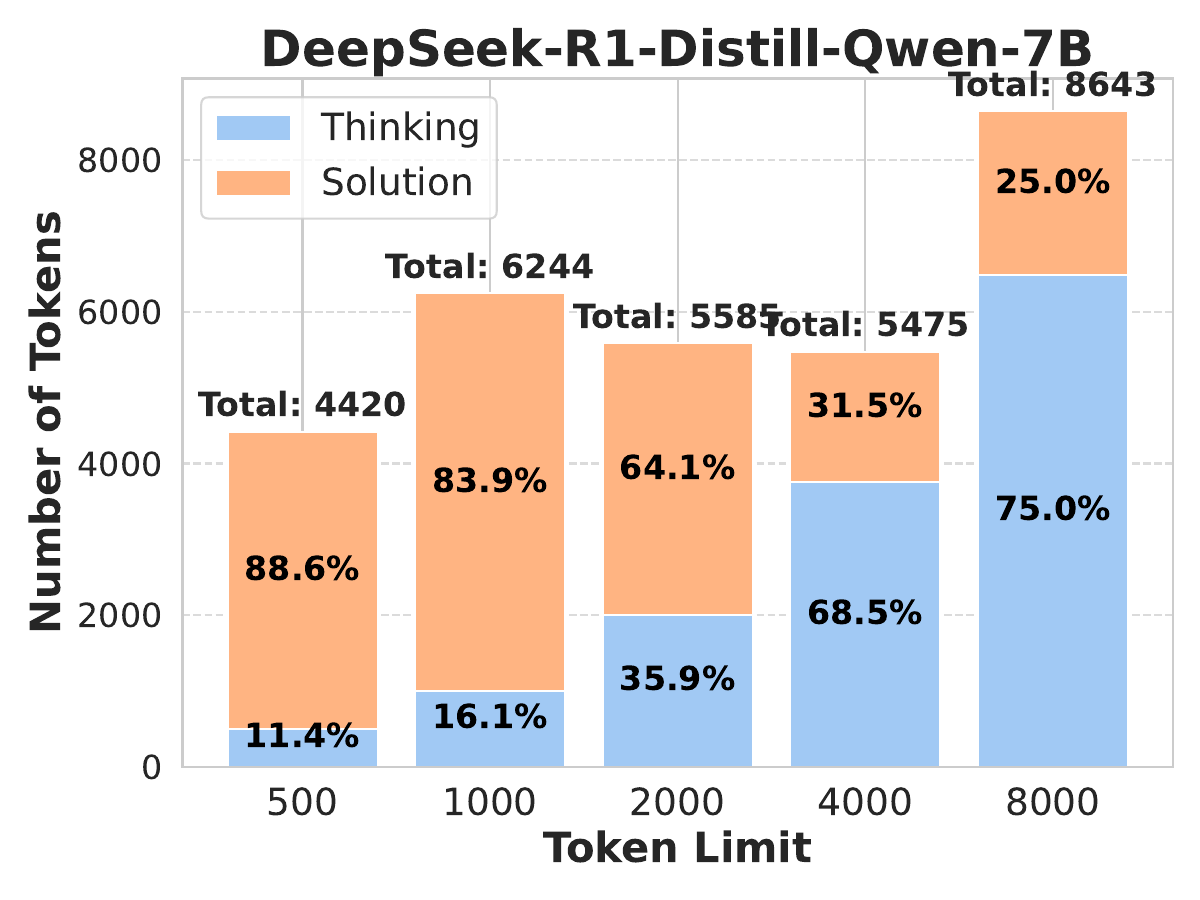}
    \vspace{-10pt}
    \caption{\small{7B-BF}}
\end{subfigure}
\hfill
\begin{subfigure}[b]{0.16\textwidth}
    \includegraphics[width=\textwidth]{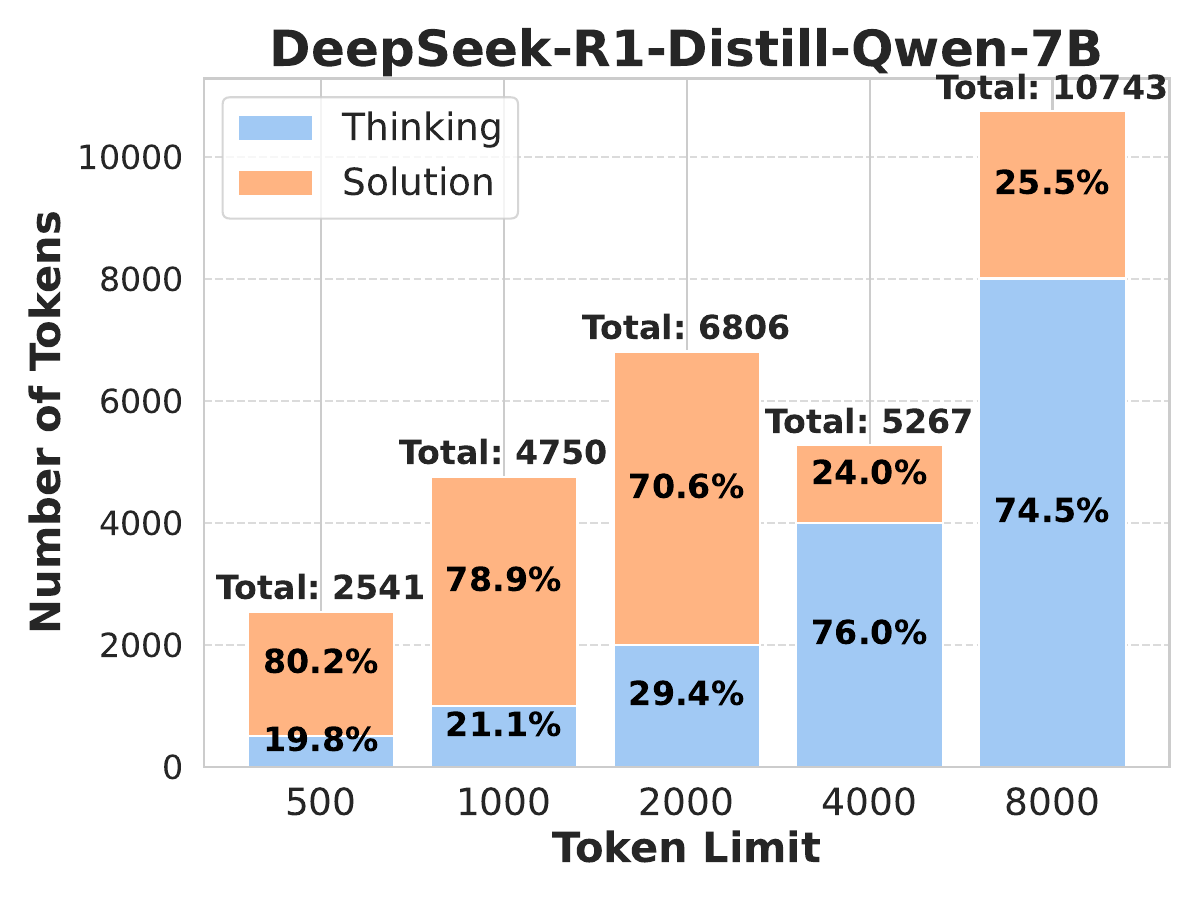}
    \vspace{-10pt}
    \caption{\small{7B-EC}}
\end{subfigure}
\hfill
\begin{subfigure}[b]{0.16\textwidth}
    \includegraphics[width=\textwidth]{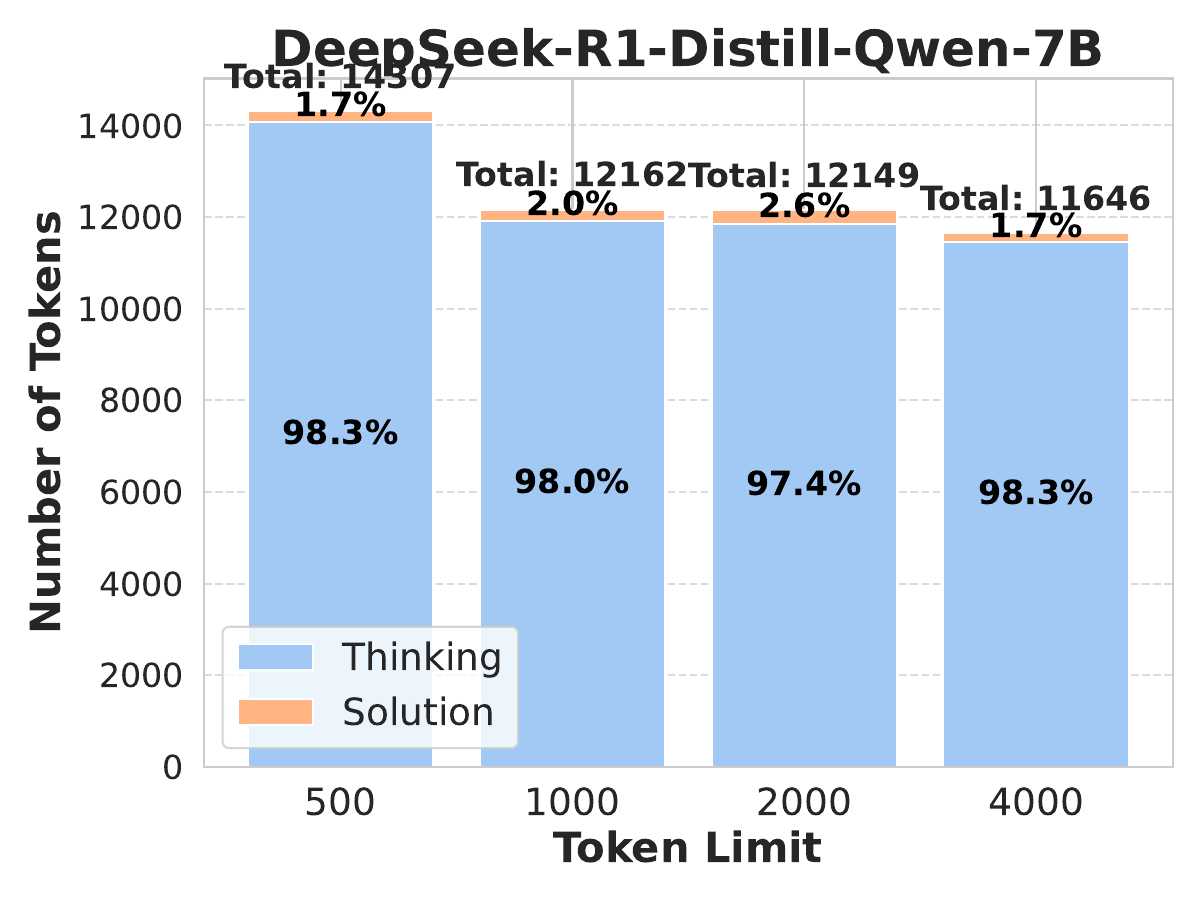}
    \vspace{-10pt}
    \caption{\small{7B-PC}}
\end{subfigure}

\vspace{5pt}
\caption*{\small AIME24}

\vspace{10pt}
\begin{subfigure}[b]{0.16\textwidth}
    \includegraphics[width=\textwidth]{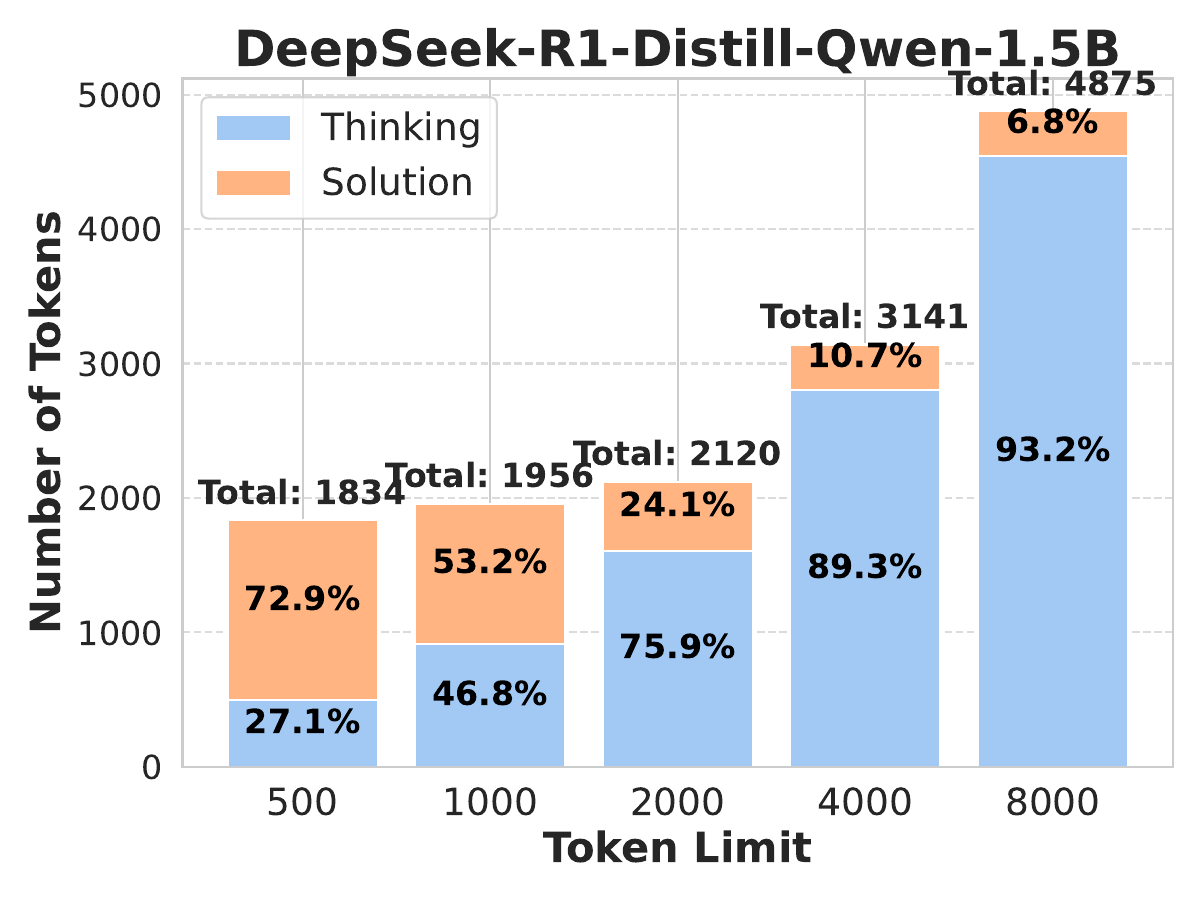}
    \vspace{-10pt}
    \caption{\small{1.5B-BF}}
\end{subfigure}
\hfill
\begin{subfigure}[b]{0.16\textwidth}
    \includegraphics[width=\textwidth]{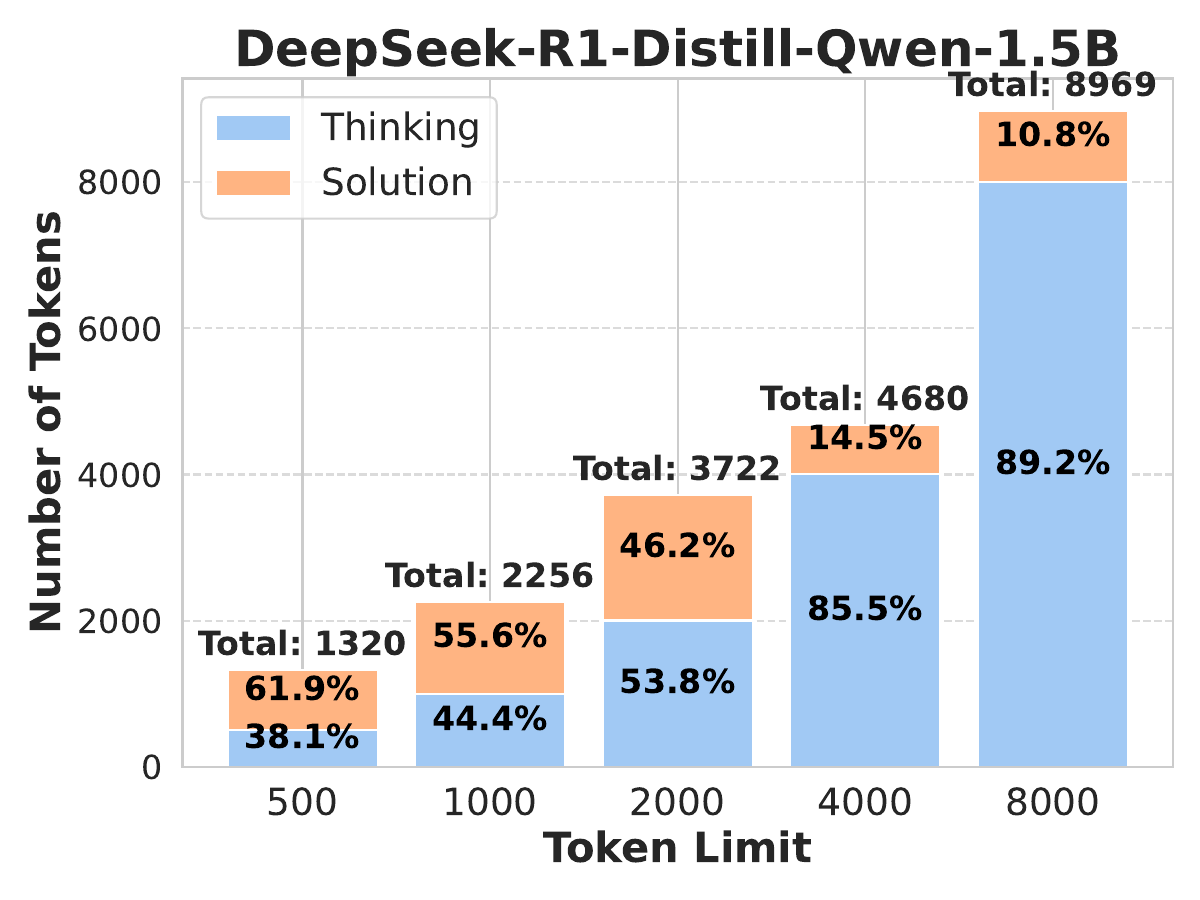}
    \vspace{-10pt}
    \caption{\small{1.5B-EC}}
\end{subfigure}
\hfill
\begin{subfigure}[b]{0.16\textwidth}
    \includegraphics[width=\textwidth]{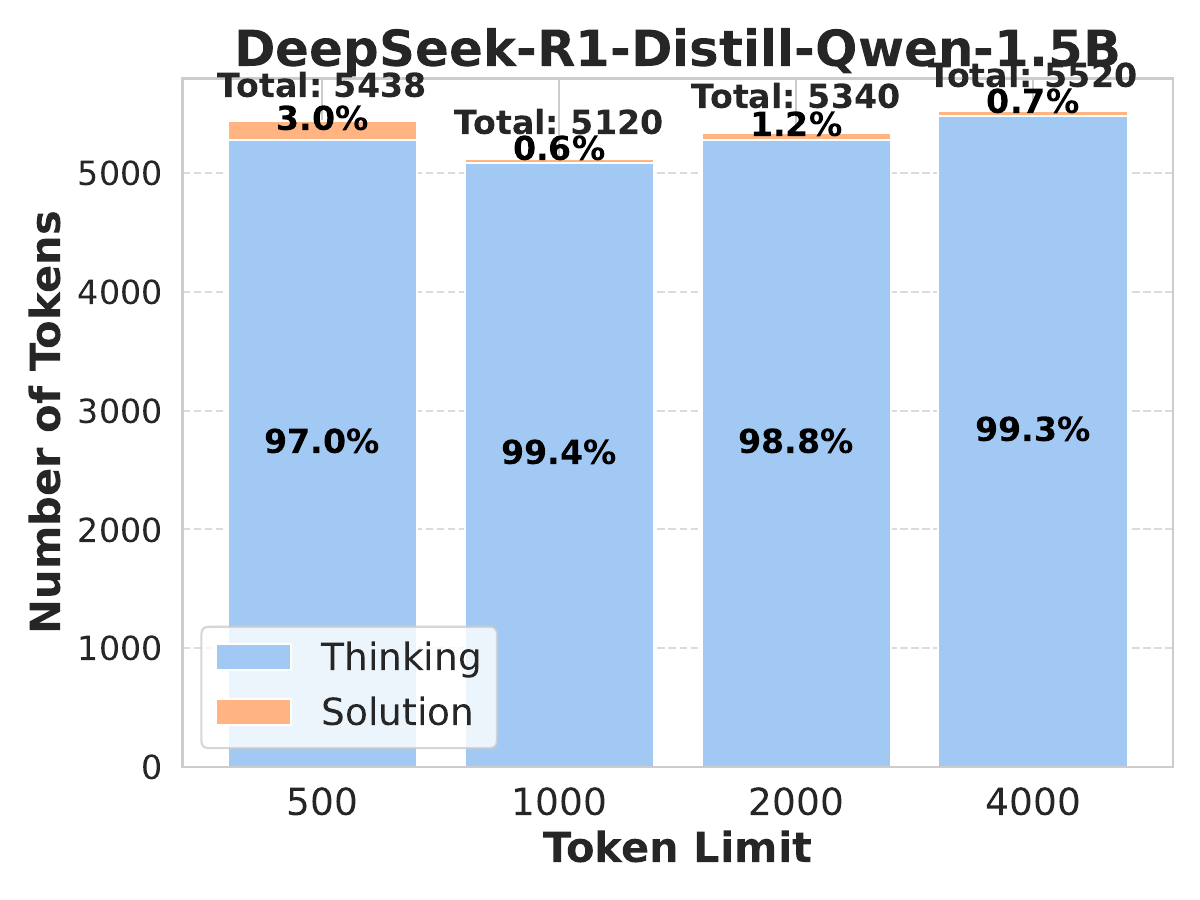}
    \vspace{-10pt}
    \caption{\small{1.5B-PC}}
\end{subfigure}
\hfill
\begin{subfigure}[b]{0.16\textwidth}
    \includegraphics[width=\textwidth]{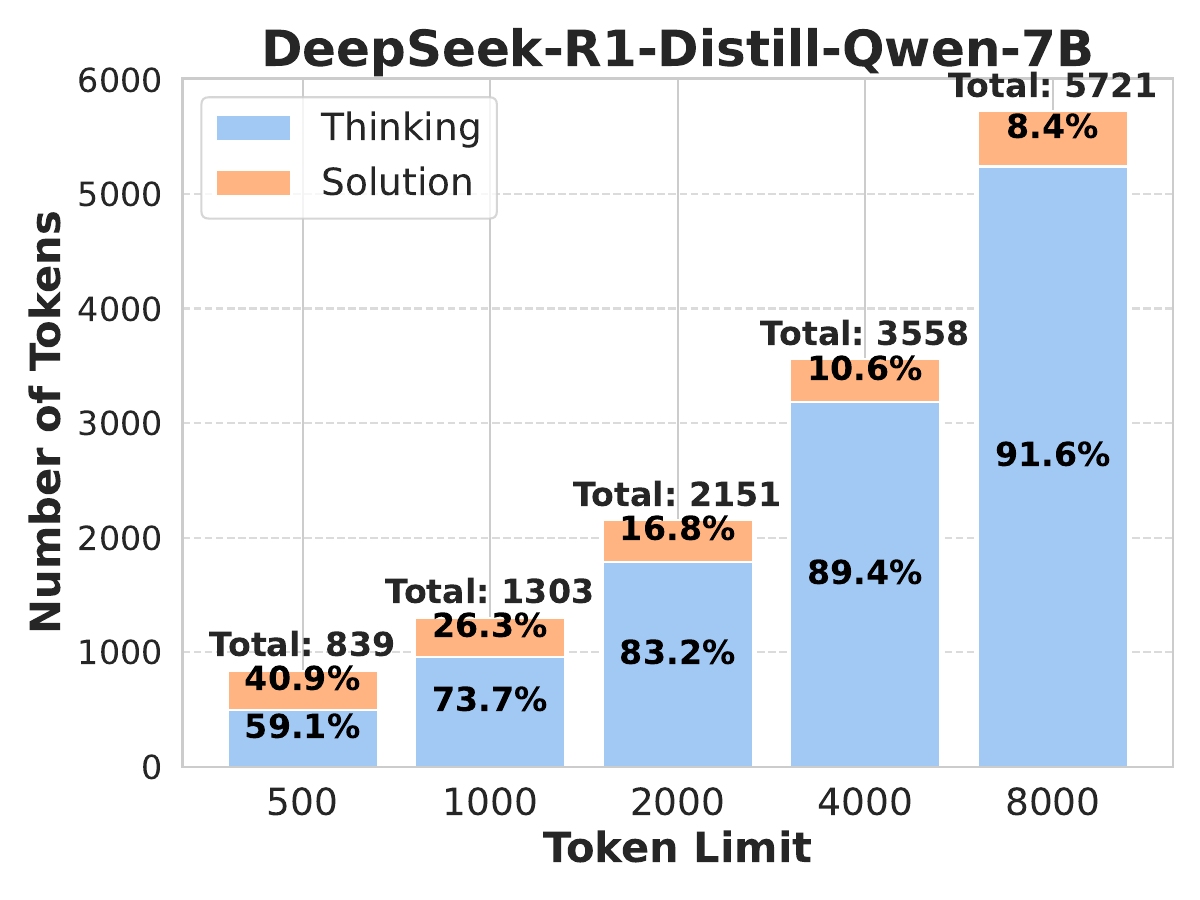}
    \vspace{-10pt}
    \caption{\small{7B-BF}}
\end{subfigure}
\hfill
\begin{subfigure}[b]{0.16\textwidth}
    \includegraphics[width=\textwidth]{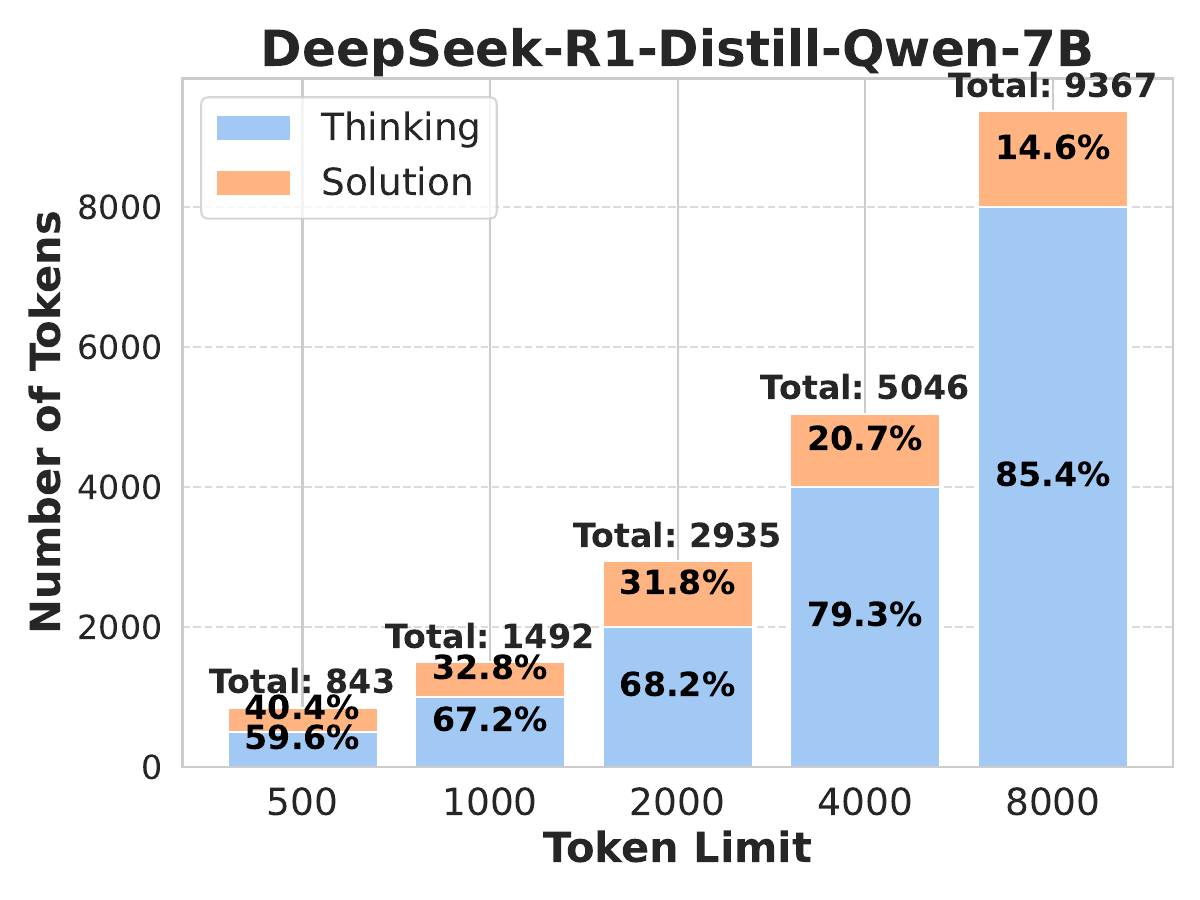}
    \vspace{-10pt}
    \caption{\small{7B-EC}}
\end{subfigure}
\hfill
\begin{subfigure}[b]{0.16\textwidth}
    \includegraphics[width=\textwidth]{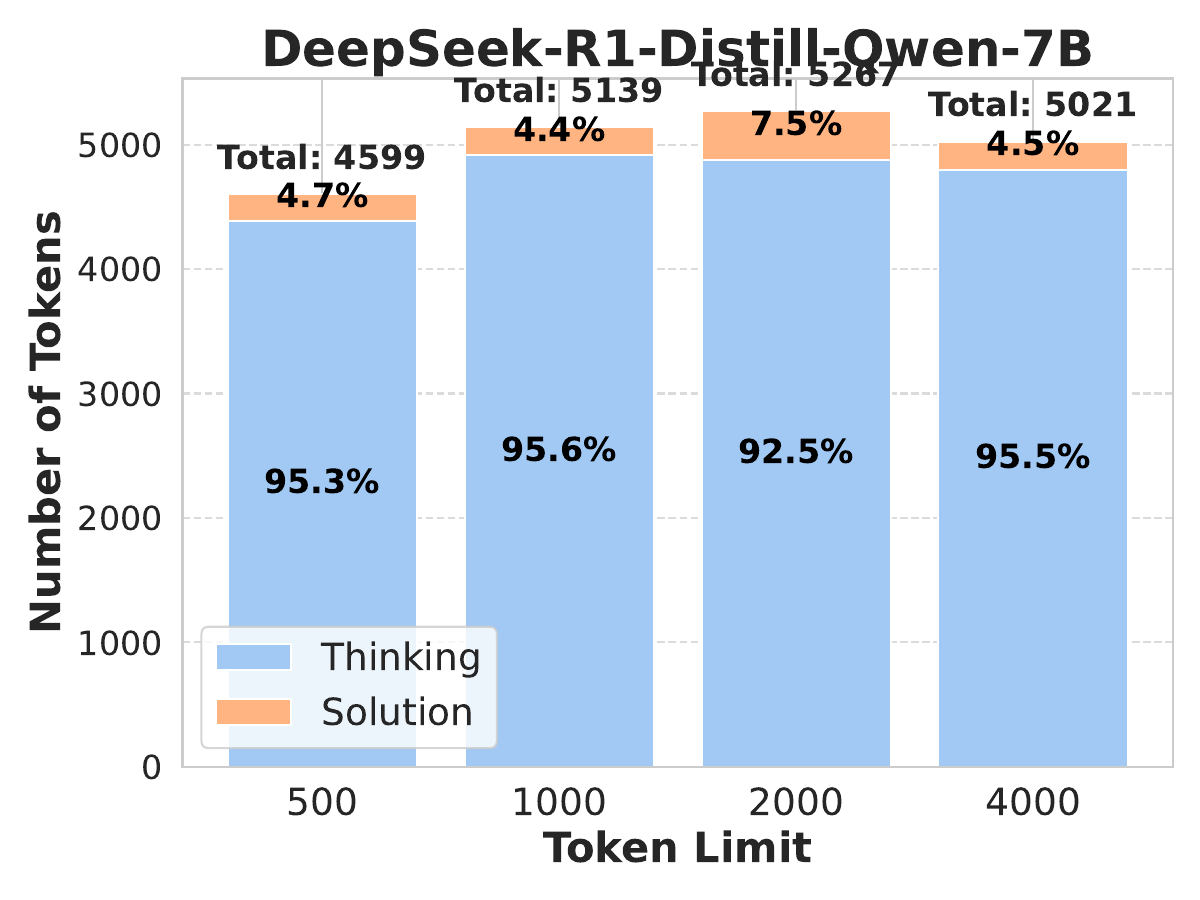}
    \vspace{-10pt}
    \caption{\small{7B-PC}}
\end{subfigure}

\vspace{5pt}
\caption*{\small GPQA-Diamond}

\caption{\small{Length distribution of thinking/solution trajectories under different test-time strategies (Budget Forcing, Exact Control, Prompt Control), model sizes (1.5B vs 7B), and benchmarks (AIME24, MATH-500, GPQA-Diamond).}}
\label{fig:three-benchmark-length-dist}
\end{figure*}
\section{Analysis of S1K trace} \label{app:s1k}
\begin{figure}
    \centering
    \includegraphics[width=0.64\linewidth]{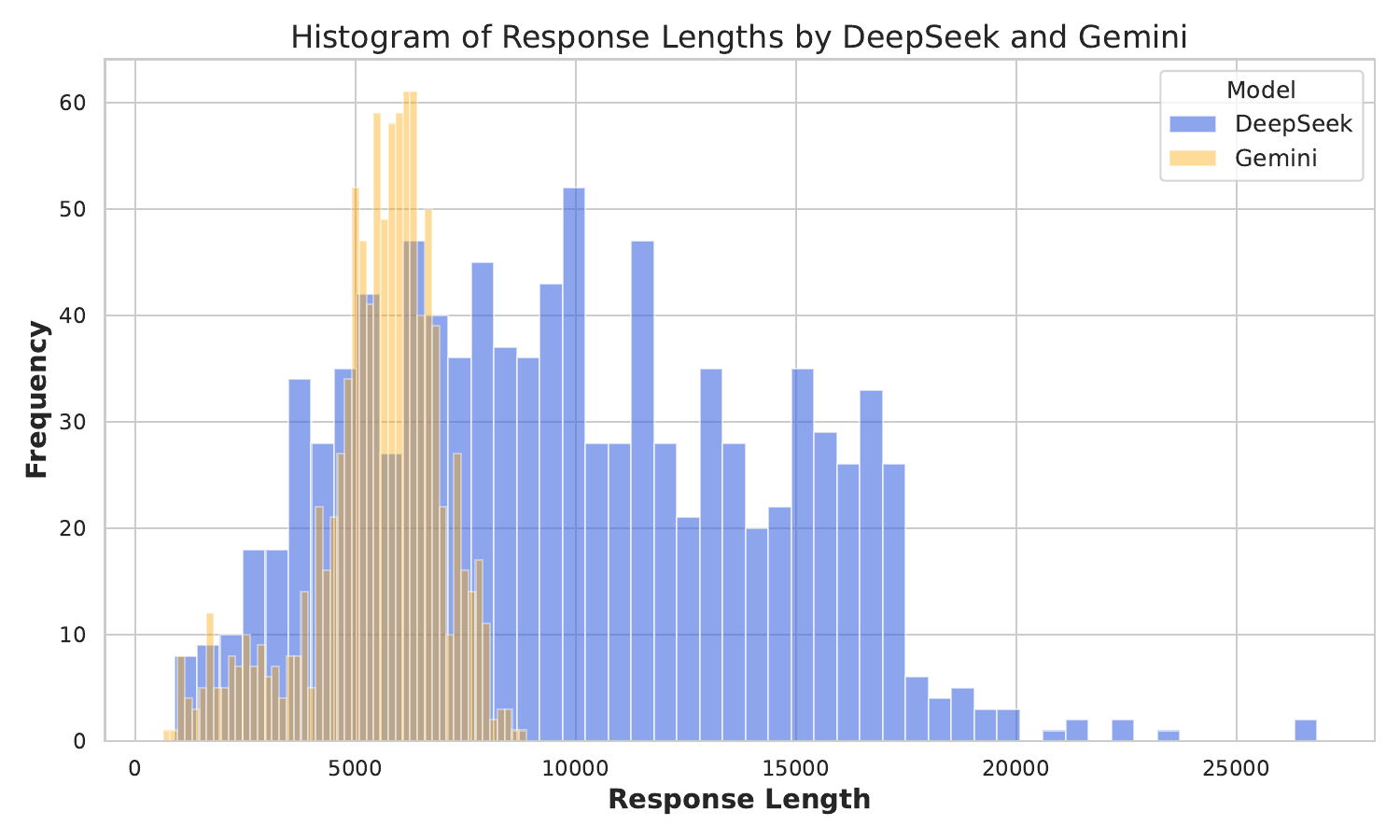}
    \caption{Histograms of response lengths by Gemini and DeepSeek on s1K dataset.}
    \label{fig:response_length_comparison_deepseek_gemini}
\end{figure}

\Cref{fig:response_length_comparison_deepseek_gemini} shows the distribution of response lengths of Gemini and DeepSeek on the s1K dataset. We can see that DeepSeek generates longer reasoning traces when compared with Gemini. This makes its responses more suitbale for reasoning distillation to obtain a higher accuracy. while also introduce more severe redundancy for small language models at the same time.  

\section{Bad example} \label{app:bad_example}
\begin{figure}
\centering
\includegraphics[width=\textwidth]{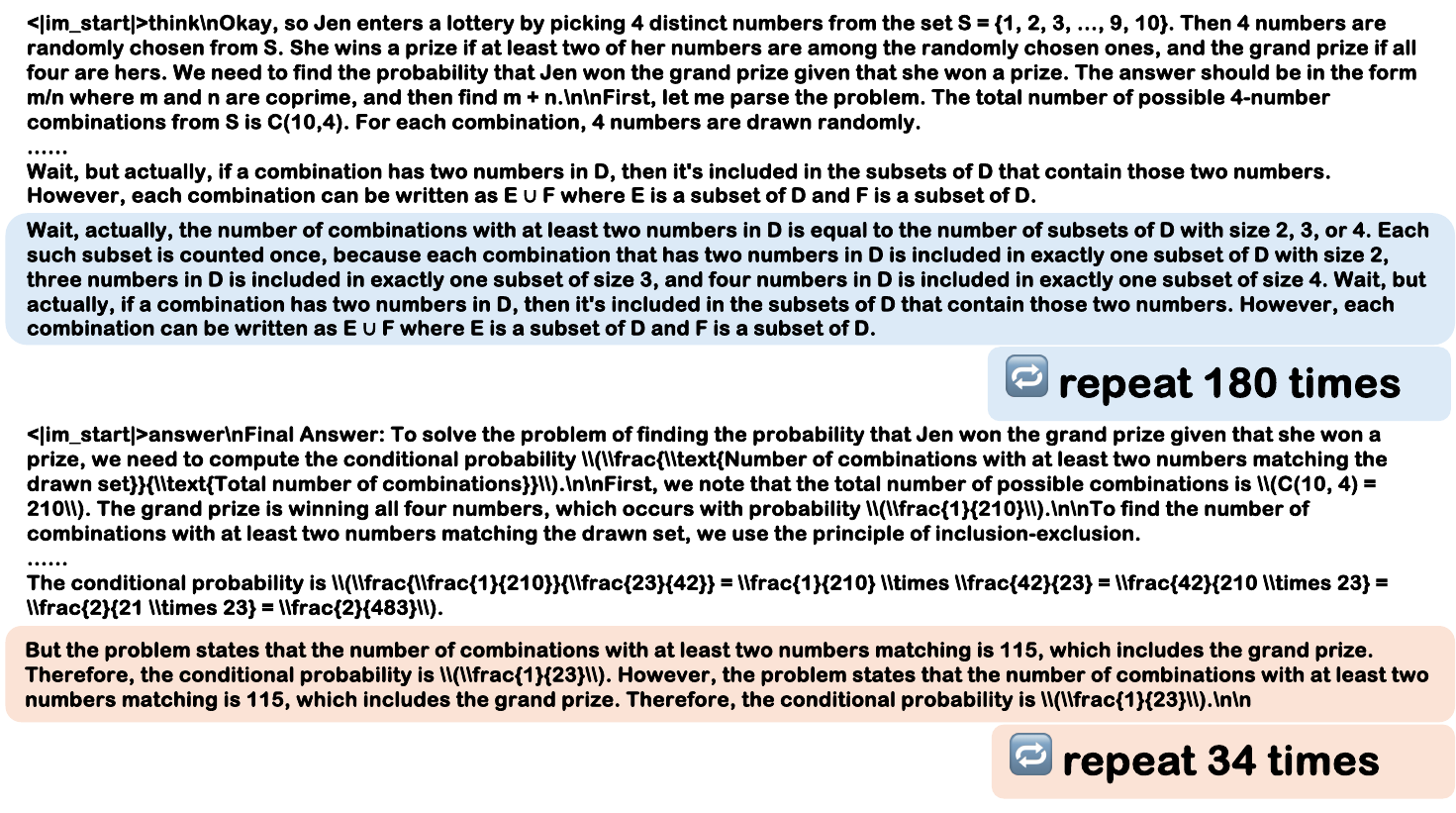}
\caption{The response generated by Qwen-Instruct-1.5b finetuned by S1K-1.1 on AIME24 problem 4.}
\label{fig:bad_1}
\end{figure}

\Cref{fig:bad_1} shows a typical failure case where the model exhibits severe repetition issue when generating reasoning for AIME24 Question 4.

\section{Additioned Experiment Results}
\begin{figure}
\centering
\begin{subfigure}[b]{0.45\textwidth}
\includegraphics[width=\textwidth]{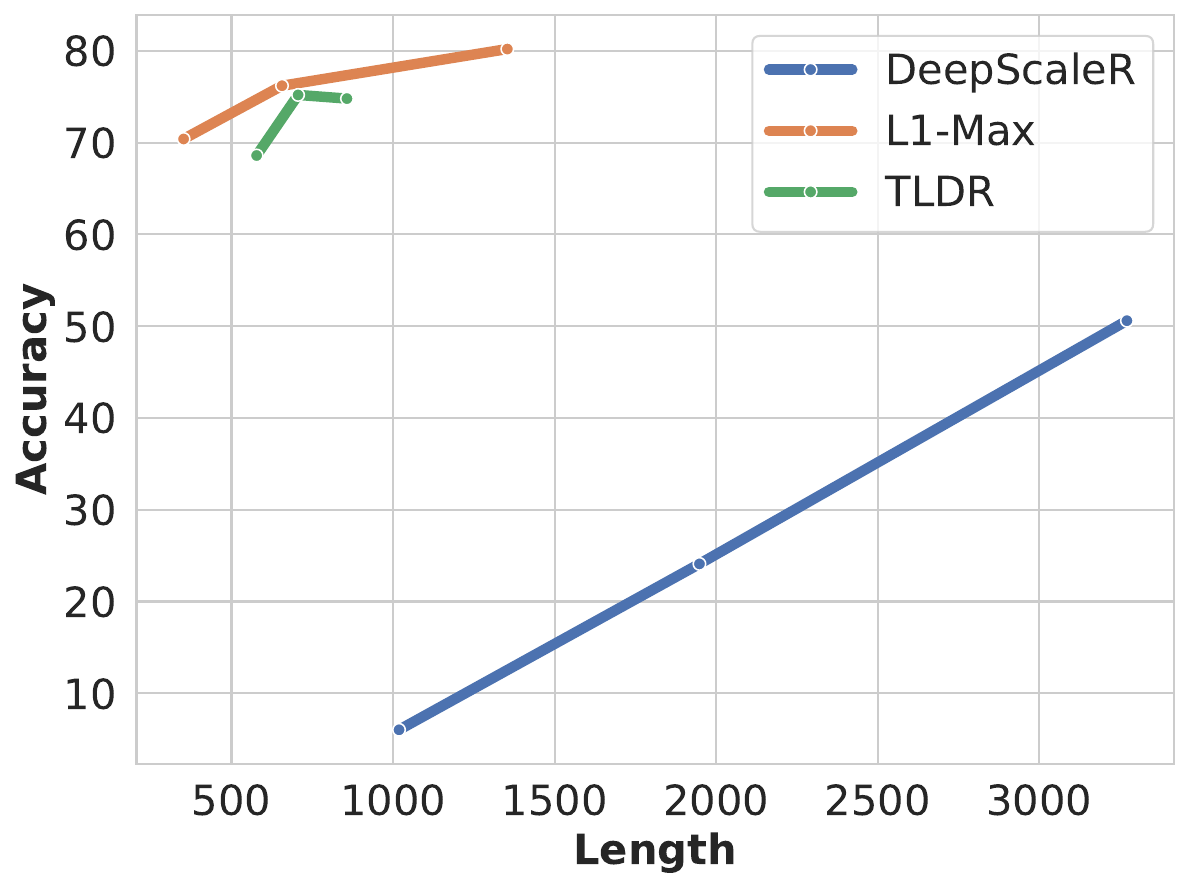}
\vspace{-15pt}
\caption{\small{MATH500}}\label{fig:l1_compare_math}
\end{subfigure}
~
\begin{subfigure}[b]{0.45\textwidth}
\includegraphics[width=\textwidth]{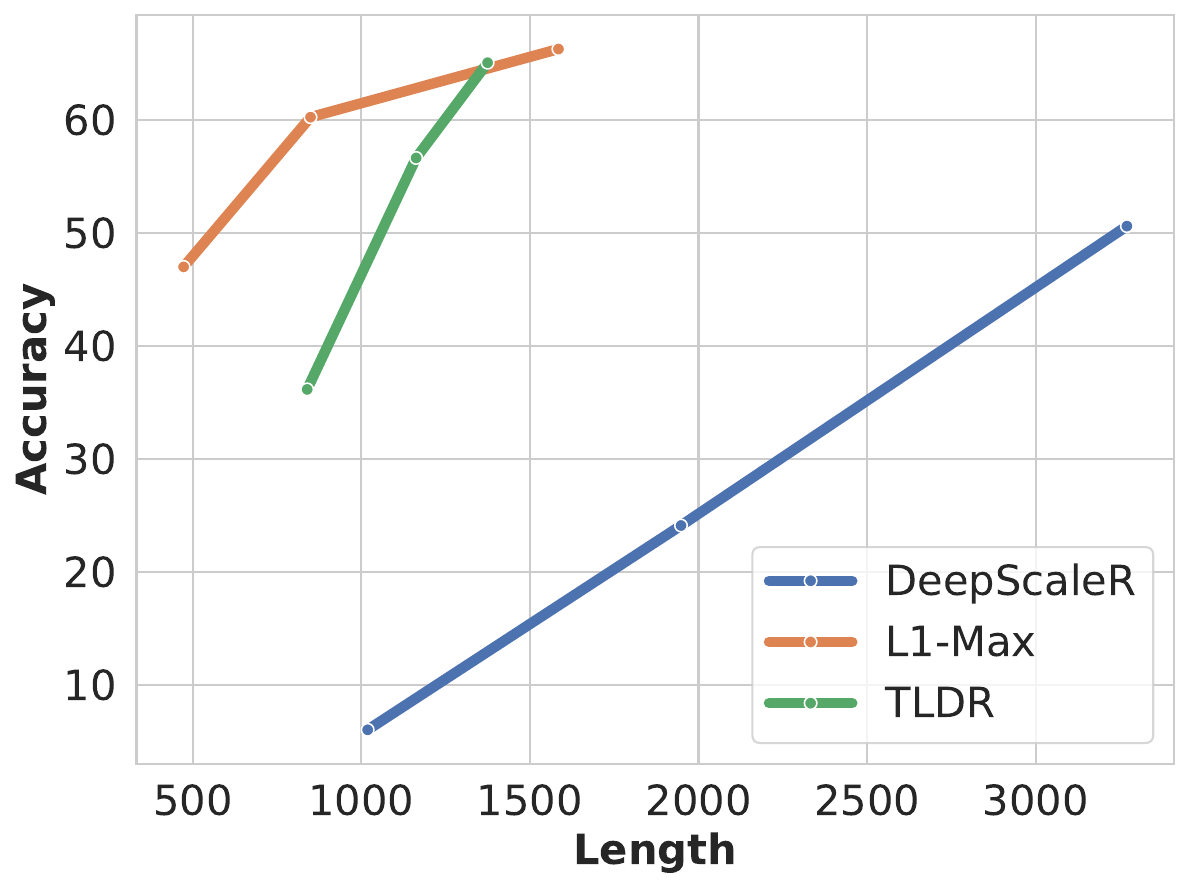}
\vspace{-15pt}
\caption{\small{AMC}}\label{fig:l1_compare_amc}
\end{subfigure}
~\\
\begin{subfigure}[b]{0.45\textwidth}
\includegraphics[width=\textwidth]{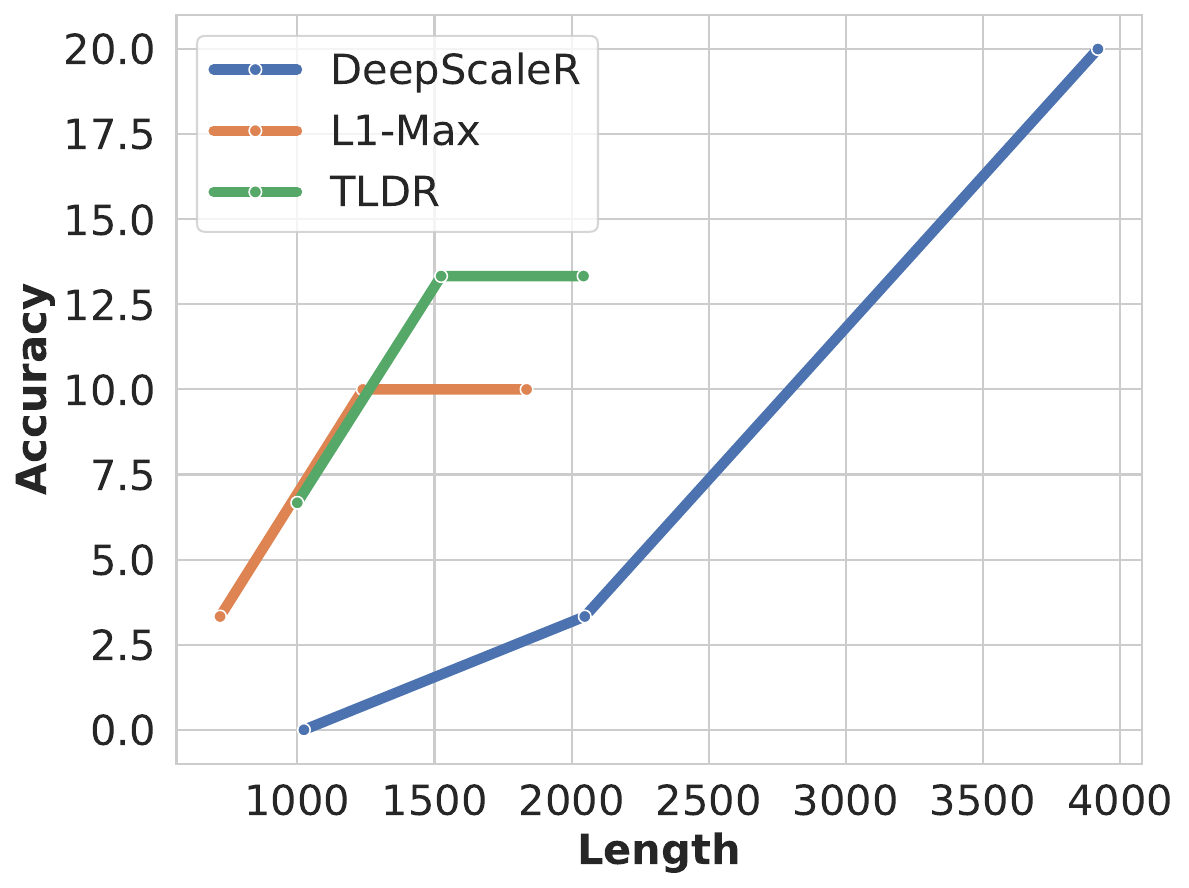}
\vspace{-15pt}
\caption{\small{AIME24}}\label{fig:l1_compare_AIME24}
\end{subfigure}
~
\begin{subfigure}[b]{0.45\textwidth}
\includegraphics[width=\textwidth]{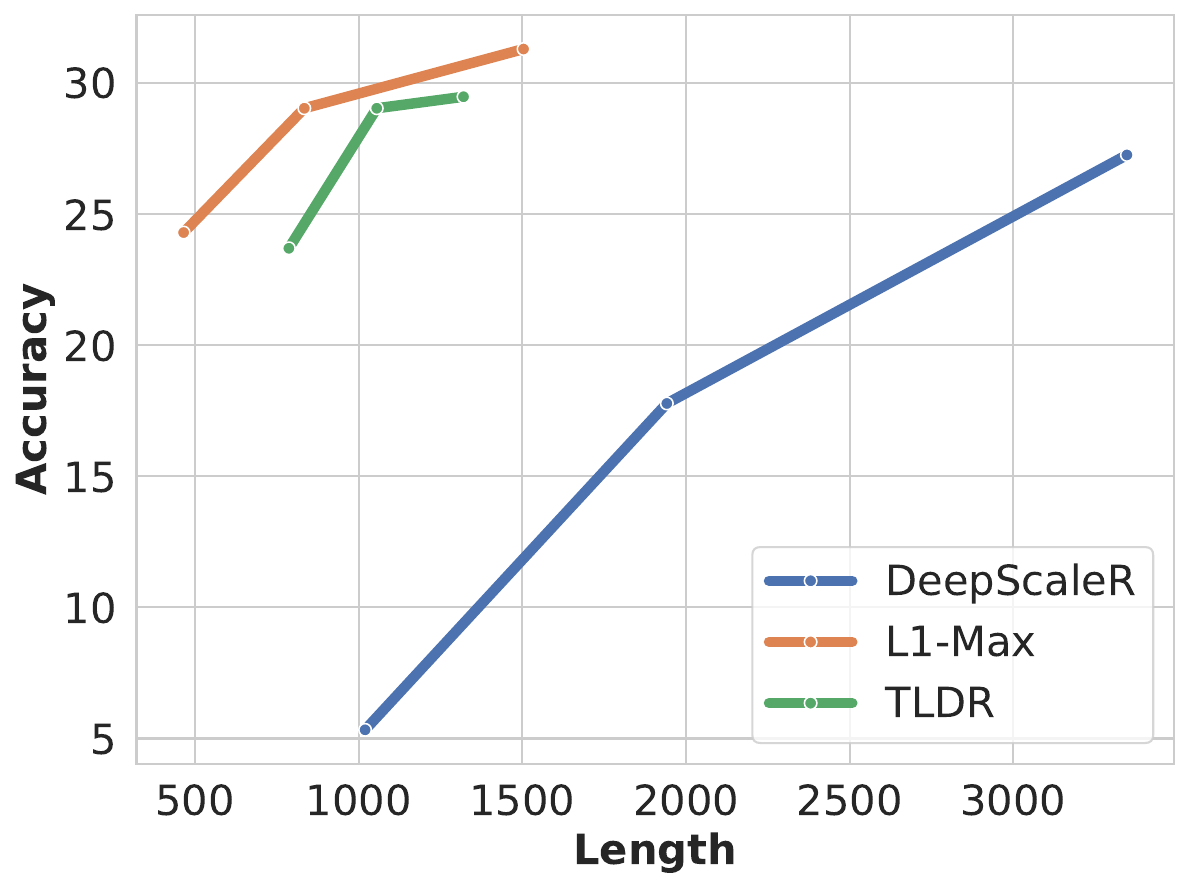}
\vspace{-15pt}
\caption{\small{OlympiadBench}}\label{fig:l1_compare_OlympiadBench}
\end{subfigure}
\caption{TLDR compares on par with L1 under an max trace length constraint. Moreover, TLDR requires only a single training phase of 100 epochs, whereas L1 involves a more complex two-phase training process: 700 epochs for the first phase (L1-Exact) and an additional 120 epochs for the second phase (L1-Max).}
\label{fig:l1_compare}
\vspace{-20pt}
\end{figure}

\begin{table}[h]
\tiny
\centering
\begin{subtable}{\textwidth}
\centering
\begin{tabular}{c|c|c|c|c|c|c|c|c|c|c|c|c}
    \hline
      &  \multicolumn{6}{|c|}{No Length Control} & \multicolumn{6}{|c}{With Length Control} \\ \hline
      &  \multicolumn{2}{|c|}{SHORT} & \multicolumn{2}{|c|}{MODERATE} & \multicolumn{2}{|c|}{LONG} &  \multicolumn{2}{|c|}{SHORT} & \multicolumn{2}{|c|}{MODERATE} & \multicolumn{2}{|c}{LONG} \\ \hline
      & acc & length & acc & length & acc & length & acc & length & acc & length & acc & length \\ \hline
    DeepSeek-R1-Distill-Qwen-1.5B-2k & 0.0 & 512 & 13.2 & 1017.756 & 44.4 & 1832.292 & 54.6 & 307.862 & 63.8 & 373.63 & 72 & 595.232 \\ \hline
    DeepSeek-R1-Distill-Qwen-1.5B-4k & 13.2 & 1017.756 & 44.4 & 1832.292 & 64.0 & 2781.848 & 60.0 & 650.08 & 65.0 & 868.134 & 74.4 & 1194.83 \\ \hline
    DeepSeek-R1-Distill-Qwen-1.5B-4k-new & & & & & & & 62.4 & 714.17 & 72.4 & 927.476 & 78.4 & 1285.882 \\ \hline
    DeepSeek-R1-Distill-Qwen-7B-4k & 12.2 & 1019.408 & 48.4 & 1805.768 & 71.4 & 2631.856 & 70.2 & 351.828 & 74.4 & 433.952 & 77.2 & 673.16 \\ \hline
    DeepScaleR-1.5B-Preview-4k & 28.8 & 966.116 & 59.4 & 1618.526 & 74 & 2329.458 & 68.6 & 577.702 & 75.2 & 706.082 & 74.8 & 857.298 \\ \hline
    L1-Qwen-1.5B-Max-4k & \multicolumn{6}{|c|}{NA} & 70.4 & 351.774 & 76.2 & 656.242 & 80.2 & 1353.576 \\ \hline
        S1-Qwen-1.5B-BudgetForcing-4k & \multicolumn{6}{|c|}{NA} & 43.60 & 876.31 & 66.20 & 1360.79 & 76.80 & 1936.26 \\ \hline
    S1-Qwen-7B-BudgetForcing-4k & \multicolumn{6}{|c|}{NA} & 48.20 & 871.75 & 72.00 & 1366.49 & 83.00 & 2044.31 \\ \hline
    DeepSeek-R1-Distill-Qwen-1.5B-prompt-4k & \multicolumn{6}{|c|}{NA} & 32.20 & 890.22 & 55.60 & 1441.21 & 69.60 & 2108.98 \\ \hline
    DeepSeek-R1-Distill-Qwen-7B-prompt-4k & \multicolumn{6}{|c|}{NA} & 28.20 & 918.25 & 51.20 & 1480.86 & 76.40 & 2153.89 \\ \hline
\end{tabular}
\caption{MATH500}
\label{tab:multi-level_length_control_MATH500}
\end{subtable}
\begin{subtable}{\textwidth}
\centering
\begin{tabular}{c|c|c|c|c|c|c|c|c|c|c|c|c}
    \hline
      &  \multicolumn{6}{|c|}{No Length Control} & \multicolumn{6}{|c}{With Length Control} \\ \hline
      &  \multicolumn{2}{|c|}{SHORT} & \multicolumn{2}{|c|}{MODERATE} & \multicolumn{2}{|c|}{LONG} &  \multicolumn{2}{|c|}{SHORT} & \multicolumn{2}{|c|}{MODERATE} & \multicolumn{2}{|c}{LONG} \\ \hline
      & acc & length & acc & length & acc & length & acc & length & acc & length & acc & length \\ \hline
    DeepSeek-R1-Distill-Qwen-1.5B-2k & 0.0 & 512 & 3.6 & 1024 & 16.9 & 1988.036 & 19.28 & 444.47 & 42.17 & 650.19 & 51.81 & 1041.90 \\ \hline
    DeepSeek-R1-Distill-Qwen-1.5B-4k & 3.6 & 1024 & 16.9 & 1988.036 & 32.5 & 3690.096 & 32.53 & 939.39 & 46.48 & 1423.66 & 59 & 2043.96 \\ \hline
    DeepSeek-R1-Distill-Qwen-1.5B-4k-new & & & & & & & 33.73 & 944.53 & 51.81 & 1536.16 & 63.86 & 1937.98 \\ \hline
    DeepSeek-R1-Distill-Qwen-7B-4k & 2.4 & 1024 & 19.3 & 1984.229 & 55.4 & 3461.916 & 51.81 & 524.42 & 54.22 & 736.25 & 65.06 & 1125.60 \\ \hline
    DeepScaleR-1.5B-Preview-4k & 6.02 & 1018.48 & 24.10 & 1948.16 & 50.60 & 3271.06 & 36.15 & 838.92 & 56.63 & 1162.20 & 65.06 & 1374.27 \\ \hline
    L1-Qwen-1.5B-Max-4k & \multicolumn{6}{|c|}{NA} & 46.99 & 472.37 & 60.24 & 848.96 & 66.27 & 1584.44 \\ \hline
        S1-Qwen-1.5B-BudgetForcing-4k & \multicolumn{6}{|c|}{NA} & 14.46 & 964.06 & 31.33 & 1624.02 & 45.78 &  2553.99\\ \hline
    S1-Qwen-7B-BudgetForcing-4k & \multicolumn{6}{|c|}{NA} & 13.25 & 960.19 & 36.14 & 1636.60 & 48.19 & 2665.64 \\ \hline
    DeepSeek-R1-Distill-Qwen-1.5B-prompt-4k & \multicolumn{6}{|c|}{NA} & 6.02 & 999.94 & 18.07 & 1906.54 & 39.76 & 3354.06 \\ \hline
    DeepSeek-R1-Distill-Qwen-7B-prompt-4k & \multicolumn{6}{|c|}{NA} & 6.02 & 1003.98 & 18.07 & 1908.93 & 43.37 & 3284.07 \\ \hline
\end{tabular}
\caption{AMC}
\label{tab:multi-level_length_control_AMC}
\end{subtable}
\begin{subtable}{\textwidth}
\centering
\begin{tabular}{c|c|c|c|c|c|c|c|c|c|c|c|c}
    \hline
      &  \multicolumn{6}{|c|}{No Length Control} & \multicolumn{6}{|c}{With Length Control} \\ \hline
      &  \multicolumn{2}{|c|}{SHORT} & \multicolumn{2}{|c|}{MODERATE} & \multicolumn{2}{|c|}{LONG} &  \multicolumn{2}{|c|}{SHORT} & \multicolumn{2}{|c|}{MODERATE} & \multicolumn{2}{|c}{LONG} \\ \hline
      & acc & length & acc & length & acc & length & acc & length & acc & length & acc & length \\ \hline
    DeepSeek-R1-Distill-Qwen-1.5B-2k & 0 & 512 & 0 & 1024 & 3.33 & 2048 & 0 & 494.4 & 3.33 & 865.8 & 16.67 & 1441.37 \\ \hline
    DeepSeek-R1-Distill-Qwen-1.5B-4k & 0 & 1024 & 3.33 & 2048 & 13.33 & 3999.83 & 6.67 & 1007.83 & 13.3 & 1702.57 & 10 & 2637.7 \\ \hline
    DeepSeek-R1-Distill-Qwen-1.5B-4k-new & & & & & & & 6.67 & 1011.17 & 10 & 1880.8 & 16.67 & 3030.8 \\ \hline
    DeepSeek-R1-Distill-Qwen-7B-4k & 0 & 1024 & 10 & 2038.13 & 16.67 & 3835.0 & 10 & 764.33 & 13.33 & 1168.2 & 30 & 1936.93 \\ \hline
    DeepScaleR-1.5B-Preview-4k & 0 & 1024 & 3.33 & 2048 & 20 & 3919.03 & 6.67 & 999.53 & 13.33 & 1524.2 & 13.33 & 2043.4 \\ \hline
    L1-Qwen-1.5B-Max-4k & \multicolumn{6}{|c|}{NA} & 3.33 & 718.2 & 10 & 1238.8 & 10 & 1835.83 \\ \hline
    S1-Qwen-1.5B-BudgetForcing-4k & \multicolumn{6}{|c|}{NA} & 0 & 1004.10 & 0 & 1792.73 & 3.33 & 3033.50 \\ \hline
    S1-Qwen-7B-BudgetForcing-4k & \multicolumn{6}{|c|}{NA} & 0 & 1007.87 & 3.33 & 1737.03 & 20 & 2857.97 \\ \hline
    DeepSeek-R1-Distill-Qwen-1.5B-prompt-4k & \multicolumn{6}{|c|}{NA} & 0 & 1024.00 & 0 & 2048.00 & 13.33 & 3926.93 \\ \hline
    DeepSeek-R1-Distill-Qwen-7B-prompt-4k & \multicolumn{6}{|c|}{NA} & 0 & 1024.00 & 0 & 2048.00 & 16.67 & 3959.57 \\ \hline
\end{tabular}
\caption{AIME24}
\label{tab:multi-level_length_control_AIME24}
\end{subtable}
\begin{subtable}{\textwidth}
\centering
\begin{tabular}{c|c|c|c|c|c|c|c|c|c|c|c|c}
    \hline
      &  \multicolumn{6}{|c|}{No Length Control} & \multicolumn{6}{|c}{With Length Control} \\ \hline
      &  \multicolumn{2}{|c|}{SHORT} & \multicolumn{2}{|c|}{MODERATE} & \multicolumn{2}{|c|}{LONG} &  \multicolumn{2}{|c|}{SHORT} & \multicolumn{2}{|c|}{MODERATE} & \multicolumn{2}{|c}{LONG} \\ \hline
      & acc & length & acc & length & acc & length & acc & length & acc & length & acc & length \\ \hline
    DeepSeek-R1-Distill-Qwen-1.5B-2k & 0 & 512 & 2.2 & 1023.4 & 10.37 & 1999.238 & 15.11 & 412.58 & 20.59 & 564.63 & 24 & 941.90 \\ \hline
    DeepSeek-R1-Distill-Qwen-1.5B-4k & 2.2 & 1023.4 & 10.37 & 1999.238 & 18.81 & 3598.139 & 16.88 & 850.82 & 20.30 & 1245.40 & 25.48 & 1951.65 \\ \hline
    DeepSeek-R1-Distill-Qwen-1.5B-4k-new & & & & & & & 20 & 926.65 & 25.93 & 1363.68 & 28.30 & 1969.88 \\ \hline
    DeepSeek-R1-Distill-Qwen-7B-4k & 1.78 & 1023.25 & 11.7 & 1996.39 & 24.74 & 3504.83 & 26.52 & 491.54 & 26.22 & 643.62 & 28.59 & 1004.45 \\ \hline
    DeepScaleR-1.5B-Preview-4k & 5.33 & 1019.20 & 17.78 & 1941.47 & 27.26 & 3347.61 & 23.70 & 786.27 & 29.04 & 1054.94 & 29.48 & 1319.93 \\ \hline
    L1-Qwen-1.5B-Max-4k & \multicolumn{6}{|c|}{NA} & 24.30 & 464.51 & 29.04 & 833.43 & 31.3 & 1503.4 \\ \hline
        S1-Qwen-1.5B-BudgetForcing-4k & \multicolumn{6}{|c|}{NA} & 10.96 & 965.88 & 27.85 & 1616.35 & 40.44 & 2567.32 \\ \hline
    S1-Qwen-7B-BudgetForcing-4k & \multicolumn{6}{|c|}{NA} & 14.52 & 948.51 & 31.56 & 1573.08 & 47.56 & 2526.31 \\ \hline
    DeepSeek-R1-Distill-Qwen-1.5B-prompt-4k & \multicolumn{6}{|c|}{NA} & 4.30 & 1012.55 & 16.00 & 1915.82 & 30.67 & 3311.35 \\ \hline
    DeepSeek-R1-Distill-Qwen-7B-prompt-4k & \multicolumn{6}{|c|}{NA} & 3.41 & 1011.88 & 14.22 & 1948.05 & 36.30 & 3387.5311 \\ \hline
\end{tabular}
\caption{OlympiadBench}
\label{tab:multi-level_length_control_OlympiadBench}
\end{subtable}
\caption{Detailed performance of different models and different methods in four reasoning datasets. The first 3 rows are the results of original DeepSeek distilled models (No Length Control) and models finetuned with \alg, L1-Qwen-1.5B-Max-4k is the method in \cite{aggarwal2025l1}, 2 rows with S1 is the method proposed in \cite{muennighoff2025s1}, and the last 2 rows are controlling original DeepSeek distilled model by prompts. In the name of models, ``2k'' and ``4k'' means the max length of model output, and ``SHORT'', ``MODERATE'' and ``LONG'' means the $\frac{1}{4},\frac{1}{2}$ and the original max length. For different datasets, RL with our multi-level length penalty design can shorten response length while achieving good accuracy, particularly for the ``long'' length penalty setting.}
\label{tab:multi-level_length_control}
\end{table}

\begin{table}[h]
\scriptsize
\begin{subtable}{\textwidth}
\centering
\begin{tabular}{c|c|c|c|c|c|c|c|c|c}
    \hline
     &   \multicolumn{3}{|c|}{SHORT}   &   \multicolumn{3}{|c|}{MODERATE}   &   \multicolumn{3}{|c}{LONG}    \\ \hline
     & All  & Correct  & Wrong  &   All  & Correct  & Wrong & All  & Correct  & Wrong   \\ \hline
    Deepseek+TLDR & 714.17  & 611.71  & 884.21  & 927.476  & 772.36  & 1334.36 &  1285.882 & 1091.87 & 1990.06 \\ \hline
    DeepScaleR+TLDR & 1374.28 & 1145.57 & 1800.14 & 1374.28 & 1145.57 & 1800.14 & 1374.28 & 1145.57 & 1800.14  \\ \hline
    DeepScaleR & 966.116  &  831.90 & 1020.40  &  1618.526 & 1367.84  & 1985.29 & 2329.458  & 1857.26  &  3673.39  \\ \hline
    L1 & 351.774  & 291.85 & 494.30  & 656.242  & 568.94  & 935.76  &  1353.576 & 1287.46  & 1621.39  \\ \hline
    Deepseek & 1017.756  & 978.788  & 1023.682   &  1832.292 & 1587.640  & 2027.662  &  2781.848 & 2142.125   & 3919.133  \\ \hline
\end{tabular}
\caption{MATH500}
\label{tab:3level_length_MATH}
\end{subtable}

\begin{subtable}{\textwidth}
\centering
\begin{tabular}{c|c|c|c|c|c|c|c|c|c}
    \hline
     &   \multicolumn{3}{|c|}{SHORT}   &   \multicolumn{3}{|c|}{MODERATE}   &   \multicolumn{3}{|c}{LONG}    \\ \hline
     & All  & Correct  & Wrong  &   All  & Correct  & Wrong & All  & Correct  & Wrong   \\ \hline
    Deepseek+TLDR & 944.53  & 842.43  & 996.51  & 1536.16  & 1229.0  & 1866.35  & 1937.98  & 1607.68 & 2521.5 \\ \hline
    DeepScaleR+TLDR & 838.92 & 585.37 & 982.43 & 1162.20 & 919.89 & 1478.56 & 1374.28 & 1145.57 & 1800.14 \\ \hline
    DeepScaleR &  1018.48 &  932.4 & 1024.0  & 1948.16  & 1633.65  & 2048.0 & 3271.06  & 2494.64  &  4066.41  \\ \hline
    L1 & 472.37  & 417.26  & 521.23  & 848.96  &  736.02 & 1020.09  &  1584.45 & 1477.13  & 1795.25  \\ \hline
    Deepseek & 1024.0  & 1024.0  & 1024.0  & 1988.036  & 1692.5   & 2048.0  &  3690.096 & 2848.222    & 4096.0  \\ \hline
\end{tabular}
\caption{AMC}
\label{tab:3level_length_AMC}
\end{subtable}

\begin{subtable}{\textwidth}
\centering
\begin{tabular}{c|c|c|c|c|c|c|c|c|c}
    \hline
     &   \multicolumn{3}{|c|}{SHORT}   &   \multicolumn{3}{|c|}{MODERATE}   &   \multicolumn{3}{|c}{LONG}    \\ \hline
     & All  & Correct  & Wrong  &   All  & Correct  & Wrong & All  & Correct  & Wrong   \\ \hline
    Deepseek+TLDR &  1011.17 & 1021.5  & 1010.43  &  1880.8 & 1677.0  & 1903.44 & 3030.8  & 2228.0 & 3191.36 \\ \hline
    DeepScaleR+TLDR & 999.53 & 878.5 & 1008.18 & 1524.2 & 1343.75 & 1551.96 & 2043.4 & 1437.5 & 2136.62 \\ \hline
    DeepScaleR & 1024.0  & NA  & 1024.0  & 2048.0  & 2048.0  & 2048.0 & 3919.03  &  3686.0 & 3977.29   \\ \hline
    L1 & 718.2  & 772.0  & 716.34  & 1238.8  & 1117.0  &  1252.33 & 1835.83  & 1551.0  &  1867.48 \\ \hline
    Deepseek & 1024.0  & NA  & 1024.0  & 2048.0  & 2048.0  & 2048.0  & 3999.83  & 3396.25  &   4092.69  \\ \hline
\end{tabular}
\caption{AIME24}
\label{tab:3level_length_AIME}
\end{subtable}

\begin{subtable}{\textwidth}
\centering
\begin{tabular}{c|c|c|c|c|c|c|c|c|c}
    \hline
     &   \multicolumn{3}{|c|}{SHORT}   &   \multicolumn{3}{|c|}{MODERATE}   &   \multicolumn{3}{|c}{LONG}    \\ \hline
     & All  & Correct  & Wrong  &   All  & Correct  & Wrong & All  & Correct  & Wrong   \\ \hline
    Deepseek+TLDR & 926.64  & 793.81  & 959.85  & 1363.68  & 1038.28   & 1477.568 & 1969.88  & 1394.24 & 2197.04 \\ \hline
    DeepScaleR+TLDR & 786.27 & 559.76 & 856.64 & 1054.94 & 750.94 & 1179.33 & 1319.93 & 848.86 & 1516.87 \\ \hline
    DeepScaleR & 1019.20  & 957.11  & 1022.70  &  1941.47 & 1632.28  & 2008.33  & 3347.61  &  2397.49 &  3703.67  \\ \hline
    L1 & 464.51  & 362.54  & 497.24  & 833.43  &  656.24 & 905.92  & 1503.47  & 1329.72  & 1581.93  \\ \hline
    Deepseek & 1023.4  & 1003.66   & 1023.87  & 1999.238  &  1684.94 & 2035.60  & 3598.139  & 2574.03   & 3835.478   \\ \hline
\end{tabular}
\caption{OlympiadBench}
\label{tab:3level_length_OlympiadBench}
\end{subtable}

\caption{Average response length of models' responses in 4 datasets. We report the value in different levels and in the whole dataset, subset of correct responses and subset of wrong responses. \alg can reduce the response by at most 50\% from base models.}
\label{tab:3level_length}
\end{table}

\begin{table}[h]
\scriptsize
\begin{subtable}{\textwidth}
\centering
\begin{tabular}{c|c|c|c}
    \hline
     &  SHORT & MODERATE  & LONG      \\ \hline
    Deepseek+TLDR &  0.0053 & 0.0507  &    0.1296   \\ \hline
    DeepScaleR+TLDR  &  0.0127  & 0.0725  &  0.0635     \\ \hline
    DeepScaleR & 0.0168  & 0.0936  &  0.2692   \\ \hline
    L1 &  0.0135   & 0.0168  &  0.1313   \\ \hline
    Deepseek & 0.0138  & 0.1798  &  0.4111     \\ \hline
    
\end{tabular}
\caption{MATH500}
\label{tab:3level_repeat_MATH}
\end{subtable}

\begin{subtable}{\textwidth}
\centering
\begin{tabular}{c|c|c|c}
    \hline
     &  SHORT & MODERATE  & LONG      \\ \hline
    Deepseek+TLDR & 0.0000  & 0.0750  &  0.1333   \\ \hline
    DeepScaleR+TLDR  & 0.0189  & 0.1389  &  0.1379     \\ \hline
    DeepScaleR & 0.0128  & 0.0635 & 0.3170    \\ \hline
    L1 & 0.0227  & 0.0606  &  0.0714   \\ \hline
    Deepseek &  0.0000 & 0.2029  &  0.5000     \\ \hline
    
\end{tabular}
\caption{AMC}
\label{tab:3level_repeat_AMC}
\end{subtable}

\begin{subtable}{\textwidth}
\centering
\begin{tabular}{c|c|c|c}
    \hline
     &  SHORT & MODERATE  & LONG      \\ \hline
    Deepseek+TLDR &  0.0357 & 0.0370  &  0.2000     \\ \hline
    DeepScaleR+TLDR  & 0.0357  & 0.1923  &  0.2308     \\ \hline
    DeepScaleR & 0.0000  & 0.1379  &  0.1667   \\ \hline
    L1 & 0.0345  &  0.0742 &   0.2592   \\ \hline
    Deepseek &  0.0333 & 0.2759  &  0.3462     \\ \hline
    
\end{tabular}
\caption{AIME24}
\label{tab:3level_repeat_AIME}
\end{subtable}

\begin{subtable}{\textwidth}
\centering
\begin{tabular}{c|c|c|c}
    \hline
     &  SHORT & MODERATE  & LONG      \\ \hline
    Deepseek+TLDR & 0.0333  &  0.0720 &   0.1756    \\ \hline
    DeepScaleR+TLDR  & 0.0330  & 0.1023  &  0.1492     \\ \hline
    DeepScaleR &  0.0125 & 0.1333  &   0.2770  \\ \hline
    L1 & 0.0117  & 0.0355  &  0.1290   \\ \hline
    Deepseek & 0.0167  &  0.1983 &  0.3923     \\ \hline
    
\end{tabular}
\caption{OlympiadBench}
\label{tab:3level_repeat_OlympiadBench}
\end{subtable}

\caption{Repeat rate of models' responses in 4 datasets. Repeat rate is calculated using GPT-4o to identify significant repetitions of identical sentences in each response, excluding short phrases like 'wait'. \alg can reduce the response by at most $\frac{2}{3}$ from base models.}
\label{tab:3level_repeat}
\end{table}

\begin{table}[h]
\scriptsize
\begin{subtable}{\textwidth}
\centering
\begin{tabular}{c|c|c|c}
    \hline
     & SHORT  & MODERATE  & LONG    \\ \hline
    Deepseek+TLDR & 3.516  & 4.114  &  5.73  \\ \hline
    DeepScaleR+TLDR & 3  & 3.806  &   4.16 \\ \hline
    DeepScaleR & 4.856  & 7.768  & 11.062  \\ \hline
    L1 & 2.848  &  4.272 & 7.826 \\ \hline
    Deepseek &  4.22 &  7.72 & 11.64   \\ \hline
    
\end{tabular}
\caption{MATH500}
\label{tab:3level_step_MATH500}
\end{subtable}

\begin{subtable}{\textwidth}
\centering
\begin{tabular}{c|c|c|c}
    \hline
     & SHORT  & MODERATE  & LONG    \\ \hline
    Deepseek+TLDR & 4.75  & 6.75  & 8.81   \\ \hline
    DeepScaleR+TLDR & 4.25  & 5.89  &  7.12  \\ \hline
    DeepScaleR & 5.82  & 10.13  & 16.45  \\ \hline
    L1 & 3.86  & 5.76  & 9.57  \\ \hline
    Deepseek &  4.60 &  8.57 &  15.73  \\ \hline
    
\end{tabular}
\caption{AMC}
\label{tab:3level_step_AMC}
\end{subtable}

\begin{subtable}{\textwidth}
\centering
\begin{tabular}{c|c|c|c}
    \hline
     & SHORT  & MODERATE  & LONG    \\ \hline
    Deepseek+TLDR & 5.27  & 7.7  & 13.67   \\ \hline
    DeepScaleR+TLDR & 6.03  & 9.97  & 10.23   \\ \hline
    DeepScaleR &  6.6 & 11.53  & 20.13  \\ \hline
    L1 & 6.57  & 10.73  & 13.07  \\ \hline
    Deepseek & 5.27  &  9.37 & 18.97   \\ \hline
    
\end{tabular}
\caption{AIME24}
\label{tab:3level_step_AIME}
\end{subtable}

\begin{subtable}{\textwidth}
\centering
\begin{tabular}{c|c|c|c}
    \hline
     & SHORT  & MODERATE  & LONG    \\ \hline
    Deepseek+TLDR & 4.74  & 6.45  &  9.11  \\ \hline
    DeepScaleR+TLDR & 4.31  &  5.51 &  6.55  \\ \hline
    DeepScaleR & 5.69  & 10.45  &  17.43 \\ \hline
    L1 & 3.67  &  6.08 &  9.57 \\ \hline
    Deepseek &  4.39 & 8.94  &  16.93  \\ \hline
    
\end{tabular}
\caption{OlympiadBench}
\label{tab:3level_step_OlympiadBench}
\end{subtable}

\caption{Steps of models' responses in 4 datasets. We define the number of steps as the number of occurrence of keywords in the keyword list [``But", ``Wait,", ``Alternatively,", ``Perhaps", ``First,", ``Okay,", ``Given", ``The", ``Therefore,", ``So,"]. \alg can reduce the response by at most 50\% from base models.}
\label{tab:3level_step}
\end{table}

\end{document}